\documentclass{article} 
\usepackage{jmlr2e}

\usepackage{amsmath,amsfonts,bm}

\def\eqref#1{equation~\ref{#1}}

\def\1{\bm{1}}

\DeclareMathAlphabet{\mathsfit}{\encodingdefault}{\sfdefault}{m}{sl}
\SetMathAlphabet{\mathsfit}{bold}{\encodingdefault}{\sfdefault}{bx}{n}

\usepackage{soul}
\usepackage{hyperref}
\usepackage{amssymb}
\usepackage{booktabs}
\usepackage{pifont}
\usepackage{xspace}
\usepackage{graphicx}
\usepackage{makecell}
\usepackage{xcolor,colortbl}
\usepackage{arydshln} 
\usepackage{multirow}
\usepackage{longtable,tabu}
\usepackage{caption}
\usepackage{verbatim}
\usepackage{subcaption}
\usepackage{lscape}
\usepackage{longtable}
\usepackage{wrapfig}
\usepackage[normalem]{ulem}

\usepackage{url}

\definecolor{tblue}{HTML}{174992}

\newcommand{\ram}[1]{{\color{purple} [Ram: {#1}]}}
\newcommand{\vic}[1]{\textcolor{teal}{[Victoria: {#1}]}}

\newcommand{\OursLargeIcon}{OPT-IML\includegraphics[scale=0.068]{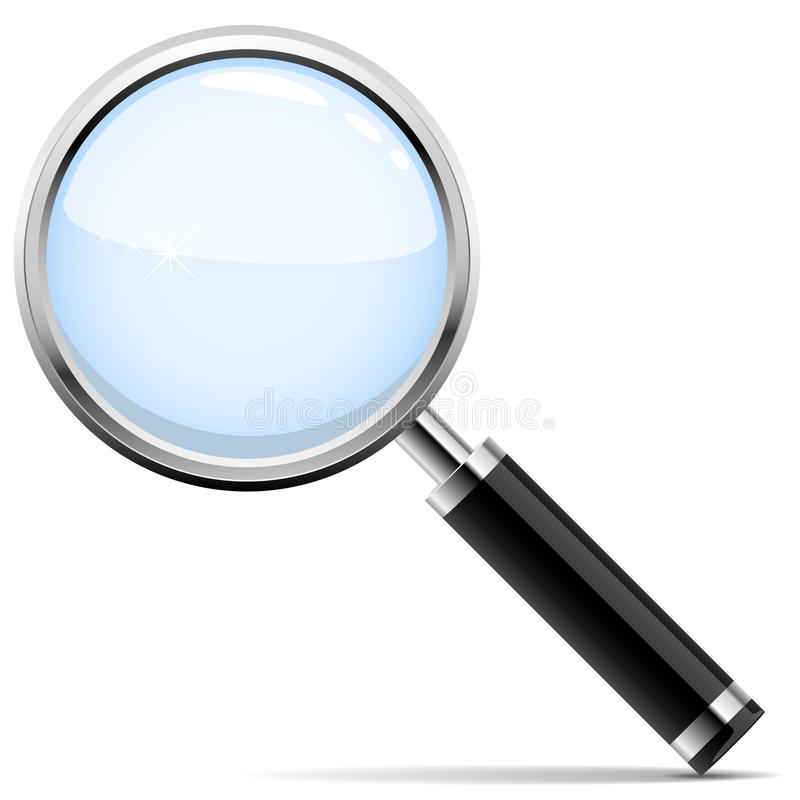}\xspace}
\newcommand{\Ours}{OPT-IML\xspace}
\newcommand{\OursM}{OPT-IML-Max\xspace}
\newcommand{\OursB}{\Ours Bench\xspace}

\newcommand{\tfive}{\textsc{T5}\xspace}

\newcommand{\tzero}{\textsc{T0}\xspace}
\newcommand{\tkinstruct}{T$k$-\textsc{Instruct}\xspace}
\newcommand{\flanpalm}{FLAN-PaLM\xspace}
\newcommand{\flantfive}{FLAN-T5\xspace}

\newcommand{\natins}{Super-NaturalInstructions\xspace}
\newcommand{\natinsShort}{SuperNatInst}
\newcommand{\promptsource}{PromptSource\xspace}
\newcommand{\crossfit}{CrossFit\xspace}
\newcommand{\flan}{FLAN\xspace}
\newcommand{\exmix}{ExMix\xspace}

\newcommand{\unifiedskg}{UnifiedSKG\xspace}

\newcommand{\cotr}{Chain-of-thought Reasoning\xspace}
\newcommand{\cotrShort}{Reasoning\xspace}

\newcommand{\highest}[1]{\textbf{#1}}
\newcommand{\hide}[1]{}

\newcommand{\sep}{{\small\texttt{[Sep]}}}
\newcommand{\outlier}{$^{\dagger}$}
\newcommand{\outliertwo}{$^{\ddag}$}

\definecolor{DarkGreen}{RGB}{30,130,30}

\definecolor{cadmiumgreen}{rgb}{0.0, 0.42, 0.24}
\definecolor{oldlace}{rgb}{0.99, 0.96, 0.9}

\extrafloats{1000}
\usepackage[T1]{fontenc}

\title{\OursLargeIcon: Scaling Language Model Instruction Meta Learning through the Lens of Generalization}

\author{ \centering
Srinivasan Iyer\footnote{Equal contribution; alphabetical order.  }, Xi Victoria Lin$^*$, Ramakanth Pasunuru$^*$, \\Todor Mihaylov, D\'aniel Simig, Ping Yu, Kurt Shuster, Tianlu Wang, Qing Liu, \\ Punit Singh Koura, Xian Li, Brian O'Horo, Gabriel Pereyra\thanks{Work done while at Meta AI.}\footnotemark[2]{\normalfont}, Jeff Wang, \\Christopher Dewan, Asli Celikyilmaz, Luke Zettlemoyer, Ves Stoyanov\footnotemark[2]{\normalfont}  \AND \begin{center} 
\addr{Meta AI}\end{center} 
}

\begin{document}

\editor{}
\maketitle

\begin{abstract}


Recent work has shown that fine-tuning large pre-trained language models on a collection of tasks described via instructions, a.k.a. instruction-tuning, improves their zero and few-shot generalization to unseen tasks. However, there is a limited understanding of the performance trade-offs of different decisions made during the instruction-tuning process. These decisions include the scale and diversity of the instruction-tuning benchmark, different task sampling strategies, fine-tuning with and without demonstrations, training using specialized datasets for reasoning and dialogue, and finally, the fine-tuning objectives themselves. In this paper, we characterize the effect of instruction-tuning decisions on downstream task performance when scaling both model and benchmark sizes. To this end, we create \Ours Bench: a large benchmark for Instruction Meta-Learning (IML) of 2000 NLP tasks consolidated into task categories from 8 existing benchmarks, and prepare an evaluation framework to measure three types of model generalizations: to tasks from fully held-out categories, to held-out tasks from seen categories, and to held-out instances from seen tasks. Through the lens of this framework, we first present insights about instruction-tuning decisions as applied to OPT-30B and further exploit these insights to train \Ours 30B and 175B, which are instruction-tuned versions of OPT. \Ours demonstrates all three generalization abilities at both scales on four different evaluation benchmarks with diverse tasks and input formats -- \promptsource, \flan, \natins, and \unifiedskg. Not only does it significantly outperform OPT on all benchmarks but is also highly competitive with 
existing models fine-tuned on each specific benchmark. We release \Ours at both scales, together with the \OursB evaluation framework.

\end{abstract}

\section{Introduction}



Instruction fine-tuning is shown \citep{wei2021flan,sanh2021t0,chung2022scaling} to significantly improve the zero- and few-shot performance of large pretrained LMs (LLM). It involves fine-tuning LLMs on collections of NLP tasks using instructional style input formats. Successful instruction-tuning of LLMs depends on a number of aspects such as the objectives used for fine-tuning, the distribution and diversity of the fine-tuning tasks, the inclusion of specialized datasets related to reasoning and dialogue, fine-tuning with demonstrations, and also, the comprehensiveness of the evaluation framework. In this paper, we develop an extensive large-scale fine-tuning and evaluation framework of 2000 NLP tasks (which we call \OursB) and use it to characterize the tradeoffs of different decisions relating to instruction meta-learning (IML) on the OPT models \citep{zhang2022opt}. We exploit insights gathered from this process, to train \Ours 30B and 175B, 
instruction-tuned versions of OPT.

\begin{figure}[ht]
    \centering
    \resizebox{\columnwidth}{!}{
    \includegraphics[width=\textwidth]{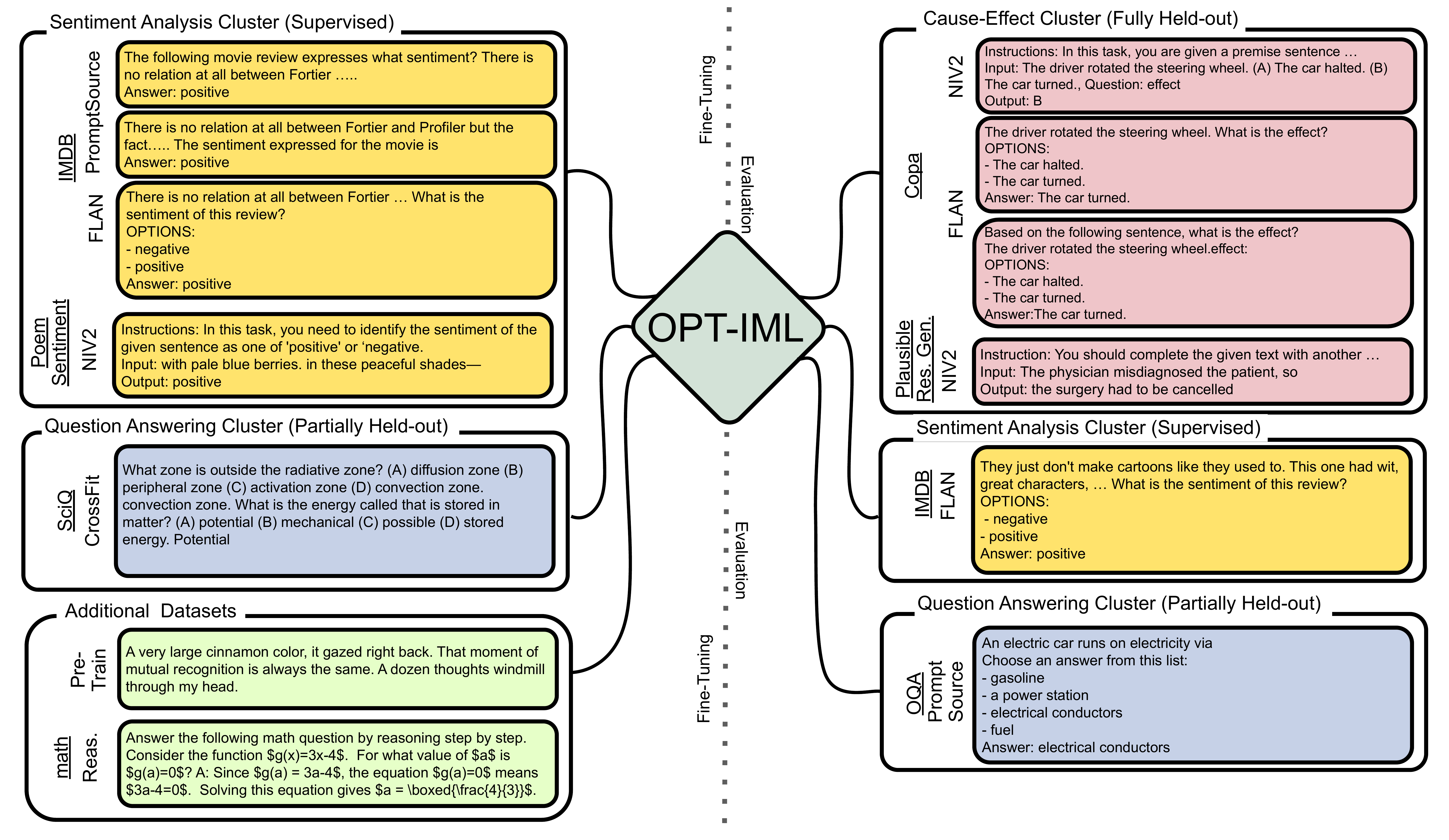}}
    \caption{We fine-tune OPT on a large collection of 1500+ NLP tasks divided into task categories (left hand side) to create \Ours. Each category contains multiple related tasks, as well as multiple prompts for the same task (e.g. IMDB), aggregated from multiple benchmarks. We evaluate \Ours on a set of evaluation categories (right hand-side) which can be disjoint, partially overlap or fully-overlap with the categories used for tuning (e.g. Sentiment Analysis fully overlaps and QA partially overlaps), corresponding to evaluating model generalization to tasks from fully held-out categories, to tasks from categories seen during training, and to instances from tasks seen during training. We release this evaluation framework as \OursB.}
    \label{fig:flow}
\end{figure}


There are a growing number of large meta-datasets of NLP tasks such as \natins \citep{wang2022niv2}, FLAN \citep{wei2021flan} and PromptSource \citep{sanh2021t0}. Recent instruction-tuning work has demonstrated success using these individual benchmarks and their combinations~\citep{chung2022flanpalm}, with a general recommendation for scaling up the number of tasks. We follow this recommendation by consolidating 8 meta-datasets into a large collection of 1,991 NLP tasks containing instructions with multiple prompts and grouping them into more than 100 task categories 
such as Question Answering and Sentiment Analysis (Figure \ref{fig:flow}). Furthermore, we transform this collection into an evaluation framework for comprehensively evaluating large-scale instruction-tuned models across three levels of generalization: 1) model performance on tasks from fully held-out task categories not used for tuning, as in prior work \citep{wei2021flan,sanh2021t0}, and additionally, 2) performance on unseen tasks from categories seen during instruction-tuning, and, 3) performance on held-out instances of tasks seen during tuning. The former two settings evaluates the cross-task generalization of instruction-tuning while the last setting evaluates 
the generalization of supervised multi-task learning~\citep{mccann2018decathlon}.
We refer to the resulting instruction-tuning framework as \OursB and illustrate its composition in Figure \ref{fig:flow} where the right hand side depicts evaluation categories, which can be completely disjoint, partially overlap, or completely overlap with the categories used for tuning on the left. Each category comprises datasets that can belong to multiple benchmarks and be associated with multiple prompts.


The effectiveness of instruction-tuning on LLMs depends on factors such as the diversity and distribution of tuning-tasks, the formatting of their prompts, and the objectives used for fine-tuning. Several recent works on instruction-tuning explore these factors 
by grouping tasks into categories and evaluating performance on tasks from completely held-out task categories \citep{sanh2021t0,wei2021flan,wang2022niv2}. Using our evaluation framework that considers multiple levels of generalization, we are able to comprehensively characterize the tradeoffs relating to these different factors when scaling up instruction-tuning to an aggregate of 8 different benchmarks. By instruction tuning OPT 30B \citep{zhang2022opt} on \OursB, we outline the tradeoffs of dataset and benchmark sampling strategies during tuning, the scaling laws with respect to tasks and categories, the effects of approaches to incorporating task demonstrations into instruction-tuning based on \cite{min2021metaicl}, as well as instruction-tuning with specialized datasets that contain reasoning chains \citep{kojima2022large, wei2022cot} and dialogue \citep{shuster2022blenderbot}. These experiments can serve to establish best practices for large scale instruction-tuning of LLMs.

Given the insights gathered from our generalization experiments on \Ours bench, we train \Ours. \Ours significantly improves over its base pre-trained model at both 30B and 175B scales 
on four different instruction-tuning benchmarks: PromptSource \citep{sanh2021t0}, FLAN \citep{wei2021flan}, \natins \citep{wang2022niv2}, and UnifiedSKG \citep{xie2022unifiedskg}. Additionally, the \Ours models also perform competitively in comparison with each of the prior instruction-tuned models individually tuned on these benchmarks on both zero and few-shot performance. Recently, along similar lines as this work, \cite{chung2022flanpalm} achieve impressive gains on the challenging 
benchmarks of MMLU \citep{hendrycks2020measuring} and Big-Bench Hard \citep{suzgun2022challenging} by instruction-tuning PaLM \citep{chowdhery2022palm} and T5 \citep{raffel2020t5} on a scaled-up collection of 1.8K tasks. \Ours trained under similar settings still 
underperforms in comparison on these challenging benchmarks and we discuss this in Section \ref{sec:discussion}. Following OPT \citep{zhang2022opt}, we will responsibly share versions of \Ours at both scales, and also release our \OursB evaluation framework to facilitate future work in this direction. 

\section{Scaling up Multi-task Benchmarks}
\label{sec:scaled-multi-task-benchmark}

To characterize the effects of extreme task scaling on instruction tuning, we build on recent task collections such as \natins \citep{wang2022niv2} and \promptsource \citep{sanh2021t0}, and aggregate 8 such collections to create the \Ours Benchmark for massive instruction fine-tuning and evaluation over diverse task categories, instruction types and prompting setups (Table \ref{tab:benchmark_compilation}). 

For the remainder of this paper, we use the terms task and dataset interchangeably; each task/dataset can be instantiated using multiple prompt templates. We refer to the original data from which the tasks are created as a data source; multiple tasks can be created from the same data source (e.g. question answering and question rewriting).  A benchmark comprises multiple tasks, where each task belongs to a single task category/cluster. 

\subsection{Task Curation}
\label{sec:unifying_benchmarks}
We expand the \natins benchmark of 1600+ tasks by~\citet{wang2022niv2} with the task collections from multiple existing work on \emph{instruction-tuning}: \flan~\citep{wei2021flan}, \tzero~\citep{sanh2021t0}; \emph{prompt crowdsourcing}: \promptsource~\citep{bach2022promptsource};
\emph{cross-task transfer studies}: \exmix~\citep{aribandi2021ext5}, \tfive~\citep{raffel2020t5}, \crossfit~\citep{ye2021crossfit}; and \emph{area-specific task consolidation}: Structured Knowledge Grounding~\citep{xie2022unifiedskg}, Dialogue~\citep{shuster2022blenderbot} and \cotr\footnote{We use 14 \cotr datasets which form a superset of those used by~\cite{chung2022flanpalm} (Appendix~\ref{subsec:reasoning-benchmark}).}~\citep{chung2022flanpalm}. The curation process of all these benchmarks can be found in Appendix~\ref{sec:data_curation}. \\


There is a significant overlap between the datasets in these benchmarks. For example, popular datasets such as SQuAD v1/v2~\citep{rajpurkar2016squad,rajpurkar2018squadv2} appear in almost all benchmarks. In addition, while \natins, \promptsource, \flan and \cotr contain long-form human-written instructions or reasoning chains, the rest of the benchmarks are designed for multi-task learning and the prompt templates often only consist of short field or task prefixes (e.g. ``question:'', ``label:''). Therefore, we only kept tasks from the \crossfit, \exmix and \tfive collections that do not appear in any other benchmarks. Since we're exploring a large number of tasks, we take maximally 100k examples (at random) per task from all benchmarks except \flan, where we take maximally 30k examples per task following the same practice as~\cite{wei2021flan}.

\begin{table*}[ht]
    \centering
    \small 
    \resizebox{.94\linewidth}{!}{
        \begin{tabular}{lccccccc}
            \toprule
            \multirow{2}{*}{Benchmark} & \multirow{2}{*}{\makecell{Instruct. \\ type}} & \multirow{2}{*}{\makecell{\# \\ clusters}} & \multirow{2}{*}{\makecell{\# \\ tasks}} & \multirow{2}{*}{\makecell{\# total \\ examples}} & 
            \multirow{2}{*}{\makecell{Avg. \# \\ prompts / task}} & \multicolumn{2}{c}{prompt length} \\ 
            & & & & & & mean & std \\ 
             \cmidrule(lr){1-1}  \cmidrule(lr){2-2}  \cmidrule(lr){3-3}  \cmidrule(lr){4-4}  \cmidrule(lr){5-5} \cmidrule(lr){6-6} \cmidrule(lr){7-8} 
             \natins   & task inst. & 76 & 1613 & 12.4M & 1.0 & 287 & 882 \\ 
             \promptsource  & instance inst. & 51 & 280 & 12.8M & 5.7 & 179 & 222 \\ 
             \crossfit & keywords & 32 & 159 & 7.1M & 1.0 & 117 & 258 \\ 
             \flan  & instance inst. & 12 & 70 & 4.4M & 8.5 & 193 & 375 \\ 
             \exmix\outliertwo  & keywords & 10 & 14 & 0.5M & 1.0 & 132 & 191 \\ 
             \tfive  & keywords & 9 & 36 & 1.9M & 1.0 & 111 & 167 \\ 
             \unifiedskg  & keywords & 7 & 21 & 0.8M & 1.0 & 444 & 297 \\ 
             \cotrShort  & task inst. & 1 & 14 & 0.4M & 1.0 & 146 & 122 \\ 
             \hdashline
             \OursB (train) & mixed & 93\outlier & 1,545 & 17.9M & 1.7 & 261 & 631 \\ 
             \OursB (dev) & mixed & 7 & 35 & 145K & 2.9 & -- & -- \\ 
             \OursB (test) & mixed & 10 & 87 & 321K & 4.6 & -- & -- \\ 
             \bottomrule 
        \end{tabular}
    }
    \caption{
        Details of \OursB. The statistics of each existing benchmark is calculated using the original data we downloaded. The statistics of \OursB is calculated using the data after we performed task filtering and taking a maximum of $M$ examples per tasks. For all benchmarks except FLAN, we set $M=100$k; for FLAN, we set $M=30$k following~\cite{wei2021flan}. 
        \outlier We only manually unify the task categorization in our evaluation sets. The estimation of the number of task clusters in our train set is based on a coarse union of the clustering tags from each original benchmark. 
    }
    \label{tab:benchmark_compilation}
\end{table*}

\subsection{Benchmark Consolidation}


\paragraph{Instruction schema.} 
Each benchmark adopts different instruction and language styles.
In Table~\ref{tab:prompt_variation}, we broadly classify their instructions into two categories: dataset-level and instance-level. \emph{Dataset-level instructions} define the overall task and may include auxiliary information such as positive/negative examples and explanations. The model is expected to learn the definition of the task based on this and apply the knowledge to each example coming after it. \emph{Instance-level instructions} are templates to be instantiated for each example individually and is sometimes designed in the cloze-style to solicit the desired output for the example.
We cast all tasks across the benchmarks we collect into the bipartite prompt formulation that include ``instructions'' and ``output'' segments (Table~\ref{tab:prompt_variation}). 
For \crossfit, \exmix and \tfive, since the original benchmarks do not provide natural language instructions, we manually write a simple instruction sentence for each of the included tasks and use them at the instance level. For example, the instructions for the GPT-2 Deepfake Detection task~\citep{radfordgpt2output} in \exmix reads ``Is the following text produced by GPT-2?''.

\paragraph{Task categorization.} 
We categorize the tasks under the conventional NLP categories following the practice of previous work~\citep{wei2021flan,sanh2021t0,wang2022niv2,ye2021crossfit}. Such grouping offers a convenient scaffold to study the generalization of models cross- and within categories.
We primarily follow the 76-category taxonomy defined by \natins. The other benchmarks also provide their own task clusters. We perform a coarse unification of the task clusters manually, e.g. merging ``hate speech detection'' with ``toxic language detection''. Besides this, benchmarks such as \crossfit and \promptsource adopt a finer-grained task categorization compared to \natins, e.g. \crossfit identifies multiple sub-classes of Question Answering. In such cases, we adopt the more coarse-grained assignment of \natins. This results in a single-level taxonomy with over 100 task categories (Table~\ref{tab:benchmark_compilation}). 

\begin{table*}[ht]
    \resizebox{\linewidth}{!}{
    \begin{tabular}{lp{0.14\textwidth}|p{0.65\textwidth}p{0.15\textwidth}}
        \toprule
         & \makecell{Inst. Type} & Instructions & Output \\
         \cmidrule(lr){1-1}  \cmidrule(lr){2-2}  \cmidrule(lr){3-3}  \cmidrule(lr){4-4}
         \natinsShort & task-level inst. & Instructions: Given a premise and two alternatives, choose the alternative that is a more plausible cause or effect of the situation described by the premise. The input format is ``premise (1) alternative\_1 (2) alternative\_2", the output should either be ``1" or ``2" based on your judgment.\newline Input: \dotuline{The terrorist set off the bomb.} (1) \dotuline{The bomb exploded.} (2) \dotuline{The bomb was deactivated.} & 1 \\
         \midrule
         \promptsource & instance-level inst. & Exercise: choose the most plausible alternative. \sep \dotuline{The terrorist set off the bomb.}  so... \sep - \dotuline{The bomb exploded.}\sep - \dotuline{The bomb was deactivated.} & The bomb exploded.\\
         \midrule
         \flan & instance-level inst. & \dotuline{The terrorist set off the bomb.} What is the effect?\sep OPTIONS: - \dotuline{The bomb exploded.} - \dotuline{The bomb was deactivated.} & The bomb exploded.\\
         \midrule
         \crossfit & keywords & \dotuline{The terrorist set off the bomb.} (A) \dotuline{The bomb exploded.} (B) \dotuline{The bomb was deactivated.} &  The bomb exploded.\\
        \bottomrule
    \end{tabular}
    }
    \caption{Different prompt formulations of the COPA task~\citep{roemmele2011copa} from \natins, \promptsource, \flan and \crossfit. \crossfit does not provide natural language instructions, which requires the models to rely on the data presentation to infer task requirements. 
    }
    \label{tab:prompt_variation}
\end{table*}

\subsection{Creating Benchmark Splits}
\label{sec:evaluation_splits}
\paragraph{Train, validation and test splits.} We split the set of all tasks in a way that allows us to perform massive instruction fine-tuning and evaluate the resulting model with respect to three levels of generalization. 
First, we hold out several task categories to evaluate model generalization to \emph{new categories of tasks}. Second, we select a subset of the remaining categories as partially held-out categories.\footnote{We manually examined the full task collection to eliminate false negatives for the held-out and partially held-out categories.} We divide the datasets in these categories into train and evaluation and use them to test model generalization to \emph{new datasets from seen task categories}. 
We select the fully and partially held-out categories by largely staying consistent with previous instruction fine-tuning work~\citep{wang2022niv2,wei2021flan,sanh2021t0} to allow direct comparison. 
Finally, for a subset of the training tasks, we hold out the validation and test sets from the original data release, and use them to test model generalization in the standard multi-task learning setting, i.e. \emph{new examples from seen tasks}. 
We reserve 35 evaluation tasks spanning 9 task categories from the evaluation tasks as the validation set\footnote{We also added the validation split of the Measuring Massive Multitask Language Understanding benchmark~\cite{hendryckstest2021mmlu} in our experiments in \S\ref{sec:ablation}.}, and use them to characterize the tradeoffs of different instruction-tuning strategies in \S\ref{sec:ablation}. The details of our validation tasks including their evaluation metrics are shown in Table~\ref{tab:validation_tasks}.

\paragraph{Task de-duplication.} We make sure that the train and evaluation tasks do not overlap on the data source they were created from, to prevent leakage\footnote{This condition is maintained for our partially held-out evaluation tasks as well.}, following the practice of~\cite{wang2022niv2}. For each pair of train and eval tasks, we compute the fraction of examples that have any 13-gram overlap between the instantiated sequences from those examples. We manually examine every pair where more than $1\%$ of the eval set overlaps with the training set ($\sim$14,000 pairs) to confirm whether tuning on the train task can unfairly benefit the eval task, and decide either to remove the train or the eval task in confirmed cases. The task pairs that share a broad contextual resource such as Wikipedia but otherwise contain unrelated output labels are retained. Table~\ref{tab:benchmark_compilation} shows the statistics of our task splits.

\subsection{Task Prompt Construction}
Each example in the zero-shot setting is formatted using the bipartite instruction scheme as described in Section \ref{sec:unifying_benchmarks}. We insert a delimiter between the instructions and the output if the instructions do not end with a ``:''. Similar to~\cite{chung2022flanpalm}, for each example we randomly sample a delimiter from a small set\footnote{The set includes ``$\backslash$nAnswer:'', `` Answer:'', ``$\backslash$nA:'', `` A:'', ``$\backslash$nOutput:'', `` Output:'', ``$\backslash$nanswer:'', ``$\backslash$output:''.} to mitigate overfitting. 
For few-shot prompts, we place the demonstration examples between the task descriptions and the target example for benchmarks that adopt task-level instructions such as \natins, and before the task example for benchmarks that adopt instance-level instructions such as \flan and \promptsource.
Examples of prompts for each of the tasks can be found in Appendix \ref{sec:templates-and-example-prompts}.

The \flan and \promptsource benchmarks contain multiple manually-written templates per task. To further increase task diversity, some templates in these benchmarks altered the original task semantics (e.g. ``question answering'' $\rightarrow$ ``question generation''). We manually examined all task templates in these benchmarks and removed the templates that altered the original task semantics to refine our task categories.


\section{Instruction Fine-tuning}
\label{sec:modeling}

We use the \OursB presented in Section \ref{sec:scaled-multi-task-benchmark} to fine-tune OPT \citep{zhang2022opt}, a suite of open-source decoder-only transformer language models released in scales from 125M to 175B parameters that performs similar to GPT-3 \citep{brown2020gpt} on a collection of standard NLP tasks. OPT is trained on 180B unique tokens from a combination of the datasets used in RoBERTa \citep{liu2019roberta}, the Pile \citep{gao2020pile}, and PushShift.io Reddit \citep {baumgartner2020pushshift,roller2020recipes} using a next-word prediction objective. 
We describe the process of instruction-tuning OPT at the scales of 30B and 175B in this section. 

\subsection{Fine-tuning Objective}

We finetune OPT in a manner similar to pre-training using a next-word prediction objective conditioned on all previous tokens as context. However, we separate the training sequence into a source context sequence and a target sequence and only include loss terms from the tokens in the target sequence (label-loss). We treat the task instructions and inputs as source tokens and the label tokens as target tokens. Formally, for a fine-tuning dataset $\mathcal{D}$ comprising source instances $s_i$ and their corresponding target tokens $t_i=\{t_{ij}\}$, a pre-trained model with parameters $\theta$ is fine-tuned to minimize the following loss over the target tokens conditioned on the source tokens and previously seen target tokens.

\begin{align}
    \mathcal{L}(\mathcal{D}; \theta) = - \sum_i \sum_j \log p_{\theta}(t_{ij}|s_i,t_{i,<j})
\end{align}

We minimize this loss across all datasets in our \OursB by mixing examples from different datasets based on their sizes and proportions assigned to the benchmarks they come from (more details in Section \ref{sec:ablation}).


\subsection{Packing and Document Attention}
\label{subsec:packing}
In order to utilize the maximum sequence length for computational efficiency, we pack multiple examples (source and target) together as a sequence of 2048 tokens \citep{raffel2020t5}, separated by \texttt{<eos>} tokens. One consequence of packing is that the tokens belonging to one example can attend to tokens from previously packed examples in the same sequence. To mitigate this, we use document attention masking i.e. we modify the token attention mask in causal LMs to attend only to the tokens that are part of the same example, rather than all the previous tokens in the sequence. This changes the attention mask from a triangular to a block triangular mask and improves both stability and performance in our experiments. 


\subsection{Fine-tuning Hyperparameters}
\label{subsec:fthp}
We fine-tune all 30B models on 64 40GB A100s, and 175B models on 128 40GB A100s. Following OPT, we use Fully Sharded Data Parallel \citep{artetxe2021efficient} and the Megatron-LM Tensor Parallelism \citep{shoeybi2019megatron}. We inherit most model hyper-parameters for each model scale following OPT. We pack our training examples into sequences of length $2048$, left-truncating examples that overflow. 
We use Adam \citep{kingma2014adam} with 32-bit state with $(\beta_1, \beta_2) = (0.9, 0.95)$, linearly warming up the learning rate for $60$ steps to the maximum, followed by linearly decaying it to 0. We conduct preliminary experiments to select learning rates from $\{1e^{-5}, 3e^{-5}, 5e^{-5}, 6e^{-5}\}$ and per-GPU batch sizes from \{2, 4, 8\} using our validation split from \S\ref{sec:scaled-multi-task-benchmark}. The resulting hyperparameters are listed in Table \ref{tab:hyp}. We use a dropout of 0.1 (including embedding dropout) and clip gradient norms to 1.0, and use dynamic loss scaling to prevent underflows \citep{micikevicius2018mixed}. During fine-tuning, our models saw approximately 2 billion tokens, which is only 0.6\% of the pre-training budget of OPT (Table \ref{tab:hyp}).

\begin{table}[h]
\centering{
\scalebox{0.85}{
\begin{tabular}{lccccccc}
\toprule
Model & \# Gpus & Batch Size & Learning Rate & Steps & Warm-up Steps & FT Time (h) & \# Tokens \\
\midrule
\Ours 30B & 64 &  256 & 5e-05 & 4000 & 60 & 19 & 2B \\
\Ours 175B & 128 & 128 & 5e-05 & 8000 & 60 & 72 & 2B \\
\bottomrule
\end{tabular}
}
\caption{Fine-tuning parameters for all \Ours models, including the fine-tuning times and the number of fine-tuning tokens.}
\label{tab:hyp}
}

\end{table}

\section{What Matters for Instruction Fine-tuning?}
\label{sec:ablation}

Recent works have explored a number of instruction fine-tuning techniques to optimize the performance of the resulting model on specific kinds of downstream tasks, and also to improve their robustness against variations in prompts, instruction styles and prompting setups. Using an OPT 30B model with the basic hyper-parameter settings chosen in \S\ref{subsec:fthp},
we run experiments to characterize the effects of dataset proportions, number of tasks and diversity, using pre-training, dialogue, and reasoning datasets, and training using demonstrations, on instruction-tuning with respect to our three levels of model generalization: \emph{fully held-out}, \emph{partially held-out} and \emph{fully supervised}. We aggregate performance along several dimensions such as clusters, and benchmarks to determine the best settings. 

\subsection{Experimental Setup}

The goal of our experimental setup is first, to characterize the effects of a multitude of factors related to the fine-tuning process, on instruction-tuning performance, and second, to use these findings to effectively instruction-tune OPT models. The factors that we experiment with are 1) the composition of the fine-tuning dataset mixture, 2) the number and diversity of the tasks used for fine-tuning, 3) using additional datasets relating to pre-training, reasoning and dialogue as part of the fine-tuning mix, and 4) different ways of fine-tuning with demonstrations. 

\paragraph{Prompt construction details.} 
To compile our train data, we merged all prompt data for a task with $N$ examples and randomly take $N$ prompts from the pool such that the training task distribution is kept the same regardless of how many prompts are given for the tasks. We merged the prompts for each task in a similar manner in our validation set, and randomly sample a maximum of 250 prompts per task to report the validation results. For our test tasks, we keep all prompt variations and all examples.

\paragraph{Generalization levels.} Starting with a baseline instruction-tuned model, we independently characterize the effect of each factor, by tuning models with several variations of that factor and evaluating the models on the tasks from our validation split from Section \ref{sec:scaled-multi-task-benchmark}, separated into three generalization levels: a) tasks from clusters not included in training (Fully Held-out), b) tasks unseen during training but from seen clusters (Partially Supervised), and c) tasks seen during training (Fully Supervised). An instruction-tuning setting is desirable if it improves performance on fully held-out and partially supervised tasks without sacrificing performance on fully supervised tasks. We use average performance across all three generalization levels on both 0-shot and 5-shot settings on the validation/test sets of the tasks in the validation split to determine the best settings for each factor. 

\paragraph{Decoding.} 
\label{sec:decoding}
Our evaluation data comprises tasks with answer candidates (of which one is correct), as well as tasks with multiple gold reference sequences. For the former set of tasks, we use rank classification similar to \cite{gpt3}, where we score each candidate based on their likelihood and output the highest-scoring candidate as the answer. This candidate is used to compute accuracy on the task. For tasks without candidates, we perform greedy decoding until an \texttt{<eos>} token is predicted or a maximum of \textit{N=256} tokens are generated. Based on the generated sequence and the references, we then compute either Exact-match or Rouge-L F1 scores.


\paragraph{Model selection.} For all experiments, we first aggregate results separately for 0-shot and 5-shot across task subtypes. For example, pro and anti versions of type 1 and type 2 Winobias \citep{zhao2018gender} tasks from PromptSource, and all 57 subtasks of MMLU \citep{hendrycks2020measuring}, would be aggregated to get per task performance. If the same task exists across multiple benchmarks, we then average performance across benchmarks as well. We then compute 0-shot and 5-shot averages of all tasks within a category (or benchmark depending on the experiment), and finally, compute a combined average of all 0 and 5-shot scores of each category (or benchmark), which we use for model selection. 

We tune each model for 4000 steps and evaluate on our validation split on both 0-shot and 5-shot settings, using 250 examples from each task for compute-efficiency. As described in Section \ref{sec:scaled-multi-task-benchmark}, our validation splits for each task include a mix of multiple prompts for FLAN and PromptSource. All but four validation tasks are generation-style tasks (where we report Rouge-L F1). We compute accuracy based on scoring for the remaining tasks and aggregate them together with Rouge-L for presentation purposes. We refer to Table~\ref{tab:validation_tasks} in the Appendix for full details about the tasks in our validation split. 

\subsection{Effects of varying task mixing-rate maximum}
\label{subsec:eps}

Prior work \citep{raffel2020t5,wei2021flan} typically uses example-proportional sampling and builds batches by sampling from datasets proportional to their sizes, while enforcing a maximum size parameter (EPS) to prevent large datasets from overwhelming the batch. To understand how this maximum mixing rate (EPS) affects performance across the different generalization levels, we perform experiments with EPS $\in \{128, 256, 512, 1024, 2048, 4096, 8192, 16384, 10^6\}$ and report results in Table \ref{tab:eps}. An EPS of 512 causes 97\% datasets to hit their maximum, while an EPS of 8192 causes 16\% datasets to hit their maximum. We also experiment without using EPS i.e. EPS=100K. 

\begin{table}[h]
\scalebox{0.68}{
\begin{tabular}{l|cccc|ccccc|ccc}
\toprule
& \multicolumn{4}{c}{Fully Held Out} & \multicolumn{5}{c}{Partially Supervised} & \multicolumn{3}{c}{Fully Supervised} \\
& \makecell{Cause \\ Effect} & \makecell{Gram. \\ Corr.} & \makecell{Stereo. \\ Det.} & \makecell{Word \\ Ana.} & Reas. & MMLU & QA & Summ. & \makecell{Toxic \\ Det.} & \makecell{Dial \\ ogue.} & QA & Summ. \\
\midrule
\textbf{$2^7$} & 61.4/62.0 & 86.2/87.5 & 59.1/82.5 & 12.1/59.1 & 2.9/22.4 & 42.5/35.6 & 67.5/59.7 & 21.0 & 61.7/66.3 & 16.8/17.5 & 86.9/83.3 & 30.7 \\
\textbf{$2^8$}  & 59.3/60.7 & 86.5/87.8 & 60.2/83.4 & 13.0/57.1 & 2.6/19.1 & 41.5/36.0 & 64.8/59.9 & 20.5 & 61.7/69.5 & 16.4/16.8 & 86.2/83.7 & 31.0 \\
\textbf{$2^9$} & 59.6/61.3 & 86.4/87.9 & 55.2/82.8 & 12.9/58.5 & 2.6/24.7 & 40.2/38.1 & 65.3/57.4 & 20.2 & 59.8/66.2 & 17.1/16.6 & 85.7/82.6 & 31.2 \\
\textbf{$2^{10}$} & 64.5/60.3 & 86.0/87.6 & 47.9/82.3 & 14.1/56.8 & 2.7/23.6 & 39.0/35.9 & 66.9/61.6 & 20.5 & 60.8/66.4 & 17.7/16.0 & 86.1/85.2 & 31.0 \\
\textbf{$2^{11}$} & 64.4/62.7 & 85.9/87.7 & 50.4/82.2 & 11.7/54.5 & 2.7/22.0 & 40.1/35.7 & 67.4/58.6 & 19.9 & 60.1/65.6 & 17.2/16.8 & 87.3/84.6 & 31.4 \\
\textbf{$2^{12}$} & 63.5/62.5 & 86.1/87.5 & 58.9/82.3 & 17.2/57.8 & 2.6/20.4 & 41.5/37.0 & 69.3/59.0 & 18.1 & 60.0/70.0 & 16.1/15.8 & 87.6/83.5 & 31.3 \\
\textbf{$2^{13}$} & 63.3/61.2 & 85.6/87.9 & 48.2/81.3 & 13.2/56.8 & 2.6/25.6 & 38.3/35.9 & 69.4/57.7 & 19.6 & 59.4/68.2 & 16.4/15.6 & 86.2/84.5 & 32.3 \\
\textbf{$2^{14}$} & 60.2/61.3 & 86.0/88.0 & 57.3/82.5 & 15.1/52.6 & 2.6/20.3 & 41.8/36.1 & 70.5/61.1 & 19.8 & 58.6/64.0 & 16.9/14.7 & 86.1/84.4 & 32.0 \\
\textbf{$10^6$} & 59.2/62.2 & 86.4/86.9 & 57.3/80.8 & 8.8/53.7 & 2.6/22.0 & 39.2/34.2 & 67.6/59.5 & 19.8 & 58.2/68.1 & 15.2/15.8 & 84.6/81.6 & 31.7 \\
\bottomrule
\end{tabular}
}
\caption{Performance variation across different task categories with different maximum mixing rates (EPS), for each generalization level on \Ours 30B, after 4000 steps. Results are in the format of 0-shot/5-shot. We use only 0-shot performance for summarization tasks. Most tasks are generation tasks, for which we report Rouge-L. We report accuracy for MMLU. Some tasks in the Cause Effect Cluster also use accuracy, which is averaged with Rouge-L for presentation purposes. We select models based on their average performance aggregated per category, benchmark and shot.}
\label{tab:eps}
\end{table}

Overall, we find that while EPS is important to instruction-tuning i.e. on average all models that use EPS outperform the model without it, after a certain threshold i.e. less than 4096 in our case, there is minimal variation in performance across all generalization levels. While based on the highest average performance, we choose 4096 (also corresponds to 50\% of the dataset lengths being capped) for our other experiments and the final \Ours models, we find that all values below 4096 also perform quite well, with EPS=128 closely matching 4096. Also note that changing EPS implicitly changes the proportion of fine-tuning data from each benchmark, which we control for explicitly in the next Section.

\subsection{Effects of varying benchmark proportions}
\label{subsec:prop}

In Section \ref{sec:scaled-multi-task-benchmark}, we describe the multiple tasks and prompt repositories \citep{sanh2021t0,wang2022niv2,wei2021flan,ye2021crossfit,aribandi2021ext5} that we unify to massively scale the number of tasks used for instruction-tuning. However, using multiple benchmarks for training, together with only example-proportional sampling, results in benchmarks with more tasks overwhelming the batch composition. For example, in our benchmark, 71\% of training examples would come from \natinsShort, with 18\% from PromptSource, and only 5\% from FLAN. Since each benchmark is associated with a specific task format, this can bias the resulting model towards certain input-output formats. We vary the proportions of different benchmarks to evaluate their effect on downstream task performance on our three generalization levels and present results in Table \ref{tab:prop}. For this experiment, we compare models based on their aggregate performance on each benchmark instead of task category, since we would like to choose the parameters that perform well on a maximum number of benchmarks.

\begin{table}[h]
\scalebox{0.70}{
\begin{tabular}{l|ccc|ccccc|cc}
\toprule
 & \multicolumn{3}{c}{Fully Held-Out} & \multicolumn{5}{c}{Partially Supervised} & \multicolumn{2}{c}{Fully Supervised} \\
\makecell{Benchmark Props. \\ Crossfit/Exmix/Flan\\/NIV2/PS/T5/U-SKG} & FLAN & NIV2 & PromptS & Reas. & FLAN & MMLU & NIV2 & PromptS & FLAN & PromptS \\
\midrule
\textbf{2/1/ 5/71/18/1/2} & 79.2/74.4 & 52.4/61.8 & 75.2/79.7 & 2.7/23.4 & 17.8 & 37.3/35.3 & 69.3/61.4 & 54.3/62.0 & 85.8/82.9 & 43.1/49.1 \\
\textbf{2/1/35/25/34/1/2} & 86.8/80.8 & 53.0/62.5 & 72.0/83.7 & 2.6/20.3 & 17.7 & 34.5/30.8 & 62.2/53.5 & 57.6/66.2 & 85.9/81.7 & 44.3/48.3 \\
\textbf{3/3/35/25/25/7/2} & 81.2/83.2 & 52.5/61.1 & 79.7/83.5 & 2.7/19.8 & 20.0 & 36.7/29.8 & 60.9/54.1 & 57.1/56.8 & 86.8/84.1 & 43.4/48.3 \\
\textbf{2/1/27/40/27/1/2} & 86.8/81.2 & 52.4/63.2 & 77.9/83.3 & 2.6/21.3 & 20.2 & 36.3/30.3 & 67.3/60.4 & 57.8/61.7 & 86.4/81.6 & 43.2/48.8 \\
\textbf{3/3/25/25/35/7/2} & 91.2/80.4 & 51.1/62.2 & 75.6/83.4 & 2.6/18.4 & 21.4 & 37.5/33.7 & 59.7/51.5 & 57.4/66.9 & 83.6/83.7 & 44.3/48.9 \\
\textbf{4/2/35/25/30/2/2} & 88.0/76.8 & 51.5/61.3 & 75.1/82.7 & 3.0/16.8 & 20.0 & 37.1/30.7 & 65.6/58.0 & 60.4/61.5 & 85.4/81.5 & 43.2/49.9 \\
\textbf{4/2/20/25/45/2/2} & 88.8/83.6 & 54.5/62.2 & 73.5/85.0 & 2.5/13.1 & 19.8 & 38.2/33.2 & 63.0/57.5 & 56.1/61.8 & 86.1/84.2 & 43.0/48.7 \\
\textbf{2/1/35/25/30/5/2} & 86.0/83.2 & 51.1/61.6 & 74.0/82.8 & 2.6/17.1 & 20.8 & 36.9/31.9 & 63.5/62.4 & 53.1/63.7 & 86.2/81.6 & 43.5/49.7 \\
\textbf{7/1/35/25/28/2/2} & 85.6/81.2 & 51.0/61.6 & 78.0/82.1 & 2.6/19.9 & 20.0 & 36.3/31.9 & 65.1/60.6 & 59.6/63.1 & 85.0/84.0 & 43.2/49.3 \\
\textbf{0/0/35/30/35/0/0} & 86.0/79.2 & 52.3/62.6 & 71.8/84.2 & 2.6/15.3 & 19.3 & 36.6/28.6 & 60.8/54.8 & 56.9/62.3 & 85.2/80.2 & 43.6/47.8 \\
\bottomrule
\end{tabular}
}
\caption{Per-benchmark performance variation at each generalization level with varying benchmark proportions; The first row represents the original proportions in the \Ours benchmark. Results are in the format of 0-shot/5-shot. We use only 0-shot performance for Summarization tasks. Most tasks are generation tasks, for which we report Rouge-L. We report accuracy for MMLU. Four tasks in the Cause Effect Cluster also use accuracy, which is averaged with Rouge-L for presentation purposes. We select models based on their average performance aggregated per benchmark and shot.}
\label{tab:prop}
\end{table}

First, we look at performance improvements within the same benchmark where the proportions were changed. As we increase the proportion of FLAN from 5\% to 25\%, its performance improves significantly on both the fully-held out and the partially held-out generalization levels, with no notable improvement on the fully-supervised tasks. \natinsShort shows a similar trend on partially-supervised tasks, but surprisingly, not so much on fully held-out tasks. It is possible that the very specific input-output format of \natinsShort makes it such that changing proportions of unrelated clusters provides no benefit to its fully held-out clusters. PromptSource is relatively unchanged on fully supervised clusters and partially supervised clusters, possibly owing to reaching performance saturation with even an 18\% proportion. However, it benefits with more proportion on the fully-held out clusters. 

Secondly, we also observe benchmarks complementing each other. For example, the highest accuracy on fully held-out FLAN i.e. 88.8/83.6\%, is achieved, not with the highest proportion of FLAN, but with improving the proportions of PromptSource and Crossfit. Similarly, the highest generation performance on fully-held out PromptSource of 79.7/83.5\% is achieved with 25\% PS, and not with 45\% PS proportions. We also observe certain tradeoffs, for example, the best proportions for FLAN and PromptSource result in a sharp drop in performance on reasoning datasets, and vice versa. Finally, setting Crossfit, Exmix, T5 and Unified-SKG proportions to 0 results in the worst model, demonstrating the benefits of using a diverse set of benchmarks for instruction-tuning. Based on average performance across benchmarks, ``2/1/27/40/27/1/2", ``7/1/35/25/28/2/2" and ``4/2/20/25/45/2/2" performed the best and we choose the last one as the proportion for our final \Ours models. Despite our choice, instruction-tuned models with different end-goals (for example, producing reasoning chains) would benefit from choosing differently. We also explore methods to improve performance on reasoning datasets in Section \ref{sec:cot}.

\begin{figure}
    \centering
    \includegraphics[width=0.95\linewidth]{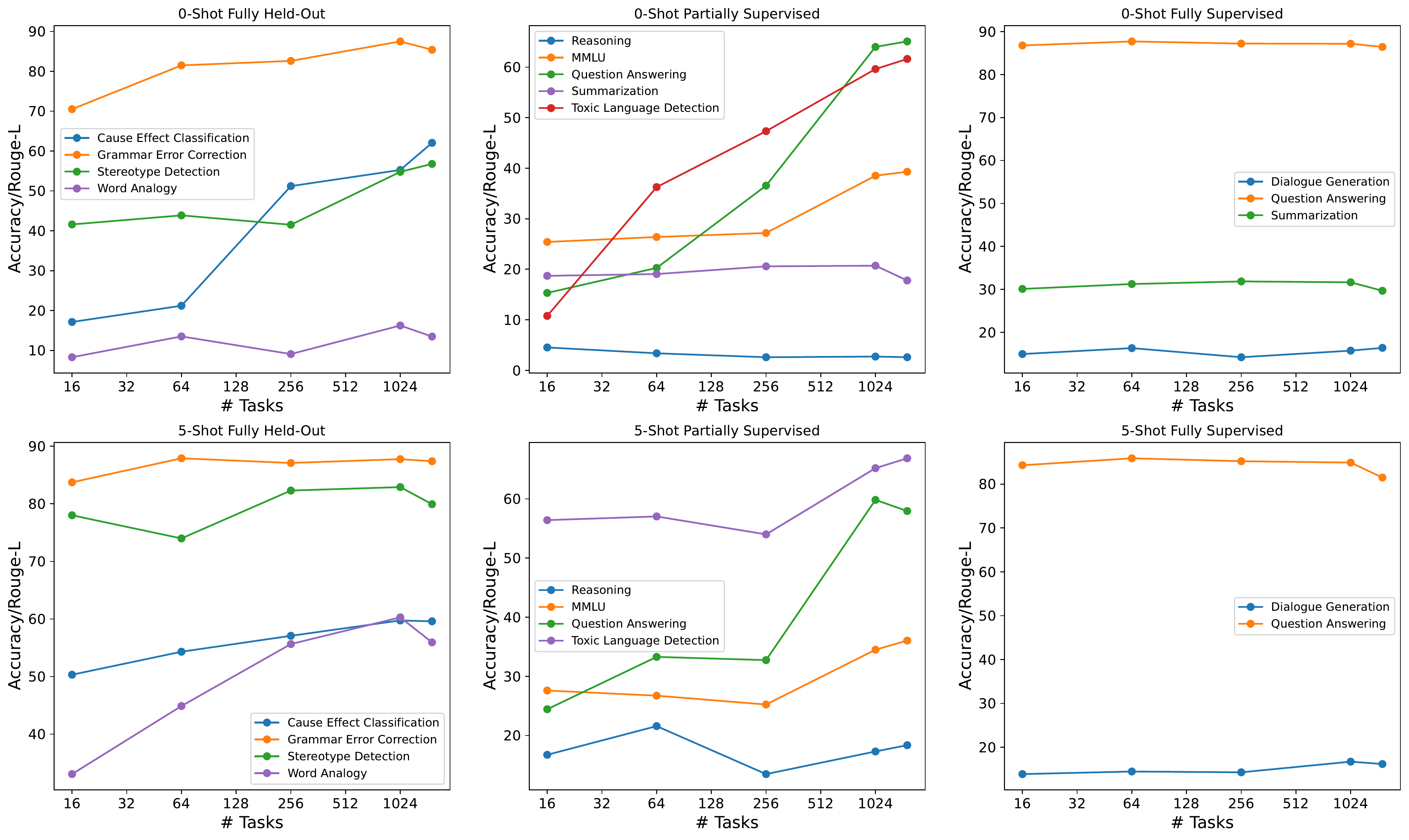}
    \caption{Effect of scaling the number of training tasks on each generalization level for \Ours 30B under both 0-shot and 5-shot settings, aggregated by task category. 
    }
    \label{fig:tasks_abalation}
\end{figure}

\subsection{Effects of Scaling Tasks or Categories}
\label{subsec:scaling-tasks-clusters}

Previous work has shown that scaling the number of training tasks or clusters improves the overall performance of the model on the fully held-out generalization setting~\citep{wei2021flan,wang2022niv2}. We study effects along similar axes but with more generalization settings such as fully held-out, partially supervised, and fully supervised tasks/categories. We use cluster/category interchangeably in this section. For the task scaling study, we randomly sample 16, 64, 256, and 1024 sets of tasks such that smaller sets are subset of bigger sets, and fully supervised tasks are always selected. Figure~\ref{fig:tasks_abalation} (full results in Appendix Table~\ref{tab:scaling-tasks}) presents these task scaling studies on the three generalization levels, aggregated at the cluster-level for both 0 and 5-shot performance. 

We observe that both fully held-out and partially supervised tasks get the most improvements with the increase in the number of training tasks. Interestingly, fully supervised tasks' performance remains unchanged even when more relevant tasks are seen from the fully supervised tasks' clusters, as we increase the training tasks. 
In the fully held-out setting, \emph{Cause Effect Classification} and \emph{Word Analogy} clusters see the biggest improvements in zero-shot and few-shot, respectively. On the partially supervised, \emph{Question Answering} and \emph{Toxic Language Detection} clusters see the biggest improvements on both zero-shot and few-shot. 

\begin{wrapfigure}{r}{0.45\textwidth}
  \begin{center}
    \includegraphics[width=0.45 \textwidth]{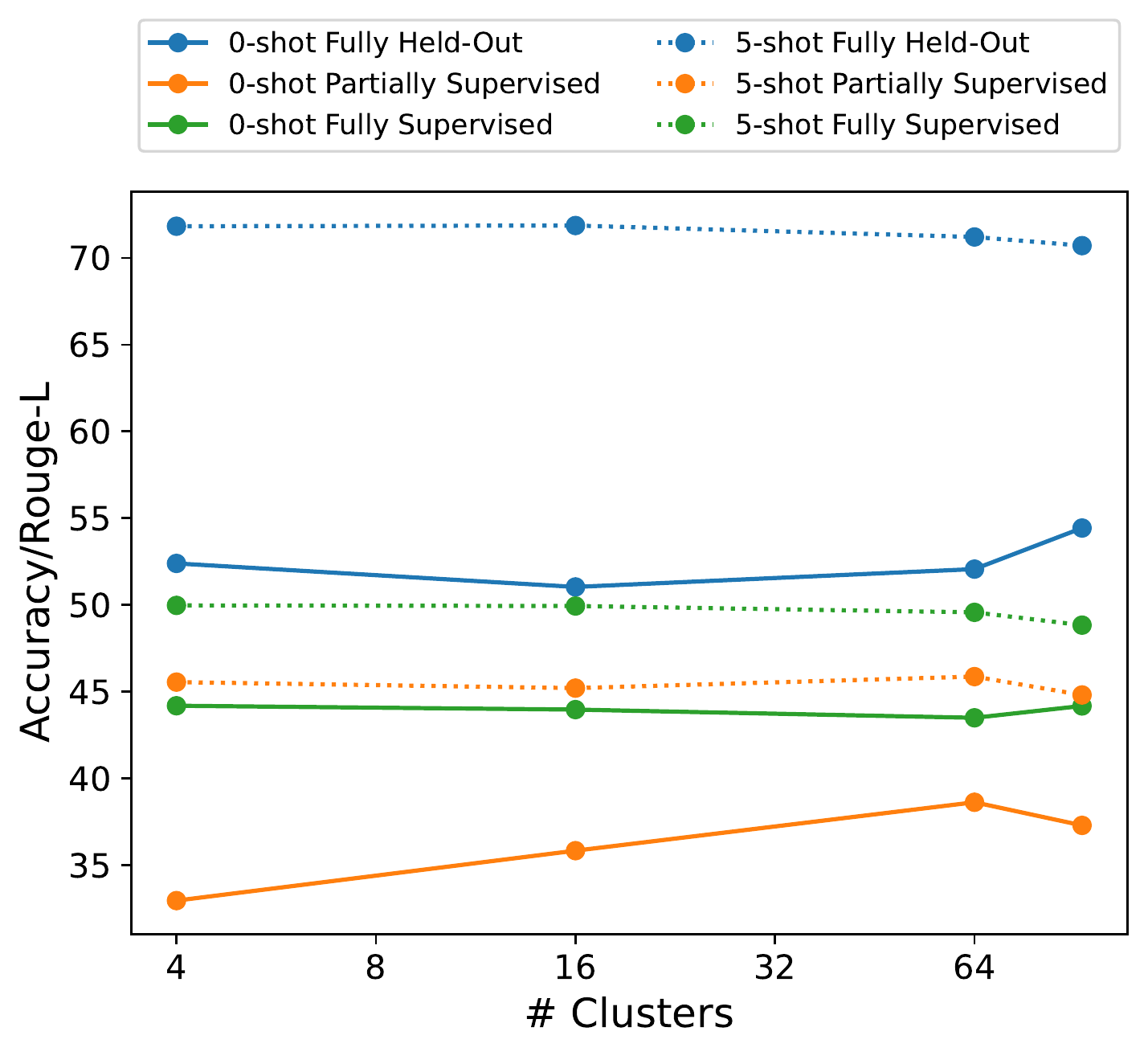}
  \end{center}
  \caption{Effect of scaling the number of training categories on each generalization level for \Ours 30B under both 0-shot and 5-shot settings.}
  \label{fig:num_clusters}
\end{wrapfigure}

For the cluster scaling study, we order the clusters based on the decreasing order of the number of tasks present in each cluster and select the first 4, 16, 64, and 93 (i.e., all) clusters. Additionally, we make sure that Question Answering, Summarization, and Dialogue Generation clusters are always represented since our fully supervised validation tasks belong to these three clusters. Figure~\ref{fig:num_clusters} (full results in Appendix Table~\ref{tab:scaling-clusters}) presents the corresponding results on all three generalization levels for both zero-shot and few-shot settings. We observe that as we increase the training clusters, the performance on fully supervised tasks either stay the same or slightly drop in the few-shot setting. On the fully held-out and partially supervised levels, the results on the zero-shot settings improve an increase in the number of clusters and the results are a bit mixed for the few-shot setting, but overall they tend to decrease with cluster scaling. Note that the first 4 clusters already cover 673 tasks (clusters belonging to the fully supervised setting have a lot of tasks). Hence, the model starts with strong performance, which might lead to the mixed results that we observe. Based on these experiments, we use all tasks and clusters to train our final \Ours models.

\begin{table*}[t]
    \centering
    \resizebox{0.88\linewidth}{!}{
    \begin{tabular}{lp{0.85\textwidth}}
        \toprule
         Dataset & Example (Input Prompt and \textit{Output}) \\
         \midrule
         Pre-training & \textit{You could make it a full group party with the kids and wives. Don't make it just about books. So have A movie night My parents made a movie group they go out to dinner then see a movie then dicuss it. You could play card games.  Watch some comedy.  Ask the members.  Do a music night when one of you has to bring a selection of their fav music.} \\
         \midrule
         Reasoning & Answer the following question by reasoning step by step. \newline How do most people feel about a person they love? \newline popularity, know all, own house, care about, flu Output: \textit{we care about people we love. The answer is care about} \\
         \midrule
         Dialogue &  \textit{I love cats and have five of them.\newline Cats are nice. How old are you?\newline Old enough to work in the construction field. You?\newline I am 68, been retired for a few years now.\newline Great. What did you work and retire from?\newline I was a tailor.} \\
        \bottomrule
    \end{tabular}
    }
    \caption{Examples from the pre-training, reasoning, and dialogue datasets. For pre-training and dialogue data, the source is empty and the entire text sequence is considered as the target.
    }
    \label{tab:ablation-data-examples}
\end{table*}

\subsection{Effects of Pre-training during Instruction-Tuning}
\label{subsec:pretrain-data}

\begin{figure}
    \centering
    \includegraphics[width=\linewidth]{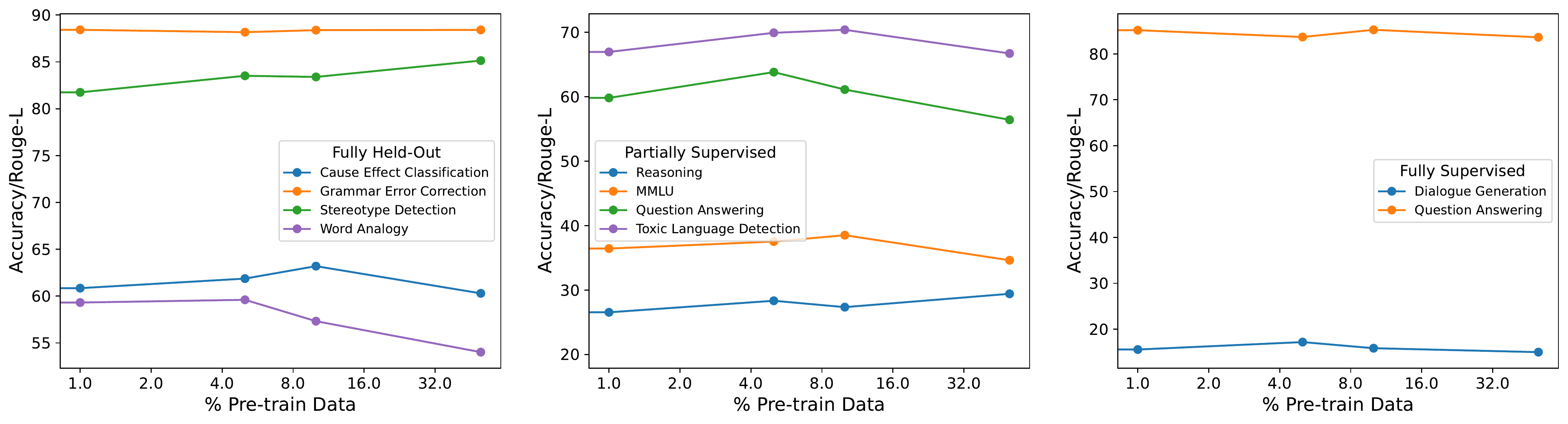}
    \caption{Effect of performing pre-training updates on entire sequences, together with instruction-tuning on each generalization level for \Ours 30B in the 5-shot setting, aggregated by task category. The x-axis represents the \% of pre-training updates performed w.r.t the total number of updates.}
    \label{fig:pretrain-data}
\end{figure}

We observe that using pre-training style updates on entire sequences during fine-tuning can make training more stable, so we explore the performance effects of using pre-training data on our three generalization levels. Table \ref{tab:ablation-data-examples} shows an example used in the pre-training style updates.
Following \cite{shuster2022blenderbot}, we use the last shard of the corpus used to train OPT~\citep{zhang2022opt} as our pre-training data for fine-tuning, since it is seen only once during the pre-training stage of OPT. We experiment with adding pre-training data by proportion in the increasing amounts of 1\%, 5\%, 10\%, and 50\%, and present results for the 5-shot setting, aggregated by task category, in Figure~\ref{fig:pretrain-data} (full 0 and 5-shot results in Appendix Table~\ref{tab:pretrain-data}).  

Overall, for the fully held-out and partially supervised generalization levels, we observe that the model improves while adding pre-training data for up to 10\% and then starts deteriorating after that. We also observe that using more pre-training data leads to better Rouge-L F1 scores but lower accuracy scores, partly owing to the influence of pre-training data on the remaining proportions of generation vs. classification tasks. Based on the average scores across generalization levels (see Appendix Table~\ref{tab:pretrain-data}), we choose to include 5\% pre-training data in instruction-tuning our \Ours models.

\subsection{Effects of Adding Reasoning Datasets}
\label{sec:cot}
Recent work \citep{wei2022cot,kojima2022large} has illustrated improvements in the performance of LLMs on reasoning tasks, when prompted to generate a reasoning chain in natural language before generating the answer. Based on these findings, we attempt to explicitly fine-tune LLMs to perform reasoning by compiling a set of 14 reasoning datasets (see Appendix \ref{subsec:reasoning-benchmark} for a list of these datasets), where the output includes a rationale before the answer and by including these datasets during instruction-tuning. This set includes the 9 datasets used by \cite{chung2022flanpalm} in their CoT category as well as some additional datasets. Each dataset has a single prompt that uses an instruction, that explicitly asks the model to generate a reasoning chain \citep{kojima2022large}, followed by examples in the few-shot setting that illustrate how the reasoning chain should be produced before the answer. We show an example with such a prompt in Table \ref{tab:ablation-data-examples}. Using benchmark proportions of ``2/1/27/40/27/1/2" as a baseline (see Section \ref{subsec:prop}), we experiment with adding 1\%, 2\%, and 4\% proportions of reasoning data (by reducing the proportion of the highest proportion benchmark i.e. \natinsShort), and present results for the 5-shot setting in Figure~\ref{fig:cot} (full 0 and 5-shot results in Appendix Table \ref{tab:cot}) by generalization level and task category. 

\begin{figure}
    \centering
    \includegraphics[width=\linewidth]{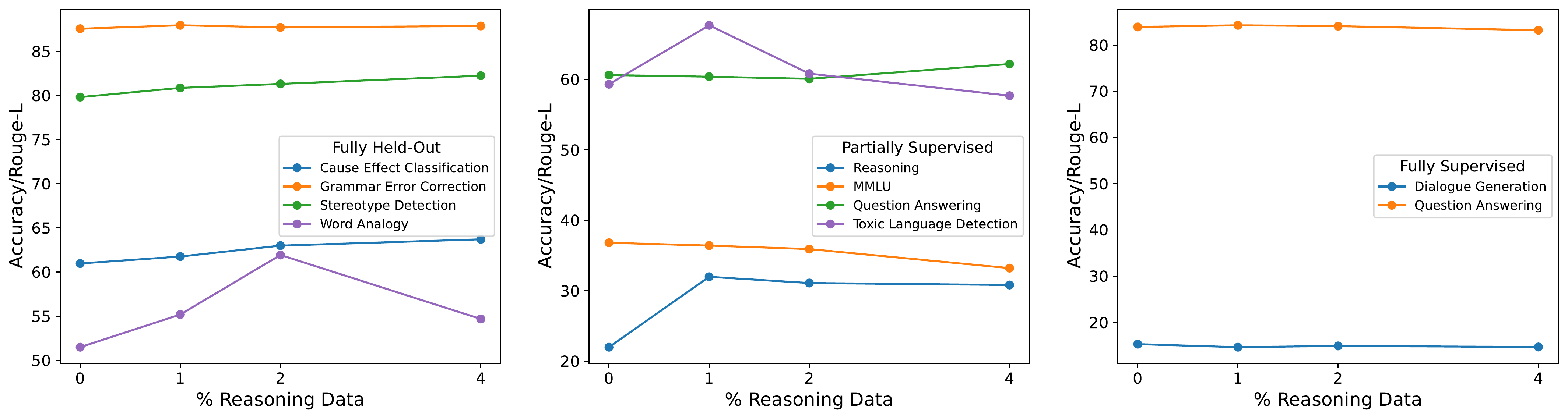}
    \caption{Effect of fine-tuning using reasoning datasets on each generalization level for \Ours 30B in a 5-shot setting, aggregated by task category. We experiment with adding 1\%, 2\% and 4\% reasoning datasets by proportion. Note that the baseline for this experiment is based on a different proportion than other experiments.}
    \label{fig:cot}
\end{figure}

We see a substantial performance improvement on the 2/14 held-out validation reasoning tasks (Rouge-L from 12.2\% to 31.6\%) when we instruction-tune with reasoning datasets, but alongside, we also see improvements on other held-out task categories such as Cause-Effect, Stereotype Detection, Toxicity Detection, and Word Analogy. Furthermore, adding 1\% reasoning data results in the largest gains overall, beyond which, the gains start to reduce on MMLU, Cause-Effect Accuracy, Toxicity, and Dialogue (averaged over 0 and 5-shot). On the other hand, the Summarization cluster (only 0-shot, see Appendix) continues to benefit from higher proportions of reasoning data. Based on average performance across categories and generalization levels, we use 1\% reasoning data for our final \Ours models.

\subsection{Effects of Adding Dialogue Datasets}
\label{sec:ablation_dialogue}
We experiment with adding dialogues as auxiliary fine-tuning data to test if it can improve the LM's ability to respond to directional input and understand referential expressions. Another goal is to evaluate if this approach can induce chat-bot behaviors~\citep{shuster2022blenderbot} and make the resulting models more conversational. Using a subset of dialogue datasets\footnote{Appendix~\ref{subset:dialogue-datasets} list the dialogue datasets we used in this experiments.} used for training BlenderBot 3~\citep{shuster2022blenderbot}, we process the dialogues into sequences of turns separated by a single newline token (see Table \ref{tab:ablation-data-examples} for an example). The data consists of  320,543 unique dialogues and we fine-tune the model to predict the entire dialogue sequence. We set the proportion of the included dialogue data to be 0.5\% and present 0 and 5-shot results by task category and generalization level on our validation split in  Table~\ref{tab:add_dialogue}.

\begin{table}[h]
\setlength{\tabcolsep}{2.2pt}
\scalebox{0.69}{
\begin{tabular}{l|cccc|ccccc|ccc|c}
\toprule
& \multicolumn{4}{c}{Fully Held Out} & \multicolumn{5}{c}{Partially Supervised} & \multicolumn{3}{c}{Fully Supervised} & \multirow{2}{*}{Average} \\
\makecell{EPS} & \makecell{Cause \\ Effect} & \makecell{Gram. \\ Corr.} & \makecell{Stereo. \\ Det.} & \makecell{Word \\ Ana.} & Reas. & MMLU & QA & Summ. & \makecell{Toxic \\ Det.
} & \makecell{Dial \\ ogue.} & QA & Summ. \\
\midrule
Baseline & 63.5/62.5 & 86.1/87.5 & 58.9/82.3 & 17.2/57.8 & 2.6/20.4 & 41.5/37.0 & 69.3/58.9 & 18.1 & 60.0/70.0 & 16.1/15.8 & 87.6/83.5 & 31.3 & \highest{46.0/57.6} \\
+ 0.5\% BB3 & 61.7/62.2 & 86.1/87.4 & \underline{51.9}/83.4 & \underline{10.4}/57.5 & 2.6/22.2 & 40.2/35.4 & 68.9/62.5 & 20.6 & 61.9/\underline{65.4} & 16.1/15.2 & 86.4/83.7 & 31.1 & 44.8/57.5 \\
\bottomrule
\end{tabular}}
\caption{Effect of fine-tuning with 0.5\% dialogue data on each generalization level for \Ours 30B after 4000 steps, aggregated by task category. Results are presented in the format 0-shot/5-shot. Most categories use Rouge-L F1, MMLU uses accuracy. Some Cause-Effect tasks use accuracy, which is averaged with Rouge-L F1 for presentation purposes.}
\label{tab:add_dialogue}
\end{table}

We observe that adding even just 0.5\% of the aforementioned dialogue data lowers 0-shot performance while 5-shot performance remains 
 unchanged. Specifically, 0-shot performance suffers mainly on stereotype detection and word analogy. On examining model predictions on these categories, we found that they are primarily generation tasks whose 
 references are either a single word or a short piece of text with a specific format (for example, a pair of phrases from the original input that refer to each other). Training with BB3 data weakened the model's ability to conform to the required format.\footnote{On one hand this behavior demonstrates a weakened instruction-following ability for the underlying model. On the other hand, it exposes a caveat in measuring model performance on tasks with instructions -- model performance on a specific task category is often the result of multiple factors and underperforming on a particular task category may not offer a useful atomic diagnosis. As in our case, we found the model to perform worse on stereotype detection tasks because it cannot parse the required output format, not because it is a more biased model.} It also significantly lowered the 5-shot performance of toxicity detection. An error analysis revealed a similar problem i.e. the model tends to perform worse on tasks that require generating a special set of decision words rather than simply generating ``yes'' or ``no''. 
Owing to severe model degeneration on these tasks, we do not add dialogue data while tuning \Ours.

\subsection{Effects of Meta-Training for In-Context Learning}
\label{subsec:meta-icl}

Recent work has shown that fine-tuning language models with demonstration examples in the instructions improves their ability to learn from the examples in context~\citep{min2021metaicl,wang2022niv2,chung2022flanpalm}. Both~\cite{min2021metaicl} and~\cite{wang2022niv2} experimented with the setup where a constant number of $k$ demonstration examples are added to each training example. The models are evaluated with the same number of $k$ demonstration examples during inference. ~\cite{chung2022flanpalm} used a mixture of data with and without exemplars. However, the proportion of each type of data used and how many exemplars were included are not clear. 

We attempt to train models that are better in-context few-shot learners, and also robust to the number of demonstration examples used during inference time.\footnote{In preliminary experiments, we found models trained with $k$ exemplars tend to perform worse when a different number of exemplars is used during inference time. Table~\ref{tab:niv2} shows a similar effect on the \tkinstruct models.} We experiment with a simple way of creating training examples that include varying numbers of demonstration examples. For each example $e$, we sample $k$ from a distribution $\mathcal{D}$ with cap\footnote{For any $k>K$, we set $k=K$.} $K$, and  randomly select $k$ other examples $E_d=\{e_1,\hdots,e_k\}, e_i \neq e$, from the train set, if $k > 0$. We add $E_d$ as the demonstration examples in $e$'s prompt, where the examples are separated by a special token \texttt{[SEP]}. For benchmarks with task-level instructions such as \natins, we place the demonstration examples before $e$ and after the instruction field; for benchmarks with instance-level instructions such as \flan and \promptsource, we place the demonstration examples before $e$. 

Because the demonstration examples significantly increase the prompt lengths, including too many few-shot training examples often leads to worse performance and reduced learning stability, owing to sparsity in the loss and lower batch diversity. As a result, we choose $\mathcal{D}$ to be the Zipf distribution\footnote{\url{https://docs.scipy.org/doc/scipy/reference/generated/scipy.stats.zipf.html}}, which can be heavily tilted towards $k=0$. We train MetaICL models with different $\mathcal{D}$'s by adjusting the shape parameter $a$ of the Zipf distribution. When $a=4$, $92.5\%$ of the examples are zero-shot examples; and when $a=2$, $67.1\%$ of the examples are zero-shot examples. We set $K=5$ and use three consecutive newline tokens as \texttt{[SEP]} following~\cite{min2021metaicl}. 

\begin{figure}[t!]
    \centering
    \includegraphics[width=\linewidth]{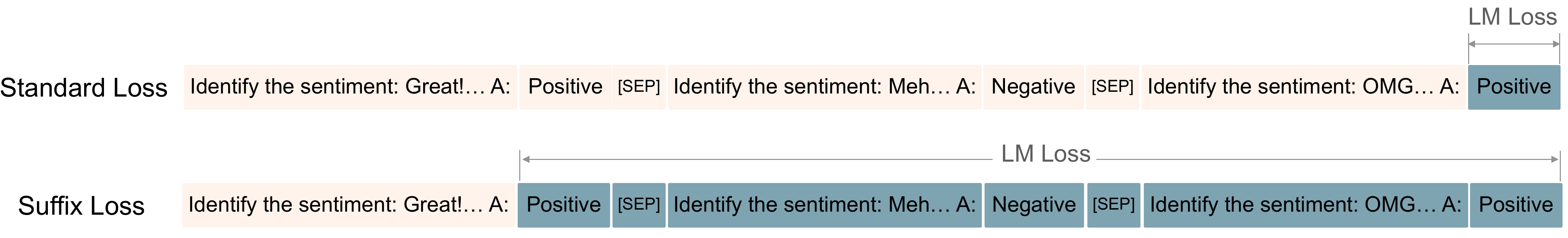}
    \caption{We experiment with two types of training losses for MetaICL: the generation loss over the label of the target example as proposed by~\cite{min2021metaicl}, and the generation loss over the label of the first demonstration example and the complete sequences of the following examples.}
    \label{fig:metaicl_loss}
\end{figure}

\paragraph{MetaICL with suffix loss.} To further address the loss sparsity problem, we also experiment with a variation of the original MetaICL loss, illustrated in Figure~\ref{fig:metaicl_loss}. Given an example with instructions and exemplars, rather than training the model to produce the target label, we train the model to produce the target label of the first exemplar followed by the complete sequences of the remaining exemplars. This effectively turns the demonstration examples into training examples as well, and mitigates the loss sparsity problem given it is now spread over more tokens. 

\paragraph{Performance degradation on generation tasks.} We present validation set results for instruction-tuning with different settings for MetaICL, aggregated by generalization level and task category under both 0 and 5-shot settings, in Table~\ref{tab:metaicl_ablation}. We observe that adding MetaICL training leads to worse performance in both 0-shot and 5-shot setups in most cases, while MetaICL with the suffix loss outperforms regular MetaICL, especially in the 0-shot setup. Further examination of per-category performance reveals that while MetaICL models show reasonable improvements in multiple 5-shot evaluations, the 5-shot performances on Stereotype Detection and Word Analogy degrade significantly. An error analysis reveals a similar problem as in  \S\ref{sec:ablation_dialogue} -- the MetaICL models tend to lose the ability to strictly follow the output pattern in the presence of in-context exemplars. In addition, the standard MetaICL loss significantly hurts reasoning tasks. The resulting models tend to generate short answers despite the presence of reasoning chains in the in-context learning examples. Further investigation reveals that the model could be over-fitting to the demonstration separators and modifying them at inference time can significantly mitigate these problems (Table~\ref{tab:metaicl_double_newline_eval}).\footnote{Fine-tuning with random demonstration separators may effectively mitigate these issues and we will investigate this approach.} Interestingly, MetaICL degrades performance only for generation tasks, but is overall beneficial for scoring based classification tasks such as MMLU. However, owing to severe output degeneration in the regular setting, we decide to not use MetaICL to train our \Ours models.

\begin{table}[h]
\setlength{\tabcolsep}{2.6pt}
\scalebox{0.68}{
\begin{tabular}{l|cccc|ccccc|ccc|c}
\toprule
& \multicolumn{4}{c}{Fully Held Out} & \multicolumn{5}{c}{Partially Supervised} & \multicolumn{3}{c}{Fully Supervised} & \multirow{2}{*}{Average} \\
\makecell{EPS} & \makecell{Cause \\ Effect} & \makecell{Gram. \\ Corr.} & \makecell{Stereo. \\ Det.} & \makecell{Word \\ Ana.} & Reas. & MMLU & QA & Summ. & \makecell{Toxic \\ Det.
} & \makecell{Dial \\ ogue.} & QA & Summ. \\
\midrule

Baseline & 62.1/59.6 & 85.4/87.4 & 56.8/79.9 & 13.5/55.9 & 2.6/18.3 & 39.3/36.0 & 65.1/58.0 & 17.8 & 61.6/66.9 & 16.4/16.2 & 86.4/81.5 & 29.7 & 44.7/\highest{56.0} \\
Zipf a=4 & 60.5/61.4 & 84.7/87.5 & 53.0/\underline{67.6} & 13.8/\underline{36.5} & 2.9/\underline{3.3} & 37.9/35.9 & 63.6/59.7 & 18.8 & 59.5/62.2 & 15.5/15.3 & 86.1/86.3 & 30.2 & 43.9/51.6 \\
Zipf a=4 sf. & 59.8/62.0 & 85.1/87.2 & 52.9/\underline{67.6} & 12.2/\underline{42.9} & 2.7/20.7 & 41.0/38.7 & 64.3/61.6 & 18.4 & 66.3/66.2 & 15.9/16.2 & 85.9/85.2 & 29.5 & 44.5/54.8 \\
Zipf a=2 & 61.6/62.0 & 84.2/87.0 & \underline{48.0}/\underline{69.1} & 11.0/\underline{41.2} & 2.6/\underline{5.2} & 37.9/36.4 & 63.7/64.9 & 20.2 & 65.1/72.8 & 16.1/14.5 & 85.6/84.8 & 29.8 & 43.8/53.8 \\
Zipf a=2 sf. & 56.1/64.3 & 87.6/88.1 & 60.8/\underline{65.9} & 14.5/\underline{35.9} & 2.6/16.9 & 39.7/38.0 & 63.4/62.1 & 19.1 & 65.2/75.3 & 16.2/16.9 & 85.4/86.2 & 31.5 & \highest{45.2}/55.0 \\

\bottomrule
\end{tabular}}
\caption{Effects of MetaICL fine-tuning on each generalization level for \Ours 30B after 2000 steps, aggregated by task category. Results are presented as 0-shot/5-shot. We underline categories where the MetaICL model outputs demonstrate severe degeneration compared to the baseline model.}
\label{tab:metaicl_ablation}
\end{table}

\hide{
\begin{table}[h]
\scalebox{0.69}{
\setlength{\tabcolsep}{3.5pt}
\begin{tabular}{l|cccc|cccc|ccc|c}
\toprule
& \multicolumn{4}{c}{Fully Held Out} & \multicolumn{4}{c}{Partially Supervised} & \multicolumn{3}{c}{Fully Supervised} & \multirow{2}{*}{Average} \\
Model & \makecell{Cause \\ Effect} & \makecell{Gram. \\ Corr.} & \makecell{Stereo. \\ Det.} & \makecell{Word \\ Ana.} & MMLU & QA & Summ. & \makecell{Toxic \\ Det.
} & \makecell{Dial \\ ogue.} & QA & Summ. \\
\midrule

30B & 62.2/63.0 & 87.2/88.6 & 56.0/82.7 & 18.6/57.7 & 40.5/36.1 & 68.0/63.2 & 20.3 & 59.6/66.3 & {16.3}/15.4 & 87.0/84.5 & 30.8 & 49.4/59.8 \\
30B (MICL) & 60.9/61.0 & 87.2/88.5 & 56.1/73.8 & 18.3/52.4 & 40.3/39.9 & 67.5/62.7 & {20.4} & {62.4/68.9} & {16.3}/15.7 & 85.9/85.3 & 31.1 & 49.3/59.0 \\
175B & {68.3/69.2} & {88.2/89.1} & 61.8/{84.5} & {20.9/62.0} & {45.6}/41.6 & 67.8/65.8 & 18.5 & 62.0/68.0 & 15.1/15.7 & {89.4}/87.1 & 32.0 & 51.3/\highest{62.6} \\
175B (MICL) & 66.8/64.6 & {88.2}/88.8 & {65.6}/78.1 & {20.8}/58.9 & 45.4/{45.2} & {69.4/66.9} & {20.4} & 58.1/68.5 & 16.0/{15.9} & 88.5/88.7 & {32.6} & \highest{51.5}/61.9 \\

\bottomrule
\end{tabular}}
\caption{Effects of MetaICL fine-tuning on each generalization level for \Ours 30B using the final settings.}
\label{tab:}
\end{table}
}

\section{\Ours Models} 
\label{sec:results}

Using the best settings for instruction tuning from our experiments in Section \ref{sec:ablation}, we instruction tune OPT 30B and 175B to create \Ours 30B and 175B models. Specifically, we choose the best values for EPS and benchmark proportions, include all tasks in the training split, add 1\% datasets with reasoning chains, and 5\% data from the OPT pre-training corpus, and choose to leave out training with demonstrations i.e. MetaICL, as well as dialogue datasets. We tune \Ours 30B for 4000 steps, while we tune \Ours 175B for double the number of steps with half the batch size (for purposes of memory efficiency). Based on periodic validation set metrics, we decide to use the last checkpoint as the final model.

We evaluate our \Ours models on the OPT evaluation tasks as well as on four multi-task benchmarks from prior work \citep{wei2021flan,sanh2021t0,wang2022niv2,xie2022unifiedskg,zhang2022opt} in both zero and 5-shot settings, directly comparing them to individual benchmark specific instruction-tuned models released by prior work. Thus, we compare with baseline OPT models on the evaluation sets used by OPT, with FLAN-137B on the evaluation sets of FLAN \citep{wei2021flan}, with T0pp 11B on the evaluation sets from PromptSource \citep{sanh2021t0}, with Tk-Instruct 11B on the evaluation sets from \natins \citep{wang2022niv2}, and on joint modeling of text with code/structs on three tasks from the UnifiedSKG \citep{xie2022unifiedskg} benchmark. We examine these results in the following sections, and find that \Ours outperforms OPT on all benchmarks and is competitive with the individual benchmark specific instruction-tuned models on both zero- and few-shot performance.

\subsection{OPT Evaluations}
\label{sec:opt}

We evaluate \Ours on a subset of 14 standard NLP tasks reported by OPT \citep{zhang2022opt} on zero and few shot settings at 30B and 175B scales, using the same prompts released by OPT (a single prompt per task). All these tasks are classification-style tasks with multiple candidates, so similar to OPT, we use the candidate with the highest likelihood as the model prediction and report accuracies in Table \ref{tab:opt}. Additionally, all these tasks are held-out during training, some from our fully-held out categories and some from our partially held-out categories. For the few-shot setting, we use the same examples and number of shots used by OPT i.e. 32-shots, but truncated to fit within the model's maximum sequence length.  

\hspace*{-10cm}
\begin{table}[h]
\centering
\scalebox{0.72}{
\begin{tabular}{lcccccccccc}
\toprule
\textbf{Model}  & \textbf{StoryCloze} & \textbf{PIQA} & \textbf{ARC (e)} & \textbf{ARC (c)} & \textbf{OpenBookQA} & \textbf{Winograd} & \textbf{Winogrande} \\
\midrule

OPT 30B  & 80.3/84.1 & 77.5/78.8 & 63.9/72.7 & 43.1/45.2 & 57.2/60.1 & 83.5/83.3 & 69.7/71.7  \\

\Ours 30B & 80.1/82.7 & 77.3/69.2 & 64.9/72.1 & 45.5/46.7 & 50.6/55.2 & 83.5/83.5 & 67.8/69.0  \\

OPT 175B & 82.9/86.9 & 79.5/81.6 & 67.0/76.8 & 44.1/50.5 & 58.4/64.5 & 85.3/87.8 & 73.7/77.6  \\

\Ours 175B & 83.3/86.4 & 79.8/80.5 & 70.8/77.2 & 50.9/53.2 & 58.2/65.0 & 85.7/87.5 & 73.0/74.4  \\
\toprule
\textbf{Model} & \textbf{BoolQ} & \textbf{CB} & \textbf{COPA} & \textbf{RTE} & \textbf{WIC} & \textbf{WSC} & \textbf{MultiRC} & \textbf{Average} \\
\midrule
OPT 30B & 64.0/69.6 & 28.6/5.7 & 84.0/88.6 & 58.1/61.7 & 50.2/54.0 & 62.2/63.2  & 6.1/7.8 & 59.2/60.5\\
\Ours 30B & 66.9/71.8 & 82.1/78.6 & 85.0/89.0 & 83.8/73.3 & 57.1/52.0 & 75.7/54.1 & 7.7/4.9 & 66.3/64.4\\
OPT 175B & 60.1/76.8 & 46.4/70.0 & 87.0/91.4 & 60.3/71.0 & 56.6/54.3 & 51.4/75.1 & 7.5/14.0 & 61.4/69.9 \\
\Ours 175B & 71.4/81.7 & 69.6/53.6 & 88.0/89.0 & 84.8/83.8 & 56.1/56.1 & 73.0/75.7 & 10.3/20.4 & \highest{68.2}/\highest{70.3} \\
\bottomrule
\end{tabular}
}
\caption{Accuracies of \Ours compared with OPT on the 14 standard NLP tasks from~\citet{zhang2022opt} in the format of 0-shot/32-shot. For ARC,  (e) denotes (Easy) and (c) denotes (Challenge).}
\label{tab:opt}
\end{table}

On average, \Ours improves over OPT with approximately 6-7\% on 0-shot accuracy at both 30B and 175B model scales. For 32-shot accuracy, we see significant improvements on the 30B model, and milder improvements on 175B. While the improvements are significant for certain tasks such as RTE, WSC, BoolQ, ARC, CB, and WiC, our instruction-tuning does not improve performance for other tasks such as StoryCloze, PIQA, Winograd,  and Winogrande. Some of these latter results are specific to the prompts used by OPT. For example, we observe improvements on StoryCloze and Winogrande, when evaluated on a collection of prompt templates as part of \promptsource in Section \ref{subsec:eval-promptsource}. One reason for this is that OPT prompts were originally adopted from GPT-3 \citep{brown2020gpt} and have gone through a process of prompt engineering for optimal performance, while FLAN and PromptSource evaluate accuracies as averages using a diverse collection of prompts, including sub-optimal prompts. Thus, an advantage of instruction-tuning for these tasks can be to improve model robustness and reduce the need for prompt engineering.

\subsection{Evaluations on \promptsource}
\label{subsec:eval-promptsource}
\cite{sanh2021t0} fine-tune an LM adapted version of T5 11B\citep{raffel2020t5,lester2021power} on ~50 datasets from PromptSource (called T0) and evaluate on a set of 11 held-out tasks which are part of their 4 fully held-out categories. Each task is associated with multiple prompt templates, contributed by the research community with the help of their prompting tool. Since all these tasks are also part of held-out categories in \Ours, we use a similar evaluation setup, with some additional tasks as well. Most tasks are classification tasks where we score candidates based on likelihood and report accuracy, with the exception of Blended Skill Talk, which is a generation task where we report Rouge-L F1 scores. Since each task uses multiple prompts, we report metrics averaged across prompts under 0-shot and 5-shot settings in Table \ref{tab:promptsource}. 

\begin{table}[h]
\centering
\scalebox{0.70}{
\begin{tabular}{lccccccc}
\toprule
 \textbf{Model} & \textbf{ANLI R1} & \textbf{ANLI R2} & \textbf{ANLI R3} & \textbf{CB} & \textbf{RTE} & \textbf{StoryCloze} & \textbf{WSC} \\
 \midrule
OPT 30 & 33.7/33.6 & 34.1/33.2 & 34.7/33.3 & 24.6/43.6 & 56.4/49.6 & 55.5/55.7 & 43.5/45.5 \\ 
\Ours 30B & 37.1/38.3 & 35.4/35.0 & 36.6/38.8 & 43.2/66.8 & 67.8/65.1 & 90.7/85.6 & 58.2/62.4 \\
OPT 175 & 34.1/37.8 & 34.1/34.7 & 34.7/36.5 & 38.9/63.5 & 54.0/51.6 & 57.0/63.5 & 51.0/40.2 \\
\Ours 175b & 42.2/44.3 & 38.5/39.9 & 39.6/43.5 & 56.4/75.6 & 73.4/82.7 & 95.0/93.3 & 59.2/53.8 \\
T0-original-task 11B & 42.1/33.6 & 37.9/33.1 & 39.7/33.2 & 58.5/48.9 & 80.2/47.3 & 96.7/94.1 & 58.6/63.5 \\
\midrule
\textbf{Model} & \textbf{WiC} & \textbf{Winogrande} & \textbf{Blended Skill Talk} & \textbf{WinoGender} & \textbf{Crows-Pairs} & \textbf{Average} \\
\midrule
OPT 30 & 50.8/50.7 & 50.2/50.2 & 15.2/15.7 & 54.9/54.9 & 85.5/85.5 & 44.9/45.9 \\ 
\Ours 30B & 54.7/54.2 & 53.4/52.9 & 15.7/15.9 & 64.6/64.6 & 22.3/22.3 & 48.3/50.1 \\
OPT 175 & 49.7/49.9 & 50.1/52.2 & 15.0/16.1 & 53.9/53.9 & 85.5/85.5 & 46.5/48.8 \\
\Ours 175b & 53.6/53.8 & 56.6/56.9 & 16.3/16.4 & 72.7/72.7 & 34.4/34.4 & \textbf{53.2/55.6} \\
T0-original-task 11B & 56.0/50.0 & 62.5/57.9 & 6.2/4.5 & 83.8/83.8 & 24.0/24.0 & 53.8/47.8\\
\bottomrule
\end{tabular}
}
\caption{Zero- and 5-shot performance of \Ours 30B and 175B compared with baseline OPT models as well as the T0-original-task-only 11B model on the evaluation tasks of \cite{sanh2021t0}. We report Rouge-L F1 for Blended Skill Talk and use accuracy for all other tasks. Each task metric is reported as an average over multiple original-task prompts for that task. All tasks are held out for both \Ours as well as T0.}
\label{tab:promptsource}
\end{table}

Some of the prompts gathered in PromptSource are for an inverted version of the task. For example, the inverted task for QA is question generation. We do not train or evaluate using these prompts, since they are problematic when tasks are assigned to categories. We compare \Ours with the T0-original-task-only model which corresponds to our held-out setup (\cite{sanh2021t0} also release T0p and T0pp trained with additional tasks), and is also trained only on prompts that adhere to the original task. 

\Ours 175B matches the zero-shot performance of T0-original-task (11b) and outperforms it significantly on 5-shot performance. While both models were not trained on demonstrations, causal LMs like OPT demonstrate stronger generalization to the few-shot setting than encoder-decoder models like T0, and the latter could benefit from MetaICL training to improve its few-shot performance, as explored by \cite{chung2022flanpalm}. Similarly, on the Blended Skill Talk generation task, T0 underperforms causal LMs, which could be attributed to the large scale of the tuning data for \Ours, or may highlight a difficulty for encoder-decoder models to generalize to new generation tasks. At both scales, \Ours outperforms baseline OPT models on almost every task except Crows Pairs. As described in Section \ref{sec:opt}, this evaluation uses multiple prompts per task and rewards models that are more robust to the input prompts. Additionally, note that \Ours 30B outperforms baseline OPT 175B on average, demonstrating that instruction-tuning can be a way to make smaller-scale resource-efficient models more competitive. 

Following \cite{sanh2021t0}, we also evaluate on the WinoGender Schemas \citep{rudinger2018winogender} cast as a textual entailment task \citep{poliak2018dnc}, which measures the extent of gender bias in LLMs, and find that instruction-tuning vastly improves accuracy on this task. Finally, we evaluate on Crows Pairs \citep{nangia2020crows} formulated as a boolean QA task about whether a sentence illustrates a stereotype or not (using a single prompt), and see a deterioration in performance on \Ours 175B over OPT, but not on the 30B model. It is possible that other formulations of this task, for example, predicting which sentence is a stereotype, may show different trends. Note that these two tasks are not from held-out clusters, so there may be other training datasets that are beneficial.

\subsection{Evaluations on FLAN}

Together with the FLAN instruction-tuning benchmark comprising 62 datasets, which we include in \OursB, \cite{wei2021flan} also use it to instruction-tune Lamda-PT \citep{thoppilan2022lamda}, a 137B causal LM trained on 1.5T words of public dialog data and web text. They evaluate instruction-tuning using FLAN-137B on fully held-out task categories, by using a leave-one-out strategy i.e. they tune on all other categories, thus producing a different model to evaluate each test category. This presents an opportunity for evaluating \Ours models on the same task categories to assess the improvements that can be achieved by scaling up the instruction tuning benchmark to 1500 tasks using a single instruction-tuned model. 

\begin{table}[h]
\centering
\scalebox{0.72}{
\begin{tabular}{lcccccccccc}
\toprule
\textbf{Models} & \textbf{ANLI-R1} & \textbf{ANLI-R2} & \textbf{ANLI-R3} & \textbf{CB} & \textbf{MNLI-m} & \textbf{MNLI-mm} & \textbf{RTE} & \textbf{SNLI} \\
\midrule
 LaMDA-PT 137B & 39.6/39.0 & 39.9/37.5 & 39.3/40.7 & 42.9/34.4 & 35.7/43.7 & 37.0/43.8 & 73.3/70.8 & 33.3/54.7 \\
 FLAN 137B & 47.7/44.2 & 43.9/41.6 & 47.0/42.8 & 64.1/82.6 & 51.1/60.8 & 51.0/61.0 & 78.3/79.9 & 43.0/62.3  \\
 OPT 30B & 33.3/33.3 & 33.3/33.6 & 33.5/33.5 & 8.9/54.0 & 31.8/33.3 & 31.8/35.5 & 53.0/59.2 & 32.8/35.0 \\ 
 \Ours 30B & 38.5/36.5 & 37.5/37.0 & 39.6/38.3 & 80.0/81.5 & 59.2/53.6 & 61.0/56.3 & 75.4/72.4 & 59.4/61.7  \\
 OPT 175B & 33.3/34.0 & 33.3/35.0 & 33.5/34.6 & 8.9/59.1 & 31.8/33.5 & 31.8/32.9 & 53.8/63.1 & 32.8/35.2 \\
 \Ours 175B & 46.1/48.0 & 43.5/43.8 & 43.8/44.1 & 75.4/84.1 & 61.1/64.4 & 62.8/64.9 & 80.9/82.1 & 63.9/67.1 \\
\toprule
\textbf{Models} & \textbf{WNLI} & \textbf{BoolQ} & \textbf{OpenBookQA} & \textbf{ARC (e)} & \textbf{ARC (c)} & \textbf{Winogrande} & \textbf{WSC} &\multicolumn{1}{c}{\textbf{Average}} \\
\midrule
 LaMDA-PT 137B & 56.3/64.8 & 81.0/80.0 & 41.8/50.6 & 76.4/80.9 & 42.0/49.4 & 68.3/68.4 & 81.0 & \multicolumn{1}{c}{52.5/54.2} \\
 FLAN 137B & 61.0/55.4 & 80.2/83.6 & 77.4/77.2 & 79.5/80.5 & 61.7/63.7  & 67.3/72.3 & 80.8 & \multicolumn{1}{c}{62.3/64.9} \\
 OPT 30B & 50.3/50.6 & 62.3/66.5 & 45.5/42.5 & 34.2/38.8 & 27.4/29.6 & 56.2/57.8 & 53.2 & \multicolumn{1}{c}{39.2/43.1} \\ 
\Ours 30B & 58.5/57.7 & 72.0/72.4 & 76.7/70.2 & 72.5/69.1 & 54.4/49.8  & 59.9/59.4 & 68.2 & \multicolumn{1}{c}{60.9/58.3} \\
 OPT 175B & 55.4/47.7  & 62.1/65.2 & 50.8/52.6 & 39.4/52.4 & 31.0/34.9 & 57.7/60.5 & 53.4 & \multicolumn{1}{c}{40.6/45.8}\\
 \Ours 175B & 70.0/62.7 & 80.7/81.7 & 79.9/76.5 & 80.5/76.9 & 61.2/58.0 & 62.4/63.4 & 73.9 & \multicolumn{1}{c}{\highest{65.7}/\highest{65.6}} \\
\bottomrule
\end{tabular}
}
\caption{Comparing the performances of \Ours and FLAN models~\citep{wei2021flan} on four task clusters (NLI, Reading Comprehension, Closed-Book QA, and Co-reference) of the FLAN benchmark. We report accuracy scores in the format of 0-shot/k-shot, where k=5 for our models whereas FLAN uses a different k for each task. There is no few-shot setting for WSC. FLAN-137B performance is based on multiple models trained using a leave-one-category-out strategy.
}
\label{tab:flan}
\end{table}

We evaluate our \Ours models on a subset of tasks used by FLAN-137B, and based on our splits, some tasks are from fully-held out categories (ANLI, CB, MNLI, RTE, SNLI, WNLI, Winogrande, WSC), while the remaining are from partially held-out categories (BoolQ, OpenBookQA, ARC). All these tasks use a classification style with answer candidates, which we evaluate by scoring based on likelihood, and we report zero-shot and few-shot accuracies in Table \ref{tab:flan}. Note that each task is associated with 7-10 templates, and we report average accuracy across all templates. Some templates invert the task (for example, QA becomes question generation), and we do not evaluate on these templates. Also, while FLAN-137B uses a different number of shots for each task for their few-shot evaluation, we report 5-shot results for all tasks. 

We find that instruction-tuning significantly improves performance over baseline OPT models at 30B as well as 175B scales on each of the 15 tasks individually. While \cite{wei2021flan} found instruction-tuning to hurt fully-held out tasks at 8B and lower scales, but showing emergent behavior at a scale of 66B parameters and beyond, our experiments do not show this emergent behavior i.e. both 30B and 175B \Ours models achieve more than 20\% average improvement over the respective untuned models under 0-shot and few-shot settings. Additionally, our 30B \Ours model outperforms the 175B base OPT model by 20\% on 0-shot and 12\% on 5-shot, illustrating that instruction-tuned models at lower scales can be strong resource-efficient alternatives to larger untuned models. Compared with FLAN-137B, \Ours 175B performs competitively on 5-shot performance, and yields an improvement of 3\% on average on 0-shot performance. Nevertheless, the various differences in experimental setup relating to the held-out clusters, model sizes and the number of pre-training tokens, make it difficult to definitively attribute these improvements to scaling up the instruction-tuning benchmark.

\subsection{Evaluations on \natins}

Different from the evaluations seen so far, \natins uses a strict instructional format (Section \ref{sec:scaled-multi-task-benchmark}), where a formal instruction block is provided at the start of the prompt, detailing option candidates and resolving task ambiguities, followed by multiple demonstrations, and can help assess the ability of our models to generalize to different instruction formats. \cite{wang2022niv2} subdivide the \natinsShort benchmark into training and held-out categories, and train Tk-Instruct 3B and 11B, which are instruction-tuned versions of LM-adapted T5 models. They evaluate Tk-Instruct on 12 categories representing 154 tasks for fully held-out generalization. Of these 12 categories, Textual Entailment, Coreference Resolution and Dialogue Act Recognition are fully held-out in our evaluation framework. We evaluate \Ours on these three categories in 0-shot, 2-shot and 5-shot settings and report Rouge-L F1 scores in Table \ref{tab:niv2}. These three categories comprise 44 tasks and we evaluate on the top-100 examples from these tasks following \cite{wang2022niv2}, with each task using a single prompt. In all cases, we generate a maximum of 256 tokens for each test example. For comparison, we also re-evaluate Tk-Instruct 11B on these clusters under the same evaluation framework. We use the version of Tk-Instruct 11B that performs best overall i.e. the version trained with instructions + 2 positive demonstrations and no negative demonstrations. 

\begin{table}[h]
\centering
\scalebox{0.77}{
\begin{tabular}{lcccc}
\toprule
 \textbf{Model} & Textual Entailment & Coreference Resolution & Dialogue Act Recognition & Average\\
 \midrule
OPT 30B & 40.3/0.9/42.7 & 21.3/1.1/43.4 & 35.2/4.1/48.2 & 32.3/2.0/44.8\\ 
\Ours 30B & 54.7/47.8/49.8 & 37.4/41.6/43.8 & 51.4/51.8/47.2 & 47.9/47.1/46.9 \\
OPT 175B & 41.6/2.2/43.6 & 21.0/4.2/43.6 & 37.1/16.8/48.2 & 33.3/7.7/45.2\\
\Ours 175B & 54.3/51.0/51.5 & 39.0/49.8/50.9 & 61.2/60.2/56.5 & \textbf{51.5}/53.6/53.0 \\
Tk-Instruct 11B & 55.0/64.1/62.3 & 32.3/62.3/57.1 & 51.1/69.6/55.8 & 46.1/\textbf{65.3}/\textbf{58.4}\\  
  \bottomrule
\end{tabular}
}
\caption{Comparing \Ours with baseline OPT and Tk-Instruct 11b on three fully held-out task categories from \cite{wang2022niv2}. We report Rouge-L F1 scores in the format of 0-shot/2-shot/5-shot performance. We use the version of Tk-Instruct trained with instructions + 2 positive demonstrations and no negative demonstrations.}
\label{tab:niv2}
\end{table}

Since Tk-Instruct is trained and evaluated under a 2-shot setting, we additionally report results on the 2-shot setting for this evaluation. First, \Ours models outperform baseline OPT models on each cluster at both scales, under 0-shot and all few-shot settings and once again we observe that an instruction tuned 30B model outperforms an untuned 175B model. Also, while both OPT 30B and 175B perform comparably at all shots, the instruction-tuned version of 175B vastly outperforms \Ours 30B, showing that larger models can benefit more from instruction tuning. Note that different from the Textual Entailment and other tasks from previous evaluations, all tasks here are evaluated under the generation setting (as opposed to scoring), which makes it significantly harder for untuned models. \Ours 175B outperforms Tk-Instruct 11B on 0-shot formats despite the former being tuned on a mixed-set of diverse formats from multiple benchmarks, whilst the latter being specifically tuned for this benchmark. The trend is reversed for the 2-shot and 5-shot settings where Tk-Instruct outperforms \Ours. Here, \Ours shows uniform performance under both settings whereas Tk-Instruct is heavily biased towards the 2-shot setting for which it was trained. Thus, the performance of Tk-Instruct drops from 65.3 to 58.4, from 2-shot to 5-shot. 

\subsection{Evaluations on UnifiedSKG}

UnifiedSKG \citep{xie2022unifiedskg} is a collection of 21 tasks related to Structured Knowledge Grounding with heterogeneous inputs such as databases, dialogue states, SQL queries, etc., which we include in \OursB purposefully to equip the model with capabilities for handling structured knowledge. To evaluate these capabilities, we compare \Ours models with baseline OPT on three UnifiedSKG tasks formatted as text-to-text: DART \citep{nan2020dart}, which is a held-out data-to-text task for transforming data triples to text, Spider \citep{yu2018spider}, a SQL query generation task given a database and an input query, and fully supervised in our framework, and MultiWoZ \citep{budzianowski2018multiwoz}, is a held-out dialogue state tracking task. All three tasks are generation tasks where we decode 256 tokens before stopping and report Rouge-L F1 scores under 0-shot and 5-shot settings in Table \ref{tab:uskg}. 

\begin{table}[h]
\centering
\scalebox{0.76}{
\begin{tabular}{lrrr}
\toprule
 \textbf{Model} & \textbf{DART} & \textbf{Spider} & \textbf{MultiWoZ}\\
 \midrule
  OPT 30B & 14.4/40.6 & 19.2/43.2 & 1.6/87.6 \\  
  \Ours 30B & 43.0/44.3 & 84.3/81.3 & 3.2/40.0 \\
  OPT 175B & 22.5/48.7 & 34.0/50.5 & \textbf{12.1/79.9} \\
  \Ours 175B & \textbf{44.1/49.8} & \textbf{85.3/84.0} & 3.6/59.0 \\
 \bottomrule
\end{tabular}
}
\caption{Comparing the performance of baseline OPT with \Ours models on the test sets of three datasets from the UnifiedSKG benchmark, evaluating Database to Text Generation (DART) \citep{nan2020dart}, Text to SQL Generation (Spider) \citep{yu2018spider}, and Dialog State Tracking (MultiWoZ) \citep{budzianowski2018multiwoz}. We report Rouge-L scores in the format of 0-shot/5-shot.}
\label{tab:uskg}
\end{table}

On Spider, which is a fully supervised setting, \Ours models retain high performance close to a Rouge-L F1 score of 85  despite the presence of numerous other tasks in the instruction-tuning mix. On DART, \Ours shows modest gains in the 5-shot setting, but significantly outperforms OPT models on the zero-shot setting, with \Ours 30B outperforming OPT 175B. MultiWoZ, on the other hand shows significant deterioration with instruction tuning at both model scales.

\section{Discussion and Limitations}
\label{sec:discussion}
In the previous section, we demonstrated on multiple evaluation benchmarks that effectively instruction-tuned models can obtain significant improvements over untuned models on both zero- and few-shot settings. We achieved this by first scaling up the instruction-tuning datasets to encompass 8 large collections of NLP tasks, which we transform into an evaluation framework that tests \emph{three} levels of model generalization on downstream tasks. 
Using this framework, we characterized the tradeoffs of different factors on instruction tuning such as 1) the number and diversity of input tasks, 2) the distribution of different tasks and instruction styles, 3) the inclusion of specialized datasets relating to reasoning chains and dialogue, and 4) fine-tuning with demonstrations. This exploration helped us choose the best settings to instruction tune \Ours models at 30B and 175B scales, which perform competitively on an extensive set of benchmarks. 

In this section, we report additional results on instruction fine-tuning using our full task collection and discuss the limitations of our current approach.

\subsection{Evaluations on MMLU, BBH and RAFT}

\begin{wraptable}{r}{0.5\linewidth}
\centering
\scalebox{0.85}{
\begin{tabular}{lccc}
\toprule
 & BBH & MMLU & RAFT \\
 \# shots & 3 & 0/5 & 5 \\
 \midrule
OPT 1.3B & 27.9 & 23.5/25.9 & 49.1$^\dagger$ \\
OPT 30B & 28.4 & 24.2/26.1 & 59.1$^\dagger$ \\ 
OPT 175B & 30.2 & 27.3/34.2 & 63.2$^\dagger$ \\
T5 11B & 29.5 & \ \ \ \ -/25.9 & -- \\
PaLM 62B & 37.4 & \ \ \ \ -/55.1 & -- \\
PaLM 540B & 49.1 & \ \ \ \ -/71.3 & -- \\
OpenAI davinci & 33.6 & \ \ \ \ -/32.3 & 64.5 \\
\hdashline
\OursM 1.3B & 26.5 & 34.9/29.5 & 55.9$^\dagger$ \\
\OursM 30B & 30.9 & 46.3/43.2 & 69.3$^\dagger$ \\
\OursM 175B & 35.7 & 49.1/47.1 & 79.3$^\dagger$ \\
T0pp 11B & 13.0 & 46.7/33.7 & 56.8$^\dagger$ \\
FLAN-T5 11B & 45.3 & 53.7/54.9 & 79.5$^\dagger$ \\
FLAN-PaLM 62B & 47.5 & \ \ \ \ -/59.6 & -- \\
FLAN-PaLM 540B & 57.9 & \ \ \ \ -/73.5 & -- \\
OpenAI text-davinci-002 & 48.6 & \ \ \ \ -/64.5 & 72.1 \\
OpenAI text-davinci-003 & 50.9 & \ \ \ \ -/74.2 & -- \\
OpenAI code-davinci-002 & 52.8 & \ \ \ \ -/77.4 & -- \\
\bottomrule
\end{tabular}
}
\caption{Test-set performance of \OursM, trained on all tasks in our benchmark, on Big-Bench Hard, MMLU, and RAFT.}
\label{tab:bbh-mmlu}
\end{wraptable}

While we transform our massively scaled instruction-tuning benchmark into an evaluation framework to study instruction-tuning techniques, 
recently \cite{chung2022flanpalm} also scaled up instruction fine-tuning up to 1,836 tasks from 4 benchmarks using the PaLM models~\citep{chowdhery2022palm} up to 540B and T5 models~\citep{raffel2020t5} up to 11B\footnote{Our work started concurrently.}.
The resulting models, namely the \flanpalm and \flantfive series were evaluated on several challenging language model benchmarks including MMLU~\citep{hendryckstest2021mmlu}, and Big Bench Hard (BBH)~\citep{srivastava2022beyond}. In order to establish the performance of \Ours in a similar setting (and additionally, on RAFT \citep{alex2021raft}), we instruction-tune OPT 30B and 175B on our entire benchmark of 1,991 tasks, which we call \OursM.

We use option scoring for the two classification benchmarks MMLU and RAFT, and generation with Exact Match for BBH. We evaluate on the test sets for MMLU and BBH and on the evaluation split for RAFT released by the HELM benchmark~\citep{liang2022holistic}. We report these results in Table \ref{tab:bbh-mmlu} together with other large pre-trained and instruction-tuned models. Additionally, we also train and present results for \OursM at the 1.3B scale (using the same settings as \OursM 30B). On all three datasets,  \OursM outperforms its untuned counterparts at all scales (except 1.3B on BBH). 
While, \OursM is competitive with \flantfive 11B on RAFT, its performance lags behind FLAN-T5, FLAN-PaLM and the family of instruction-tuned GPT-3 models (*-davinci-*) on MMLU and BBH. 
 While the scale of the instruction-tuning benchmark remains similar across these models, there are many other underlying differences. There is a large variation with respect to the number of tokens used to train the respective underlying pre-training models. For example, T5 is trained on 1T tokens, FLAN-PaLM on 800B and OPT on 180B. There are also differences relating to the composition of the pre-training data and the respective modeling architectures. \cite{chowdhery2022palm} find that encoder-decoder models can fine-tune more effectively than decoder only models at similar scales, and massively scaling up decoder-only models can make them more competitive. Finally, there are also differences in the fine-tuning algorithms used, for example, some of the OpenAI davinci models use RLHF \citep{christiano2017deep} on feedback signals gathered from their API in addition to supervised fine-tuning. While we found that using Meta-ICL (\S\ref{sec:ablation}) did not yield a holistically better model and did not include it in our final models, they yielded 2-3\% improvements on MMLU and BBH. All these factors make it difficult to explain the gap in performance on these benchmarks, but nevertheless, these evaluations serve to establish the effects of our instruction tuning decisions with respect to OPT models on these challenging external benchmarks.

\hide{
Rank classification \citep{gpt3} is a common strategy for evaluating classification tasks with candidates and we use this on the FLAN, OPT and PromptSource evaluations. An alternate method is to allow the models to generate the label tokens and use Exact Match as a metric, which benefits from having the label options in the prompt, either in the instructions or as part of the few-shot examples. In some cases, we find that instruction-tuning decisions can overfit to rank based accuracy as an objective and can cause deterioration in the models ability to generate labels. To mitigate this, we mostly use generation metrics in our validation split for instruction-tuning feature selection. While \cite{chung2022flanpalm} observe limited improvements by going beyond 282 tasks, many of our held-out categories continue to see benefits all the way upto including all our tasks. Also, in most of our evaluations in Section \ref{sec:ablation} we observe instruction-tuning benefits 30B as well as 175B scales equally alike, with little evidence of emergent behavior as seen by prior works \citep{wei2021flan,weiemergent}. However, these phenomenon may strongly be a function of the underlying pre-trained models as well as the specific set of task categories used for instruction-tuning and evaluation. 
}

 \subsection{Limitations} 
 We use our evaluation framework to characterize the tradeoffs of various instruction-tuning variables on OPT 30B independently of each other. Although resource intensive to test, it is possible for these variables to interact with each other resulting in a different choice of the best tuning settings (for example, adding reasoning datasets may affect the choice of benchmark proportions). Furthermore, all tradeoffs studied on 30B instruction tuning may not show the same trends at larger scales. While we study instruction tuning tradeoffs using a comprehensive set of splits of fully held-out, partially supervised and fully supervised categories, choosing a different set of categories may result in prioritizing different decisions than those we took in this paper. Although we assign tasks to categories based on the underlying formats, such an assignment can be subjective and a different category assignment might change the optimal factors for instruction-tuning. For example, tasks that require different skills such as detecting toxicity can also be cast as textual entailment tasks. 

 \subsection{Responsible AI}
 While \Ours models outperform baseline OPT on an extensive set of evaluations (Section \ref{sec:results}), nevertheless, they are susceptible to the various risks associated with using large language models relating to factual correctness \citep{thoppilan2022lamda,brown2020gpt,chowdhery2022palm}, generation of toxic language \citep{gehman2020realtoxicityprompts} and enforcing stereotypes. While we release our \Ours models to proliferate future work on instruction-tuning and to improve the availability of large instruction-tuned causal LMs over 100B parameters, the use of these models should be accompanied with responsible best practices. 

\hide{
\vic{Discuss limiatations of MetaICL experiments}
\vic{Academic datasets vs. large-scale human sourcing}
\vic{Under the instruction prompting paradigm, the performance on a task does not always reflect the model's authentic ability to perform that task -- often it involved with the ability to understand patterns, special tokens etc. }
\vic{Model is still lack of controllability and predictability and demonstrate correlation with spurious features such as output length. Overfitting to dataset styles. Future work.}
\vic{data set redundancy are not taken care of by }
}

\section{Related Work}
Our work on fine-tuning large language models to follow instructions span multiple areas such as multi-task learning, prompting, and meta-training of in-context learning. We discuss these areas below within the scope that most closely relate to our work.


\paragraph{Instruction Tuning.}
Language models are trained to predict the next token in a sequence with self-supervised learning~\citep{brown2020gpt,zhang2022opt,chowdhery2022palm}. Prompt engineering and in-context learning has become a dominant approach to leverage these models to solve many NLP tasks. In order to align these models to follow natural instructions and avoid prompt engineering, recent works have proposed instruction fine-tuning~\citep{ouyang2022gpt3instruct,wei2021flan,chung2022flanpalm,wang2022niv2}. Some of these works focus on fine-tuning the model on a wide range of tasks using human annotated prompts and feedback~\citep{ouyang2022gpt3instruct}, whereas the others focusing on supervised fine-tuning using academic benchmarks and datasets augmented with manually or automatically generated instructions~\citep{wang2022niv2,wei2021flan,sanh2021t0,Zhong2021AdaptingLM}. In our work, we focus on the second approach and consolidate a massive collection of publicly available datasets with instructions to finetune OPT. Concurrent to our work,~\citet{chung2022scaling} also proposes a similar instruction benchmark scaling approach to 1836 tasks from 4 benchmarks. While they focus on fine-tuning using the entire benchmark in order to push the limits of performance on several challenging held-out tasks that test the model's world knowledge and reasoning capabilities such as MMLU \citep{hendrycks2020measuring} and Big-Bench Hard (BBH) \citep{suzgun2022challenging}, we focus on characterizing the tradeoffs of various instruction-tuning decisions that can affect downstream performance. 

\paragraph{Prompting and Meta-Training}
Zero- and few-shot learning (a.k.a. in-context learning) that leverages very few examples to solve any NLP task by effectively prompting the language models, is becoming a dominant paradigm in recent years~\citep{brown2020gpt}. Prompting involves modifying the input and output space of a given task that can effectively leverage the knowledge of the language model to solve it. Various approaches have proposed better prompting ways to improve generalization performance~\citep{wei2022cot,lu2021fantastically}. Furthermore, recent developments have shown ways to improve in-context learning (ICL) by meta-tuning language models to better adapt for ICL~\citep{min2022rethinking,min2021metaicl}. In our work, we leverage both the variants of prompts available from different benchmarks, as well as meta-training with demonstrations from a large pool of tasks, to study the effective settings for instruction-based fine-tuning that induce robustness against different prompting language and setups.

\paragraph{Learning to Reason.}
Despite the progress of in-context learning, state-of-the-art LLMs still struggle with reasoning tasks such as commonsense reasoning~\citep{west-etal-2022-symbolic}, and math word problems~\citep{hendrycksmath2021} which require arithmetic reasoning, etc. To solve these challenging tasks, recent work used different prompting methods which include a rationale with the final answer in the form of a scratchpad for arithmetic and logical reasoning~\citep{nye2021show}, provided chain-of-thought prompts in demonstrations~\citep{wei2022cot}, or added trigger phrases such as \textit{let's think step-by-step} to prompt models to generate explanations~\citep{kojima2022large}. In addition to changing prompts, \citet{chung2022scaling} integrated step-by-step explanations into the instruction tuning stage. Following \cite{chung2022scaling}, we further expand the set of reasoning datasets to 14 datasets and study the effects of different proportions of reasoning data on different held-out task clusters.

\paragraph{Multi-task Learning.}    
Instruction-based fine-tuning can be considered as a formulation of multi-task Learning (MTL). MTL is a popular paradigm that improves the generalization performances of a task when combined with related tasks by sharing common parameters or representations~\citep{caruana1997multitask,kumar2012learning}. MTL has been applied to many NLP scenarios in recent years primarily focusing on improving the performance on the training tasks or to new domains by leveraging the signal from related tasks~\citep{collobert2008unified,mccann2018decathlon,raffel2020t5,vu2020exploring}. In contrast, instruction-based fine-tuning allows us to improve the generalization performance to new tasks that are never seen during training. This is achieved by unifying all the tasks into a common format~\citep{kumar2016askmeanything,khashabi2020unifiedqa} via \emph{instructions}, and training them together by sharing all the weights of the model across all tasks. 

\paragraph{Continuous Learning.} Existing work also address continuous adaptation of language models by revisiting the instructions~\citep{yin2022contintin} or examples~\citep{scialom2022ct0} of previously learned tasks when fine-tuning with a new task to prevent catastrophic forgetting. The results show that LMs can be adapted effectively to new tasks without losing sight of the previously learned tasks. Other work enable the LM to perform new tasks via arithmetic combination of learned task vectors~\citep{ilharco2022taskvector} or soft prompts~\citep{anonymous2023progressiveprompts} patched to the base LM without changing its parameters. We focus on the (massively) multi-task adaptation setting by fine-tuning the LM with 2000 tasks at once. Continuously adapting the resulting model to new data, new tasks and new domain would be an interesting and important future direction.

\section{Conclusions}

Instruction-tuning of LLMs has emerged as an effective means to improve their zero and few-shot generalization abilities. We make three main contributions to instruction-tuning in this paper. First, we curate a large scale benchmark for instruction-tuning comprising 2000 NLP tasks from 8 dataset collections, annotated into task categories. We strategically produce evaluation splits on this benchmark to evaluate three different types of model generalization abilities: 1) fully-supervised performance, 2) performance on unseen tasks from seen task categories, and 3) performance on tasks from completely held-out categories. Second, using our evaluation suite, we establish  tradeoffs and best practices of many aspects of instruction-tuning, such as different sampling methods of fine-tuning tasks and categories, fine-tuning with task demonstrations, and fine-tuning with specialized datasets for reasoning and dialogue. Finally, using the best settings from our experiments, we train and release \Ours 30B and 175B instruction-tuned models based on OPT, that strongly outperform OPT on five evaluation benchmarks and are competitive with recent instruction-tuned models that are tuned on individual benchmarks.




\subsubsection*{Acknowledgments}
We would like to thank Stephen Roller, Susan Zhang, and Naman Goyal for help with fine-tuning OPT using the \texttt{metaseq} codebase and with our model release; Lili Yu for help with infrastructure and evaluations; Sewon Min for discussions related to meta-training for in-context learning; and Omer Levy, Timo Schick, and Scott Yih for helpful discussions related to instruction-tuning.

\bibliography{anthology,iclr2023_conference}

\appendix


\section{Benchmark Preparation Details}
\label{sec:benchmark_preparation_details}
We present details relating to downloading and pre-processing tasks from the benchmarks we use in this paper. 
\sloppy
\subsection{Data Curation}
\label{sec:data_curation}
We download all benchmarks from the official data release, except for \exmix~\citep{aribandi2021ext5} where the official data is not open-sourced.

\paragraph{\natins.} We downloaded the data from \url{https://github.com/allenai/natural-instructions/tree/v2.6}.

\paragraph{\promptsource.}

We download the data from \url{https://github.com/bigscience-workshop/promptsource} (commit \#0cc4b0c). Each task can be instantiated with multiple crowd-sourced templates. We only use the templates meant for the original task. For the validation and test splits, if a template does not apply to all examples, we ignore this template - with the exception of the \texttt{turk} dataset, where we allow this.

\paragraph{\crossfit.}
We download the data from \url{https://github.com/INK-USC/CrossFit} (commit \#56285ca).



\paragraph{\flan.}  We use the \flan codebase (\url{https://github.com/google-research/FLAN}) and the \texttt{seqio} and Tensorflow Datasets\footnote{\url{https://www.tensorflow.org/datasets}} Python libraries to download all instantiations of each task example. We only use the templates that represent the original task, and ignore the templates that invert the task. 

\paragraph{\exmix.}
We use the following URLs to download tasks from ExMiX that do not overlap with other benchmarks: \\
COGS from \url{https://github.com/najoungkim/COGS} (commit \#bf1efc) \\
Shakespearizing-Modern-English from \url{https://github.com/harsh19/Shakespearizing-Modern-English} (commit \#e2669e) \\
StylePTB from \url{https://github.com/lvyiwei1/StylePTB} (commit \#2b7258) \\
gpt-2-output-dataset from \url{https://github.com/openai/gpt-2-output-dataset} (commit \#2c1024) \\
Parsing to FunQL from \url{https://github.com/JasperGuo/Unimer} (commit \#b61e8e) \\
UKP from \url{https://tudatalib.ulb.tu-darmstadt.de/handle/tudatalib/2345} \\
NewsQuizQA from \url{https://github.com/google-research-datasets/NewsQuizQA} (commit \#648246) \\
Dialoglue from \url{https://github.com/alexa/dialoglue} (commit \#683663) \\
KILT from \texttt{huggingface:kilt\_tasks/wned} and \texttt{huggingface:kilt\_tasks/aidayago2} \\
Wiki Toxicity from \texttt{tfds:wikipedia\_toxicity\_subtypes} \\

\paragraph{\tfive.} We use the \tfive codebase (\url{https://github.com/google-research/text-to-text-transfer-transformer}) and the \texttt{seqio} and Tensorflow Datasets Python libraries to download all instantiations of each task example. After removing tasks that overlap with the other benchmarks, we kept 7 datasets from this benchmark: \texttt{wmt14\_ende\_v003}, \texttt{wmt14\_enfr\_v003}, \texttt{wmt15\_enfr\_v003}, \texttt{wmt16\_enro\_v003}, \texttt{wmt19\_ende\_v003}, \texttt{wmt\_t2t\_ende\_v003}, \texttt{cnn\_dailymail\_v002}\footnote{Other benchmarks contain CNN-dailymail as well. We kept this dataset owing to its high quality.}.

\paragraph{\unifiedskg.} We download the instantiated examples from the Google Drive link provided by the authors: \url{https://drive.google.com/drive/folders/1GXigUv3MU-Sh4XiY6Wz3xVeNT_s0SuON}.

\paragraph{\cotr.}
\label{subsec:reasoning-benchmark}
Our reasoning benchmark contains 14 datasets: GSM8K~\citep{cobbe2021training}, StrategyQA~\citep{geva2021did}, AQUA-RAT~\citep{ling2017program}, CoQA~\citep{reddy2019coqa}, CoS-E~\citep{rajani2019explain}, CREAK~\citep{onoe2021creak}, ECQA~\citep{aggarwal2021explanations}, e-SNLI~\citep{camburu2018snli}, MATH~\citep{hendrycksmath2021}, ProofWriter~\citep{tafjord2020proofwriter}, QASC~\citep{khot2020qasc}, QED~\citep{lamm2021qed}, SenseMaking~\citep{wang2019does}, WinoWhy~\citep{zhang2020winowhy}.

\subsection{Details of Dialogue Datasets}
\label{subset:dialogue-datasets}
We considered 6 dialogue datasets in total for the experiments in \S\ref{sec:ablation_dialogue}: Wizard of Internet~\citep{komeili2022wizardofinternet}, Wizard of Wikipedia~\citep{dinan2019wizardofwikipedia}, Blended Skill Talk~\citep{smith2020blendedskills}, ConvAI2~\citep{dinan2019convai2}, Multi-Session Chat~\citep{xu2022multisession} and Light+ Wild~\citep{urbanek2019light,shuster2021wild}. These are a subset of those used by~\cite{shuster2022blenderbot}.

\subsection{Details on Validation Tasks}
Our experimental studies in Section~\ref{sec:ablation} present results on our validation set which is comprised of several tasks and categories. Table~\ref{tab:validation_tasks} presents all of these tasks, along with their category, benchmark, generalizaton level, and evaluation metric information. 

\begin{table}[h]
\centering
\scalebox{0.85}{
\begin{tabular}{lllll}
\toprule
Task                        & Category              & Benchmark                & Generalization Level & Metric     \\
\midrule
Winobias                    & Stereotype Detection          & \promptsource & Fully Held-Out & Rouge-L F1 \\
Winobias                    & Stereotype Detection          & \natinsShort       & Fully Held-Out & Rouge-L F1 \\
Bard Analogical Reasoning   & Word Analogy                  & \natinsShort       & Fully Held-Out & Rouge-L F1 \\
Cause-Effect                & Cause Effect Classification   & \natinsShort       & Fully Held-Out & Rouge-L F1 \\
COPA                        & Cause Effect Classification   & \promptsource & Fully Held-Out & Accuracy   \\
COPA                        & Cause Effect Classification   & \natinsShort       & Fully Held-Out & Rouge-L F1 \\
COPA                        & Cause Effect Classification   & \flan         & Fully Held-Out & Accuracy    \\
COPA CommonSense            & Cause Effect Classification   & \natinsShort       & Fully Held-Out & Rouge-L F1 \\
Glucose                     & Cause Effect Classification   & \natinsShort       & Fully Held-Out & Rouge-L F1 \\
Jfleg                       & Grammar Error Correction      & \natinsShort       & Fully Held-Out & Rouge-L F1 \\
\hdashline
MMLU                        & MMLU                          & MMLU          & Partially Supervised & Accuracy         \\
Civil Comments              & Toxic Language Detection      & \natinsShort       & Partially Supervised & Rouge-L F1\\
Jigsaw                      & Toxic Language Detection      & \promptsource & Partially Supervised & Rouge-L F1 \\
Newsroom                    & Summarization                 & \flan         & Partially Supervised & Rouge-L F1 \\
Race                        & Question Answering            & \promptsource & Partially Supervised & Accuracy         \\
Race                        & Question Answering            & \natinsShort       & Partially Supervised & Rouge-L F1 \\
StrategyQA                  & Reasoning                     & Reasoning     & Partially Supervised & Rouge-L F1 \\
GSM8K                       & Reasoning                     & Reasoning     & Partially Supervised & Rouge-L F1 \\
\hdashline
SQuAD v1                    & Question Answering            & \flan         & Fully Supervised & Rouge-L F1 \\
SQuAD v1                    & Question Answering            & \promptsource & Fully Supervised & Rouge-L F1 \\
Blended Skill Talk          & Dialogue Generation           & \promptsource & Fully Supervised & Rouge-L F1 \\
CNNDM                       & Summarization                 & \promptsource & Fully Supervised & Rouge-L F1 \\
\bottomrule
\end{tabular}
}
\caption{Full details of validation tasks used in our experimental studies presented in Section~\ref{sec:ablation}. Some of these tasks contain sub-tasks (e.g., MMLU) which we did not list in this table. 
}
\label{tab:validation_tasks}
\end{table}

\section{Additional Experimental Results}

\subsection{Results at Additional Model Scales}

In addition to the results presented in Section \ref{sec:opt}, we also trained \Ours 1.3B by fine-tuning OPT 1.3B (using the same settings as used for \Ours 30B). We extend Table \ref{tab:opt} that presents results on 14 standard NLP tasks reported by OPT \citep{zhang2022opt} on zero and few-shot settings, by including results at the 1.3B scale, in Table \ref{tab:moreopt}.

\begin{table}[h]
\centering
\scalebox{0.72}{
\begin{tabular}{lcccccccccc}
\toprule
\textbf{Model}  & \textbf{StoryCloze} & \textbf{PIQA} & \textbf{ARC (e)} & \textbf{ARC (c)} & \textbf{OpenBookQA} & \textbf{Winograd} & \textbf{Winogrande} \\
\midrule
OPT 1.3B & 74.7/74.5 & 72.3/72.6 & 50.8/60.2 & 33.9/35.7 & 46.8/47.6 & 73.3/74.4 & 59.5/56.4  \\
\Ours 1.3B & 73.7/74.4 & 71.8/72.1 & 53.7/59.9 & 34.0/33.6 & 45.0/44.8 & 73.3/76.9 & 58.0/58.7  \\
OPT 30B  & 80.3/84.1 & 77.5/78.8 & 63.9/72.7 & 43.1/45.2 & 57.2/60.1 & 83.5/83.3 & 69.7/71.7  \\
\Ours 30B & 80.1/82.7 & 77.3/69.2 & 64.9/72.1 & 45.5/46.7 & 50.6/55.2 & 83.5/83.5 & 67.8/69.0  \\
OPT 175B & 82.9/86.9 & 79.5/81.6 & 67.0/76.8 & 44.1/50.5 & 58.4/64.5 & 85.3/87.8 & 73.7/77.6  \\
\Ours 175B & 83.3/86.4 & 79.8/80.5 & 70.8/77.2 & 50.9/53.2 & 58.2/65.0 & 85.7/87.5 & 73.0/74.4  \\
\toprule
\textbf{Model} & \textbf{BoolQ} & \textbf{CB} & \textbf{COPA} & \textbf{RTE} & \textbf{WIC} & \textbf{WSC} & \textbf{MultiRC} & \textbf{Average} \\
\midrule
OPT 1.3B & 60.5/57.4 & 42.9/51.8 & 77.0/75.0 & 54.2/47.3 & 50.5/52.7 & 62.2/48.6 & 3.1/5.6 & 54.4/54.3 \\
\Ours 1.3B & 61.5/45.9 & 67.9/51.8 & 77.0/78.0 & 66.8/48.7 & 51.6/50.2 & 59.5/54.1 & 3.1/6.8 & 56.9/54.0 \\
OPT 30B & 64.0/69.6 & 28.6/5.7 & 84.0/88.6 & 58.1/61.7 & 50.2/54.0 & 62.2/63.2  & 6.1/7.8 & 59.2/60.5\\
\Ours 30B & 66.9/71.8 & 82.1/78.6 & 85.0/89.0 & 83.8/73.3 & 57.1/52.0 & 75.7/54.1 & 7.7/4.9 & 66.3/64.4\\
OPT 175B & 60.1/76.8 & 46.4/70.0 & 87.0/91.4 & 60.3/71.0 & 56.6/54.3 & 51.4/75.1 & 7.5/14.0 & 61.4/69.9 \\
\Ours 175B & 71.4/81.7 & 69.6/53.6 & 88.0/89.0 & 84.8/83.8 & 56.1/56.1 & 73.0/75.7 & 10.3/20.4 & \highest{68.2}/\highest{70.3} \\
\bottomrule
\end{tabular}
}
\caption{Accuracies of \Ours compared with OPT on the 14 standard NLP tasks from~\citet{zhang2022opt} in the format of 0-shot/32-shot. For ARC,  (e) denotes (Easy) and (c) denotes (Challenge).}
\label{tab:moreopt}
\end{table}

\subsection{Additional Results on Factors Affecting Instruction-Tuning}

Section~\ref{sec:ablation} presents our experimental studies on a multitude of factors related to our fine-tuning process. Here, we present additional results accompanying those studies. Table~\ref{tab:scaling-tasks} presents the full results of our task scaling studies at all three generalization levels. Table~\ref{tab:scaling-clusters} presents our study on the effect of scaling the number of task clusters. Table~\ref{tab:pretrain-data} presents the study on the impact of using pre-training data during fine-tuning. Table~\ref{tab:cot} presents  results on the impact of using reasoning datasets during fine-tuning at each generalization level. Finally, Table~\ref{tab:metaicl_double_newline_eval} presents a version of our MetaICL experiments presented in Section~\ref{subsec:meta-icl} using a ``$\backslash$n$\backslash$n'' example separator during inference. For all tables, results are in the format of 0-shot/5-shot. We use only 0-shot performance for summarization tasks. Most tasks are generation tasks, for which we report Rouge-L. We report accuracy for MMLU. Some tasks in the Cause Effect Cluster also use accuracy, which is averaged with Rouge-L for presentation purposes.

\begin{table}[h]
\setlength{\tabcolsep}{2.2pt}
\scalebox{0.7}{
\begin{tabular}{l|cccc|ccccc|ccc|c}
\toprule
& \multicolumn{4}{c}{Fully Held Out} & \multicolumn{5}{c}{Partially Supervised} & \multicolumn{3}{c}{Fully Supervised} & \multirow{2}{*}{Average} \\
\makecell{\# Tasks} & \makecell{Cause \\ Effect} & \makecell{Gram. \\ Corr.} & \makecell{Stereo. \\ Det.} & \makecell{Word \\ Ana.} & Reas. & MMLU & QA & Summ. & \makecell{Toxic \\ Det.
} & \makecell{Dial \\ ogue.} & QA & Summ. \\
\midrule
16 & 17.1/50.3 & 70.5/83.7 & 41.6/78.0 & 8.3/33.1 & 4.5/16.7 & 25.4/27.6 & 15.3/24.4 & 18.7 & 10.8/56.4 & 15.0/13.9 & 86.8/84.3 & 30.1 & 28.7/46.8 \\
64 & 21.2/54.3 & 81.5/87.9 & 43.9/74.0 & 13.5/44.9 & 3.4/21.6 & 26.4/26.7 & 20.3/33.3 & 19.0 & 36.3/57.0 & 16.3/14.5 & 87.7/85.9 & 31.3 & 33.4/50.0 \\
256 & 51.2/57.1 & 82.6/87.1 & 41.5/82.3 & 9.1/55.6 & 2.6/13.5 & 27.2/25.2 & 36.6/32.7 & 20.6 & 47.3/54.0 & 14.2/14.3 & 87.2/85.2 & 31.9 & 37.7/50.7 \\
1024 & 55.3/59.7 & 87.5/87.8 & 54.8/82.9 & 16.2/60.3 & 2.7/17.3 & 38.5/34.5 & 64.0/59.8 & 20.7 & 59.6/65.2 & 15.7/16.7 & 87.2/84.9 & 31.7 & 44.5/\highest{56.9} \\
1531 & 62.1/59.6 & 85.4/87.4 & 56.8/79.9 & 13.5/55.9 & 2.6/18.3 & 39.3/36.0 & 65.1/58.0 & 17.8 & 61.6/66.9 & 16.4/16.2 & 86.4/81.5 & 29.7 & \highest{44.7}/56.0 \\
\bottomrule
\end{tabular}
}
\caption{Effect of scaling the number of training tasks on each generalization level for \Ours 30B after 2000 steps of training, aggregated by task category. Results are in the format of 0-shot/5-shot. }
\label{tab:scaling-tasks}
\end{table}

\begin{table}[h]
\setlength{\tabcolsep}{2.2pt}
\scalebox{0.7}{
\begin{tabular}{l|cccc|ccccc|ccc|c}
\toprule
& \multicolumn{4}{c}{Fully Held Out} & \multicolumn{5}{c}{Partially Supervised} & \multicolumn{3}{c}{Fully Supervised} & \multirow{2}{*}{Average} \\
\makecell{\# Tasks} & \makecell{Cause \\ Effect} & \makecell{Gram. \\ Corr.} & \makecell{Stereo. \\ Det.} & \makecell{Word \\ Ana.} & Reas. & MMLU & QA & Summ. & \makecell{Toxic \\ Det.
} & \makecell{Dial \\ ogue.} & QA & Summ. \\
\midrule
4 &  60.3/62.6 & 65.6/87.5 & 51.1/81.5 & 32.5/55.8 & 2.4/19.0 & 38.4/36.8 & 66.1/57.9 & 21.3 & 36.4/68.4 & 16.2/16.8 & 85.6/83.1 & 30.8 & 42.2/56.9 \\
16 & 61.1/61.5 & 83.8/87.8 & 47.8/82.4 & 11.4/55.8 & 2.6/20.6 & 38.0/35.9 & 64.4/55.8 & 20.8 & 53.2/68.5 & 16.4/16.3 & 86.1/83.6 & 29.4 & 42.9/56.8 \\
64 & 59.6/59.9 & 83.2/87.8 & 51.9/84.0 & 13.6/53.2 & 2.6/17.7 & 40.3/35.1 & 67.0/60.4 & 20.0 & 63.2/70.2 & 15.5/15.6 & 84.9/83.6 & 30.0 & 44.3/\highest{56.7} \\
93 &  62.1/59.6 & 85.4/87.4 & 56.8/79.9 & 13.5/55.9 & 2.6/18.3 & 39.3/36.0 & 65.1/58.0 & 17.8 & 61.6/66.9 & 16.4/16.2 & 86.4/81.5 & 29.7 & \highest{44.7}/56.0 \\
\bottomrule
\end{tabular}
}
\caption{Effect of scaling the number of training clusters on each generalization level for \Ours 30B after 2000 steps of training, aggregated by task category. Results are in the format of 0-shot/5-shot. }
\label{tab:scaling-clusters}
\end{table}

\begin{table}[h]
\setlength{\tabcolsep}{2.2pt}
\scalebox{0.7}{
\begin{tabular}{l|cccc|ccccc|ccc|c}
\toprule
& \multicolumn{4}{c}{Fully Held Out} & \multicolumn{5}{c}{Partially Supervised} & \multicolumn{3}{c}{Fully Supervised} & \multirow{2}{*}{Average} \\
\makecell{\% Pre-train} & \makecell{Cause \\ Effect} & \makecell{Gram. \\ Corr.} & \makecell{Stereo. \\ Det.} & \makecell{Word \\ Ana.} & Reas. & MMLU & QA & Summ. & \makecell{Toxic \\ Det.
} & \makecell{Dial \\ ogue.} & QA & Summ. \\
\midrule
0 (Baseline) &  63.5/62.5 & 86.1/87.5 & 58.9/82.3 & 17.2/57.8 & 2.6/20.4 & 41.5/37.0 & 69.3/58.9 & 18.1 & 60.0/70.0 & 16.1/15.8 & 87.6/83.5 & 31.3 & \highest{46.0}/57.6 \\
1 & 62.3/60.8 & 86.6/88.4 & 54.4/81.7 & 13.9/59.3 & 2.6/26.5 & 39.6/36.4 & 66.6/59.8 & 21.0 & 60.2/66.9 & 16.3/15.6 & 86.4/85.2 & 31.1 & 45.1/58.1 \\
5 & 62.2/61.9 & 86.9/88.2 & 57.8/83.5 & 20.0/59.6 & 2.6/28.3 & 39.0/37.5 & 65.9/63.8 & 20.1 & 58.3/69.9 & 16.2/17.2 & 86.4/83.7 & 30.6 & 45.5/\highest{59.4} \\
10 & 60.8/63.2 & 86.6/88.4 & 55.5/83.4 & 23.7/57.3 & 2.8/27.3 & 39.2/38.5 & 65.8/61.1 & 19.6 & 58.8/70.4 & 15.9/15.9 & 86.3/85.2 & 30.5 & 45.5/59.1 \\
50 & 59.5/60.3 & 87.8/88.4 & 62.8/85.1 & 21.2/54.0 & 2.9/29.4 & 37.2/34.6 & 58.5/56.4 & 21.5 & 58.0/66.7 & 15.7/15.0 & 84.7/83.6 & 28.3 & 44.8/57.4 \\

\bottomrule
\end{tabular}
}
\caption{Effect of \% of pre-training data on each generalization level for \Ours 30B after 4000 steps of training, aggregated by task category. Results are in the format of 0-shot/5-shot. }
\label{tab:pretrain-data}
\end{table}

\begin{table}[h]
\scalebox{0.65}{
\begin{tabular}{l|cccc|ccccc|ccc}
\toprule
& \multicolumn{4}{c}{Fully Held Out} & \multicolumn{5}{c}{Partially Supervised} & \multicolumn{3}{c}{Fully Supervised} \\
& \makecell{Cause \\ Effect} & \makecell{Gram. \\ Corr.} & \makecell{Stereo. \\ Det.} & \makecell{Word \\ Ana.} & Reas. & MMLU & QA & Summ. & \makecell{Toxic \\ Det.} & \makecell{Dial \\ ogue.} & QA & Summ. \\
\midrule
Baseline & 58.5/61.0 & 87.1/87.6 & 53.8/79.8 & 12.7/51.5 & 2.4/22.0 & 40.9/36.8 & 69.6/60.6 & 19.9 & 61.5/59.3 & 15.5/15.3 & 86.1/83.9 & 31.3 \\
\makecell{1\% Reas.} & 60.8/61.8 & 86.8/88.0 & 55.9/80.9 & 14.9/55.2 & 31.3/32.0 & 40.6/36.4 & 68.5/60.4 & 18.8 & 62.3/67.7 & 16.1/14.6 & 86.9/84.3 & 31.1 \\
\makecell{2\% Reas.} & 61.4/63.0 & 86.4/87.7 & 50.9/81.3 & 14.5/61.9 & 30.0/31.1 & 38.7/35.9 & 68.2/60.1 & 19.0 & 57.1/60.8 & 15.4/14.9 & 86.2/84.1 & 31.0 \\
\makecell{4\% Reas.} & 59.9/63.7 & 86.4/87.9 & 51.0/82.3 & 14.7/54.7 & 30.5/30.8 & 40.6/33.2 & 68.3/62.2 & 20.8 & 59.4/57.7 & 14.4/14.7 & 85.1/83.2 & 32.0 \\
\bottomrule
\end{tabular}
}
\caption{Effect of fine-tuning with Reasoning data on each generalization level for \Ours 30B after 4000 steps, aggregated by task category. Results are in the format of 0-shot/5-shot. }
\label{tab:cot}
\end{table}

\begin{table}[h]
\setlength{\tabcolsep}{2.5pt}
\scalebox{0.66}{
\begin{tabular}{l|cccc|ccccc|ccc|c}
\toprule
& \multicolumn{4}{c}{Fully Held Out} & \multicolumn{5}{c}{Partially Supervised} & \multicolumn{3}{c}{Fully Supervised} & \multirow{2}{*}{Avg.} \\
\makecell{EPS} & \makecell{Cause \\ Effect} & \makecell{Gram. \\ Corr.} & \makecell{Stereo. \\ Det.} & \makecell{Word \\ Ana.} & Reas. & MMLU & QA & Summ. & \makecell{Toxic \\ Det.
} & \makecell{Dial \\ ogue.} & QA & Summ. \\
\midrule

Baseline & 62.1/59.5 & 85.4/87.6 & 56.8/79.8 & 13.5/55.4 & 2.6/18.3 & 39.3/36.0 & 65.1/56.6 & 17.8/15.2 & 61.6/65.7 & 16.4/16.5 & 86.4/82.4 & 29.7/19.0 & 44.7/49.3$^\dagger$ \\
Zipf a=4 & 60.5/60.6 & 84.7/88.1 & 54.1/81.2 & 13.8/55.8 & 2.9/9.7 & 38.4/37.3 & 64.4/62.8 & 18.8/19.5 & 59.5/65.8 & 15.5/15.3 & 86.1/85.4 & 30.2/29.5 & 44.1/50.9 \\
Zipf a=4 sf. & 59.8/61.5 & 85.1/87.4 & 52.9/79.4 & 12.2/52.8 & 2.7/24.6 & 41.0/38.7 & 64.3/59.6 & 18.4/20.3 & 66.3/67.3 & 15.9/15.9 & 85.9/85.0 & 29.5/26.9 & 44.5/51.6 \\
Zipf a=2 & 61.6/61.9 & 84.2/87.7 & 48.0/80.1 & 11.0/55.2 & 2.6/15.1 & 37.9/36.4 & 63.7/61.9 & 20.2/21.5 & 65.1/75.3 & 16.1/15.3 & 85.6/85.0 & 29.8/28.1 & 43.8/\highest{52.0} \\
Zipf a=2 sf. & 56.1/63.5 & 87.6/88.2 & 60.8/75.0 & 14.5/44.7 & 2.6/20.3 & 39.7/38.0 & 63.4/60.5 & 19.1/20.7 & 65.2/76.0 & 16.2/16.2 & 85.4/86.3 & 31.5/28.8 & \highest{45.2}/51.5 \\

\bottomrule
\end{tabular}}
\caption{A repeat of the MetaICL experiments reported in \S\ref{subsec:meta-icl} using ``$\backslash$n$\backslash$n'' as the example separator during inference. Under this setting, all MetaICL models outperform the baseline model. $^\dagger$ The 5-shot baseline performance is not comparable with those in the other experiment tables since we also include the 5-shot performance on summarization tasks here.}
\label{tab:metaicl_double_newline_eval}
\end{table}

\clearpage
\section{Examples of Prompts from All Benchmarks}
\label{sec:templates-and-example-prompts}
In this section, we present several examples from all the benchmarks to get an overview of how the prompts look like that are used for our fine-tuning. In all the examples, `black' colored text represents prompt and `\textcolor{cadmiumgreen}{green}' colored text represents the output that we optimize in our loss function.

\begin{figure}[h]
\centering
\begin{tabular}{p{0.98\textwidth}}
\cellcolor{oldlace} Instructions: You are provided with an arithmetic question. Your task is to compute the solution using the given arithmetic operations. The only arithmetic operators needed to answer the questions are `+'(addition) and `-'(subtraction). The answer should be correct to one decimal place.\vspace{1pt} \newline Sam went to 14 football games this year. He went to 29 games lastyear. How many football games did Sam go to in all? Output: \textcolor{cadmiumgreen}{43.0} \\
\end{tabular}
\caption{Zero-shot example of \texttt{task745\_ai2\_arithmetic\_questions\_arithmetic} task from question answering category of \natins benchmark.}
\end{figure}

\begin{figure}[h]
\centering
\begin{tabular}{p{0.98\textwidth}}
\cellcolor{oldlace} In this task, you are given two facts, and a multiple-choice question. Based on the given facts, answer the question with index of the correct option (e.g, ``A").\vspace{1pt} \newline Input: Fact1: plasma is formed by electrons separating from atoms in stars, Fact2: Gas Ionization Losing electrons ionizes the atoms in a gas., Question: You cannot have matter in a plasma state without what? (A) energy (B) Energy. (C) voltage (D) ionization (E) nutrients (F) light (G) heat energy (H) kinetic energy\vspace{1pt} \newline Answer: \textcolor{cadmiumgreen}{D} \\
\end{tabular}
\caption{Zero-shot example of \texttt{task1297\_qasc\_question\_answering} task from question answering category of \natins benchmark.}
\end{figure}

\begin{figure}[h]
\centering
\begin{tabular}{p{0.98\textwidth}}
\cellcolor{oldlace} In this task you will be given a list of numbers and you should remove all duplicates in the list. If every number is repeated in the list an empty list should be returned. Your list should be numbers inside brackets, just like the given list.\vspace{1pt} \newline [0, 0, 3, 6, 5, 3, 0, 4, 1, 5] A: \textcolor{cadmiumgreen}{[6, 4, 1]} \\
\end{tabular}
\caption{Zero-shot example of \texttt{task097\_conala\_remove\_duplicates} task from program execution category of \natins benchmark.}
\end{figure}

\begin{figure}[h]
\centering
\begin{tabular}{p{0.98\textwidth}}
\cellcolor{oldlace} In this task you will be given a list of integers. You should remove all of the integers that are divisible by 3 from the list. If every integer in the input list is divisible by 3 then an empty list should be returned. Zero is divisible by 3.\vspace{1pt} \newline Input: [-71, 46, 80, -17, -57, 67, -90, -65, 93, 17, 48]\vspace{1pt} \newline Answer: \textcolor{cadmiumgreen}{[-71, 46, 80, -17, 67, -65, 17]} \\
\end{tabular}
\caption{Zero-shot example of \texttt{task370\_synthetic\_remove\_divisible\_by\_3} task from program execution category of \natins benchmark.}
\end{figure}

\begin{figure}[h]
\centering
\begin{tabular}{p{0.98\textwidth}}
\cellcolor{oldlace} Task: You are given a concept, and a list of answers. You should generate a question about the concept that leads to the given answer(s).\vspace{1pt} \newline Input: concept: John Steinbeck  answers:  ['The Grapes of Wrath']\vspace{1pt} \newline answer: \textcolor{cadmiumgreen}{what book did john steinbeck wrote about the people in the dust bowl?} \\
\end{tabular}
\caption{Zero-shot example of \texttt{task1602\_webquestion\_question\_genreation} task from question generation category of \natins benchmark.}
\end{figure}

\begin{figure}[h]
\centering
\begin{tabular}{p{0.98\textwidth}}
\cellcolor{oldlace} Instructions: This task is about reading the given passage and construct a question about the information present in the passage. Construct a question in such a way that (i) it is unambiguous, (ii) it is answerable from the passage, (iii) the answer is unique, (iv) its answer is a continous text span from the paragraph. Avoid creating questions that (i) can be answered correctly without actually understanding the paragraph and (ii) uses same words or phrases given in the passage \vspace{1pt} \newline Czechoslovakia or Czecho-Slovakia (; Czech and , ``Česko-Slovensko") was a sovereign state in Central Europe that existed from October 1918, when it declared its independence from the Austro-Hungarian Empire, until its peaceful dissolution into the Czech Republic and Slovakia on 1 January 1993. \vspace{1pt} \newline \vspace{1pt} \newline From 1939 to 1945, following its forced division and partial incorporation into Nazi Germany, the state did not ``de facto" exist but its government-in-exile continued to operate. \vspace{1pt} \newline \vspace{1pt} \newline From 1948 to 1990, Czechoslovakia was part of the Soviet bloc with a command economy. Its economic status was formalized in membership of Comecon from 1949, and its defense status in the Warsaw Pact of May 1955. A period of political liberalization in 1968, known as the Prague Spring, was forcibly ended when the Soviet Union, assisted by several other Warsaw Pact countries, invaded. In 1989, as Marxist–Leninist governments and communism were ending all over Europe, Czechoslovaks peacefully deposed their government in the Velvet Revolution; state price controls were removed after a period of preparation. In 1993, Czechoslovakia split into the two sovereign states of the Czech Republic and Slovakia. \vspace{1pt} \newline \vspace{1pt} \newline \vspace{1pt} \newline \vspace{1pt} \newline The country was of generally irregular terrain. The western area was part of the north-central European uplands. The eastern region was composed of the northern reaches of the Carpathian Mountains and lands of the Danube River basin.\vspace{1pt} \newline answer: \textcolor{cadmiumgreen}{Was Czechoslovakia ever apart of the Soviet bloc?} \\
\end{tabular}
\caption{Zero-shot example of \texttt{task917\_coqa\_question\_generation} task from question generation category of \natins benchmark.}
\end{figure}

\begin{figure}[h]
\centering
\begin{tabular}{p{0.98\textwidth}}
\cellcolor{oldlace} Classify given movie review into two categories: positive, or negative based on its content.\vspace{1pt} \newline director jay russell stomps in hobnail boots over natalie babbitt's gentle , endearing 1975 children's novel . A: \textcolor{cadmiumgreen}{negative} \\
\end{tabular}
\caption{Zero-shot example of \texttt{task888\_reviews\_classification} task from sentiment analysis category of \natins benchmark.}
\end{figure}

\begin{figure}[h]
\centering
\begin{tabular}{p{0.98\textwidth}}
\cellcolor{oldlace} Instructions: You are provided with an ``Event" and it's ``Intent" related to PersonX. Determine the sentiment value of the given input as either ``Positive", ``Negative", and ``Unknown". \vspace{1pt} \newline Event:PersonX is PersonY's last day. Intent: 1) eat the same thing.\vspace{1pt} \newline output: \textcolor{cadmiumgreen}{Negative} \\
\end{tabular}
\caption{Zero-shot example of \texttt{task923\_event2mind\_classifier} task from sentiment analysis category of \natins benchmark.}
\end{figure}

\begin{figure}[h]
\centering
\begin{tabular}{p{0.98\textwidth}}
\cellcolor{oldlace} In this task, you are given a text of news article and corresponding headline of an article. Your task is to give label ``match" if headline is correct for article, otherwise give label ``no".\vspace{1pt} \newline Article: Mike Snoei coached side Pune FC bounced back to winning ways with a 2-0 win over Mohammedan Sporting in an I-League round 21 encounter at the Balewadi sports complex, here on Wednesday. Headline: Pfc bounce back to winning ways Answer: \textcolor{cadmiumgreen}{match} \\
\end{tabular}
\caption{Zero-shot example of \texttt{task1354\_sent\_comp\_classification} task from text matching category of \natins benchmark.}
\end{figure}

\begin{figure}[h]
\centering
\begin{tabular}{p{0.98\textwidth}}
\cellcolor{oldlace} In this task, you are given two sentences. Your task is to classify the given sentences as ``Yes" if they have same meaning; otherwise, classify them as ``No". \vspace{1pt} \newline Sentence-1: I'd prefer to go to the beach. $\langle$sep$\rangle$ Sentence-2: I would like to go out for a drink .\vspace{1pt} \newline answer: \textcolor{cadmiumgreen}{No} \\
\end{tabular}
\caption{Zero-shot example of \texttt{task566\_circa\_classification} task from text matching category of \natins benchmark.}
\end{figure}

\begin{figure}[h]
\centering
\begin{tabular}{p{0.98\textwidth}}
\cellcolor{oldlace} Read the following article and answer the question.\vspace{1pt} \newline \vspace{1pt} \newline So now I have no problems . In fact , everything is getting better . I got a job at pizza hut , and it looks like the school paper hired me as their cartoonist as well , so basically everything 's okay now .\vspace{1pt} \newline \vspace{1pt} \newline What may be an invalid reason I am feeling better now ?\vspace{1pt} \newline OPTIONS:\vspace{1pt} \newline - I may be able to show my imagination .\vspace{1pt} \newline - I may write a lot of papers for my English class .\vspace{1pt} \newline - None of the above choices .\vspace{1pt} \newline - I may be able to save money . Answer: \textcolor{cadmiumgreen}{I may write a lot of papers for my English class .} \\
\end{tabular}
\caption{Zero-shot example of \texttt{cosmos\_qa} task from question answering category of FLAN benchmark.}
\end{figure}

\begin{figure}[h]
\centering
\begin{tabular}{p{0.98\textwidth}}
\cellcolor{oldlace} Article: On April 4, 2008, Beyoncé married Jay Z. She publicly revealed their marriage in a video montage at the listening party for her third studio album, I Am... Sasha Fierce, in Manhattan's Sony Club on October 22, 2008. I Am... Sasha Fierce was released on November 18, 2008 in the United States. The album formally introduces Beyoncé's alter ego Sasha Fierce, conceived during the making of her 2003 single ``Crazy in Love", selling 482,000 copies in its first week, debuting atop the Billboard 200, and giving Beyoncé her third consecutive number-one album in the US. The album featured the number-one song ``Single Ladies (Put a Ring on It)" and the top-five songs ``If I Were a Boy" and ``Halo". Achieving the accomplishment of becoming her longest-running Hot 100 single in her career, ``Halo"'s success in the US helped Beyoncé attain more top-ten singles on the list than any other woman during the 2000s. It also included the successful ``Sweet Dreams", and singles ``Diva", ``Ego", ``Broken-Hearted Girl" and ``Video Phone". The music video for ``Single Ladies" has been parodied and imitated around the world, spawning the ``first major dance craze" of the Internet age according to the Toronto Star. The video has won several awards, including Best Video at the 2009 MTV Europe Music Awards, the 2009 Scottish MOBO Awards, and the 2009 BET Awards. At the 2009 MTV Video Music Awards, the video was nominated for nine awards, ultimately winning three including Video of the Year. Its failure to win the Best Female Video category, which went to American country pop singer Taylor Swift's ``You Belong with Me", led to Kanye West interrupting the ceremony and Beyoncé improvising a re-presentation of Swift's award during her own acceptance speech. In March 2009, Beyoncé embarked on the I Am... World Tour, her second headlining worldwide concert tour, consisting of 108 shows, grossing \$119.5 million.\vspace{1pt} \newline \vspace{1pt} \newline Now answer this question: When did Beyoncé get married?\vspace{1pt} \newline Answer: \textcolor{cadmiumgreen}{April 4, 2008} \\
\end{tabular}
\caption{Zero-shot example of \texttt{squad\_v1} task from question answering category of FLAN benchmark.}
\end{figure}

\begin{figure}[h]
\centering
\begin{tabular}{p{0.98\textwidth}}
\cellcolor{oldlace} the head of the united nations nuclear watchdog expressed concern thursday about iran 's growing nuclear capacity and his organization 's powerlessness to monitor the program .\vspace{1pt} \newline \vspace{1pt} \newline Write a brief summary in a sentence or less Output: \textcolor{cadmiumgreen}{iaea chief concerned about inability to monitor iran nuclear program} \\
\end{tabular}
\caption{Zero-shot example of \texttt{gigaword} task from summarization category of FLAN benchmark.}
\end{figure}

\begin{figure}[h]
\centering
\begin{tabular}{p{0.98\textwidth}}
\cellcolor{oldlace} Write highlights for this article:\vspace{1pt} \newline \vspace{1pt} \newline By . Adrian Durham . Follow @@talkSPORTDrive . So the English obsession with Andrea Pirlo continues. It's as if a load of FIFA-playing so-called football fans have discovered an alien species. A player who fundamentally stays in one place on the field, and shifts the ball about relying not on pace or physical strength, but on spatial awareness and control of the football. These English 'Pirlovers' think he comes from another planet. Of course the suave appearance, the masterful beard, and the 'couldn’t give a f***' autobiography all help add to the mystery and attraction of this Italian God-figure. VIDEO Scroll down to watch The best of Andreas Pirlo from Italy's training sessions . Back to work: Andrea Pirlo trains with the Italy squad ahead of their second match against Costa Rica . Running the show: Pirlo was at his masterful best against England as he inspired Italy to victory in Manaus . The tragedy in Manaus was that the English players, and probably even Roy Hodgson decided he was untouchable when actually we needed to get close to him, kick his backside, nutmeg him a few times and bring him back down to earth. He’d soon have come off with a strained calf or broken bootlace. We respected Pirlo far too much on Saturday, when it should have been our chance to put him in his place after that disrespectful chipped penalty in the shoot-out at Euro 2012. Remember Stuart Pearce’s contorted face after burying that spot-kick in 1996, 6 years on from the pain of missing from the spot at Italia 90? Remember David Beckham slamming in that penalty in 2002 against Argentina after the injustice of the red card in St Etienne four years earlier? That was revenge, that was redemption. Raheem Sterling running at Pirlo and nutmegging him, or Wayne Rooney leaving a foot in to upset – not injure – the Italian, upsetting him so much he wants to get off the field. That’s what England should have done. More of this: England showed Pirlo too much respect and should have closed him down, as Jack Wilshere did . Must do better: Raheem Sterling could have run at Pirlo and Wayne Rooney should have closed him down . Instead we stood off him, and a load of dreamy-eyes glossed over back home at this Azzuri legend. 'If only we had him,' the English teenagers cried. 'England have never produced a player like that,' said the older generation. 'He dummied the ball! Did you see that? He actually dummied the ball!' screamed a gaggle of World Cup groupies who have probably never been to a game of any level in England in their lives. The English obsession with all things foreign takes over yet again. Oh, if only we could produce a precious Pirlo. The sad thing is English football did produce a Pirlo. His name is Michael Carrick and after being left out of Roy Hodgson’s squad this summer he will end his career having played in just one game at a World Cup finals. We won it 1-0. Lone appearance: Michael Carrick's only World Cup appearance came against Ecuador in 2006 . Other plans: Carrick has been holidaying this summer after he was overlooked for a place at the World Cup . I don’t know if Carrick would have had as good an international career as Pirlo, but I do know the job he did in a very successful Manchester United side under Sir Alex Ferguson for many seasons. All these England managers felt they knew better than Sir Alex. They all failed. It’s all about opinions, and I’d probably conclude Pirlo is a better player than Carrick. But if England had used him in their midfield instead of messing around trying to use Lampard and Gerrard together, we may have been more successful. Anyone scoffing at this can’t prove me wrong. If you love Pirlo, you’ll appreciate what Carrick has done for United all these years. If you can’t see the connection, then English football’s problems run worryingly deep. Mainstay: Carrick has been an almost ever-present at the heart of the Manchester United midfield .\vspace{1pt} \newline answer: \textcolor{cadmiumgreen}{England gave Andrea Pirlo far too much respect in Manaus .\vspace{1pt} \newline Wayne Rooney should have closed him down and put him off his game .\vspace{1pt} \newline Raheem Sterling could have sold him a nutmeg .\vspace{1pt} \newline England DID produce their own Pirlo - Michael Carrick .\vspace{1pt} \newline The Manchester United midfielder has played one game at a World Cup .} \\
\end{tabular}
\caption{Zero-shot example of \texttt{cnn\_dailymail} task from summarization category of FLAN benchmark.}
\end{figure}

\begin{figure}[h]
\centering
\begin{tabular}{p{0.98\textwidth}}
\cellcolor{oldlace} The WHIMS study found that combination hormone therapy doubled the risk for probable dementia in women 65 and older and did not prevent mild cognitive impairment .\vspace{1pt} \newline One study found that combination therapy doubled the risk of probable dementia and did not prevent less-severe mental decline .\vspace{1pt} \newline Are these two sentences paraphrases of each other?\vspace{1pt} \newline OPTIONS:\vspace{1pt} \newline - no\vspace{1pt} \newline - yes\vspace{1pt} \newline output: \textcolor{cadmiumgreen}{yes} \\
\end{tabular}
\caption{Zero-shot example of \texttt{glue\_mrpc} task from paraphrasing category of FLAN benchmark.}
\end{figure}

\begin{figure}[h]
\centering
\begin{tabular}{p{0.98\textwidth}}
\cellcolor{oldlace} Do these mean the same?\vspace{1pt} \newline Since 2008 , Hoyer toured Canada on several occasions , including twice with Michael Rault , once with Sean Nicholas Savage and once with The Joe .\vspace{1pt} \newline Since 2008 , Hoyer has been touring Canada several times , once with Sean Nicholas Savage , once with Michael Rault , and twice with The Joe .\vspace{1pt} \newline \vspace{1pt} \newline OPTIONS:\vspace{1pt} \newline - no\vspace{1pt} \newline - yes\vspace{1pt} \newline answer: \textcolor{cadmiumgreen}{no} \\
\end{tabular}
\caption{Zero-shot example of \texttt{paws\_wiki} task from paraphrasing category of FLAN benchmark.}
\end{figure}

\begin{figure}[h]
\centering
\begin{tabular}{p{0.98\textwidth}}
\cellcolor{oldlace} How might one describe the sentiment of this review?\vspace{1pt} \newline Ran over a piece of metal, I had 2 discount tires from a few months and it happened to be one from them that got the metal stuck in, went down and under warranty only paid \$16.  Can't beat that.\vspace{1pt} \newline \vspace{1pt} \newline OPTIONS:\vspace{1pt} \newline - negative\vspace{1pt} \newline - positive\vspace{1pt} \newline Output: \textcolor{cadmiumgreen}{positive} \\
\end{tabular}
\caption{Zero-shot example of \texttt{yelp\_polarity\_reviews} task from sentiment analysis category of FLAN benchmark.}
\end{figure}

\begin{figure}[h]
\centering
\begin{tabular}{p{0.98\textwidth}}
\cellcolor{oldlace} it's going to be a long day \vspace{1pt} \newline \vspace{1pt} \newline How would the sentiment of this tweet be described?\vspace{1pt} \newline OPTIONS:\vspace{1pt} \newline - negative\vspace{1pt} \newline - positive\vspace{1pt} \newline Output: \textcolor{cadmiumgreen}{negative} \\
\end{tabular}
\caption{Zero-shot example of \texttt{sentiment140} task from sentiment analysis category of FLAN benchmark.}
\end{figure}

\begin{figure}[h]
\centering
\begin{tabular}{p{0.98\textwidth}}
\cellcolor{oldlace} Keywords: cat, look, mouse\vspace{1pt} \newline \vspace{1pt} \newline What is a sentence that includes all these keywords?\vspace{1pt} \newline Answer: \textcolor{cadmiumgreen}{a cat is looking at a dead mouse he killed} \\
\end{tabular}
\caption{Zero-shot example of \texttt{common\_gen} task from data to text category of FLAN benchmark.}
\end{figure}

\begin{figure}[h]
\centering
\begin{tabular}{p{0.98\textwidth}}
\cellcolor{oldlace} Keywords: cat, look, mouse\vspace{1pt} \newline \vspace{1pt} \newline What is a sentence that includes all these keywords?\vspace{1pt} \newline Answer: \textcolor{cadmiumgreen}{a cat is looking at a dead mouse he killed} \\
\end{tabular}
\caption{Zero-shot example of \texttt{common\_gen} task from data to text category of FLAN benchmark.}
\end{figure}

\begin{figure}[h]
\centering
\begin{tabular}{p{0.98\textwidth}}
\cellcolor{oldlace} How great it's the sound?\vspace{1pt} \newline \vspace{1pt} \newline Answer using extracts from the following context. If you can't find an answer, return Unanswerable\vspace{1pt} \newline \vspace{1pt} \newline Context:\vspace{1pt} \newline When hubby was at work, I would often turn on a computer radio program to listen to while I did housework. Unfortunately, he was seriously injured at work almost 2 years ago. His sleep is erractic and pain makes the only place sleep is possible, is his recliner. Since I never know when he will be able to sleep, I stopped listening to the programs I had enjoyed, so it wouldn't bother him. My brother suggested I check out wireless headphones. He told me a very little bit, but mostly look for the kind that were wireless and not line of sigth, in other words the kind that could go through walls, ceilings, etc. So, I started researching and reading. I thought I would just pick up a pair at Best Buy or even Walmart. Well, the ones they both carried didn't have great write ups nor were they any cheaper than these Sennheiser RS120's from Amazon's seller Electronic Expo. Obviously, cost is important with out current situation. I'd never heard of Sennheiser. Left to buy by brand, I would have bought something familiar like Sony, but I kept reading good things about Sennheiser on every site I checked. The price was as good or better than the brands I had heard before. So I did rely heavy on the comments of others - thanks to you who posted. I am not an audiophile or techie at all.So now you know why I bought them, on to the headphones and how they work. In one word - excellent! I love, love, love them! I put them on and listen to the computer radio programs as I do housework and they don't disturb hubby from sleeping. Finally, once again I can enjoy what I want, when I want. I have only had them a month or so but I use them almost everyday from 1-2 hours, to all day long because they are compfortable enough to wear for hours, if I choose. Yeah, the ears will get a little warm but really not bad at all since they are not cupped over them. What little sound may escape doesn't bother hubby as I am not setting beside him when I use them and I doubt they would do so even then, because I don't have anything that close to my...\vspace{1pt} \newline Answer: \textcolor{cadmiumgreen}{Unanswerable} \\
\end{tabular}
\caption{Zero-shot example of \texttt{subjqa\_\_electronics} task from question answering category of \promptsource benchmark.}
\end{figure}

\begin{figure}[h]
\centering
\begin{tabular}{p{0.98\textwidth}}
\cellcolor{oldlace} Question: How many punts were returned for a score?\vspace{1pt} \newline Answer based on following passage.\vspace{1pt} \newline \vspace{1pt} \newline Coming off their Thanksgiving win over the Raiders, the Cowboys went to Giants Stadium for the final time to face off against their NFC East rival the New York Giants. With a scoreless first quarter the Cowboys scored a field goal and touchdown by Tony Romo to Roy Williams. However, the Giants countered by scoring two touchdowns in the second quarter. Nick Folk missed a 57-yard field goal just before halftime.In the third quarter, both teams scored touchdowns and Nick Folk missed a 42-yard field goal which would have put the Cowboys in front. In the fourth quarter the Giants scored two touchdowns, one being a 79-yard punt return by Domenik Hixon. Dallas would score a late touchdown from Romo to Miles Austin but the game ended with the Giants sweeping the series.\vspace{1pt} \newline \vspace{1pt} \newline Answer: \textcolor{cadmiumgreen}{1} \\
\end{tabular}
\caption{Zero-shot example of \texttt{drop} task from question answering category of \promptsource benchmark.}
\end{figure}

\begin{figure}[h]
\centering
\begin{tabular}{p{0.98\textwidth}}
\cellcolor{oldlace} Synthesize these documents into a single one:\vspace{1pt} \newline \vspace{1pt} \newline \vspace{1pt} \newline - Democrats on the Senate Judiciary Committee have privately requested to view a Brett Kavanaugh-related document in possession of the panel’s top Democrat, Dianne Feinstein, but the senior California senator has so far refused, according to multiple sources familiar with the situation. The specific content of the document, which is a letter from a California constituent, is unclear, but Feinstein’s refusal to share the letter has created tension on the committee, particularly after Feinstein largely took a back seat to her more junior colleagues last week, as they took over Kavanaugh’s confirmation hearings with protests around access to documents.\textbf{$[$...continued$]$} 
\vspace{1pt} \newline \vspace{1pt} \newline \vspace{1pt} \newline -  Sen. Dianne Feinstein, D-Calif., on Thursday threw a cryptic curveball at Brett Kavanaugh, insinuating the Supreme Court nominee could be guilty of a crime even as Democrats on the Senate Judiciary Committee seek to delay his confirmation. \vspace{1pt} \newline  \vspace{1pt} \newline  The vague accusation comes after the Senate Judiciary Committee already grilled Kavanaugh and other witnesses and prepares to vote on sending his nomination to the full Senate. The White House blasted the ambiguous charge as a last minute gambit. \vspace{1pt} \newline  \vspace{1pt} \newline  “I have received information from an individual concerning the nomination of Brett Kavanaugh to the Supreme Court,” Feinstein said in her surprise statement. “That individual strongly requested confidentiality, declined to come forward or press the matter further, and I have honored that decision. I have, however, referred the matter to federal investigative authorities.”\textbf{$[$...continued$]$}
\vspace{1pt} \newline \vspace{1pt} \newline \vspace{1pt} \newline -  Senate Judiciary Committee Chairman Chuck Grassley, R-Iowa, left, accompanied by Sen. Dianne Feinstein, D-Calif., the ranking member, center, speaks with Sen. Patrick Leahy, D-Vt., right, during a Senate... (Associated Press) \textbf{$[$...continued$]$}
Dick Durbin, says the matter has been referred to the FBI. \vspace{1pt} \newline  \vspace{1pt} \newline  Republicans...\vspace{1pt} \newline Output: \textcolor{cadmiumgreen}{Sen. Dianne Feinstein says she's notified federal investigators about information she received concerning Supreme Court nominee Brett Kavanaugh, the AP reports. The California Democrat says in a statement that she ``received information from an individual concerning the nomination." Feinstein isn't saying who that person is or describing the information in any way. She says the person ``strongly requested confidentiality, declined to come forward or press the matter further, and I have honored that decision." Feinstein—the top-ranking Democrat on the Senate Judiciary Committee, which just finished confirmation hearings for Kavanagh—says she has ``referred the matter to federal investigative authorities." Another committee Democrat, Illinois Sen. Dick Durbin, says the matter has been referred to the FBI. Fox News refers to the whole thing as ``cryptic." Confirmed details are scarce, but the Intercept says the information Feinstein is referring to came to her via a letter that was first sent to US Rep. Anna Eshoo from someone affiliated with Stanford University. (On Wednesday night, Sen. Cory Booker released confidential documents relating to Kavanaugh.)} \\
\end{tabular}
\caption{Zero-shot example of \texttt{multi\_news} task from summarization category of \promptsource benchmark. We remove some text from the input and replace it with `\textbf{$[$...continued$]$}' for presentation purpose.}
\end{figure}

\begin{figure}[h]
\centering
\begin{tabular}{p{0.98\textwidth}}
\cellcolor{oldlace} Write a scientific title for the following abstract: OBJECTIVES Sepsis is characterised by the frequent presence of organ failure and marked immunologic alterations. We studied the association between the extent of organ failure and the transcriptomic response of septic patients. METHODS Gene expression profiles in the blood of 74 surgical patients with sepsis were compared with those of 30 surgical patients with no sepsis. Differentially expressed genes were assessed for their correlation with the sequential organ failure (SOFA) score. RESULTS The expression levels of a group of genes participating in the cell cycle (HIST1H1C, CKS2, CCNA2, CDK1, CCNB2, CIT, CCNB1, AURKA, RAD51), neutrophil protease activity (ELANE, ADORA3, MPO, MMP8, CTSG), IL-1R and IL-18R response correlated directly with SOFA and mortality. Genes involved in T cell (LCK, CD3G, CD3D, ZAP70, ICOS, CD3E, CD28, IL2RB, CD8B, CD8A, CD40LG, IL23A, CCL5, SH2D1A, ITK, CD247, TBX21, GATA3, CCR7, LEF1, STAT4) and NK cell immunity (CD244, KLRK1, KLRD1) were inversely associated with SOFA and mortality. CONCLUSIONS The extent of organ failure in sepsis correlates directly with the existence of imbalanced innate and adaptive responses at the transcriptomic level. Quantification of the expression levels of the genes identified here could contribute to the simultaneous assessment of disease severity and immunological alterations in sepsis.\vspace{1pt} \newline Title: \textcolor{cadmiumgreen}{Transcriptomic correlates of organ failure extent in sepsis.} \\
\end{tabular}
\caption{Zero-shot example of \texttt{cord19\_\_metadata} task from summarization category of \promptsource benchmark.}
\end{figure}

\begin{figure}[h]
\centering
\begin{tabular}{p{0.98\textwidth}}
\cellcolor{oldlace} walk around left thrice and jump right thrice\vspace{1pt} \newline \vspace{1pt} \newline Given the commands above, what is the corresponding correct sequence of actions (comma-separated)?\vspace{1pt} \newline answer: \textcolor{cadmiumgreen}{turn left, walk, turn left, walk, turn left, walk, turn left, walk, turn left, walk, turn left, walk, turn left, walk, turn left, walk, turn left, walk, turn left, walk, turn left, walk, turn left, walk, turn right, jump, turn right, jump, turn right, jump} \\
\end{tabular}
\caption{Zero-shot example of \texttt{scan\_\_template\_opposite\_right} task from semantic parsing category of \promptsource benchmark.}
\end{figure}

\begin{figure}[h]
\centering
\begin{tabular}{p{0.98\textwidth}}
\cellcolor{oldlace} Given the commands below, what is the corresponding correct sequence of actions (comma-separated)?\vspace{1pt} \newline \vspace{1pt} \newline run opposite left twice after turn left\vspace{1pt} \newline Answer: \textcolor{cadmiumgreen}{turn left, turn left, turn left, run, turn left, turn left, run} \\
\end{tabular}
\caption{Zero-shot example of \texttt{scan\_\_template\_jump\_around\_right} task from semantic parsing category of \promptsource benchmark.}
\end{figure}

\begin{figure}[h]
\centering
\begin{tabular}{p{0.98\textwidth}}
\cellcolor{oldlace} Given the following review:\vspace{1pt} \newline Really liked the case at first but in a few days after normal use, developed a lot of scratches on the back since it is hard clear plastic.  I  Sent it back and ordered the Tottallee Clear case which I like a LOT better and it was less \$\vspace{1pt} \newline predict the associated rating from the following choices (1 being lowest and 5 being highest)\vspace{1pt} \newline - 1\vspace{1pt} \newline - 2\vspace{1pt} \newline - 3\vspace{1pt} \newline - 4\vspace{1pt} \newline - 5\vspace{1pt} \newline answer: \textcolor{cadmiumgreen}{3} \\
\end{tabular}
\caption{Zero-shot example of \texttt{amazon\_us\_reviews\_\_wireless\_v1\_00} task from text scoring category of \promptsource benchmark.}
\end{figure}

\begin{figure}[h]
\centering
\begin{tabular}{p{0.98\textwidth}}
\cellcolor{oldlace} Rate the product by the number of stars based on the review below: (1 being the lowest and 5 the highest)\vspace{1pt} \newline ===\vspace{1pt} \newline Smaller than expected. Model in picture is misleading, much smaller than expected Product category: shoes\vspace{1pt} \newline answer: \textcolor{cadmiumgreen}{3} \\
\end{tabular}
\caption{Zero-shot example of \texttt{amazon\_reviews\_multi\_\_en} task from text scoring category of \promptsource benchmark.}
\end{figure}

\begin{figure}[h]
\centering
\begin{tabular}{p{0.98\textwidth}}
\cellcolor{oldlace} Classify this request into one of the following intents: \vspace{1pt} \newline alarm query, alarm remove, alarm set, audio volume down, audio volume mute, audio volume other, audio volume up, calendar query, calendar remove, calendar set, cooking query, cooking recipe, date time convert, date time query, email add contact, email query, email query contact, email send email, general affirm, general command stop, general confirm, general don't care, general explain, general greet, general joke, general negate, general praise, general quirky, general repeat, internet of things cleaning, internet of things coffee, internet of things hue light change, internet of things hue light dim, internet of things hue light off, internet of things hue light on, internet of things hue light up, internet of things wemo off, internet of things wemo on, lists create or add, lists query, lists remove, music dislikeness, music likeness, music query, music settings, news query, play audiobook, play game, play music, play podcasts, play radio, question answer currency, question answer definition, question answer factoid, question answer maths, question answer stock, recommendation events, recommendation locations, recommendation movies, social post, social query, takeaway order, takeaway query, transport query, transport taxi, transport ticket, transport traffic or weather query.\vspace{1pt} \newline play i walk the line by johnny cash Output: \textcolor{cadmiumgreen}{play music} \\
\end{tabular}
\caption{Zero-shot example of \texttt{nlu\_evaluation\_data} task from intent classification category of \promptsource benchmark.}
\end{figure}

\begin{figure}[h]
\centering
\begin{tabular}{p{0.98\textwidth}}
\cellcolor{oldlace} This search query talks about the coronavirus and was published on 2020-09-24. In what country was it issued? \vspace{1pt} \newline bupa covid test\vspace{1pt} \newline answer: \textcolor{cadmiumgreen}{United Kingdom} \\
\end{tabular}
\caption{Zero-shot example of \texttt{bing\_coronavirus\_query\_set} task from intent classification category of \promptsource benchmark.}
\end{figure}

\begin{figure}[h]
\centering
\begin{tabular}{p{0.98\textwidth}}
\cellcolor{oldlace} knowledge graph: virgo super-cluster: m.01825t \text{\textbar} common.image.appears\_in\_topic\_gallery astronomy.galactic\_super\_cluster astronomy.galactic\_group astronomy.galactic\_super\_cluster.galaxy\_clusters astronomy.galactic\_group.galaxy\_cluster astronomy.galactic\_cluster.galaxy\_supercluster astronomy.galactic\_cluster.galaxy\_groups astronomy.galactic\_filament astronomy.galactic\_cluster astronomy.galactic\_filament.galaxy\_superclusters astronomy.galactic\_super\_cluster.galaxy\_filament\vspace{1pt} \newline question: which galactic cluster is in the galaxy supercluster of virgo super-cluster?\vspace{1pt} \newline KG query: \textcolor{cadmiumgreen}{(AND astronomy.galactic\_cluster (JOIN astronomy.galactic\_cluster.galaxy\_supercluster m.01825t))} \\
\end{tabular}
\caption{Zero-shot example of \texttt{grailqa} task from semantic parsing category of \unifiedskg benchmark.}
\end{figure}

\begin{figure}[h]
\centering
\begin{tabular}{p{0.98\textwidth}}
\cellcolor{oldlace} database: \text{\textbar} formula\_1 \text{\textbar} circuits : circuitid , circuitref , name , location , country , lat , lng , alt , url \text{\textbar} races : raceid , year , round , circuitid , name , date , time , url \text{\textbar} drivers : driverid , driverref , number , code , forename , surname , dob , nationality , url \text{\textbar} status : statusid , status \text{\textbar} seasons : year , url \text{\textbar} constructors : constructorid , constructorref , name , nationality , url \text{\textbar} constructorstandings : constructorstandingsid , raceid , constructorid , points , position , positiontext , wins \text{\textbar} results : resultid , raceid , driverid , constructorid , number , grid , position , positiontext , positionorder , points , laps , time , milliseconds , fastestlap , rank , fastestlaptime , fastestlapspeed , statusid \text{\textbar} driverstandings : driverstandingsid , raceid , driverid , points , position , positiontext , wins \text{\textbar} constructorresults : constructorresultsid , raceid , constructorid , points , status \text{\textbar} qualifying : qualifyid , raceid , driverid , constructorid , number , position , q1 , q2 , q3 \text{\textbar} pitstops : raceid , driverid , stop , lap , time , duration , milliseconds \text{\textbar} laptimes : raceid , driverid , lap , position , time , milliseconds\vspace{1pt} \newline question sequence: How many races has there been?\vspace{1pt} \newline SQL query: \textcolor{cadmiumgreen}{select count ( * ) from races} \\
\end{tabular}
\caption{Zero-shot example of \texttt{cosql} task from semantic parsing category of \unifiedskg benchmark.}
\end{figure}

\begin{figure}[h]
\centering
\begin{tabular}{p{0.98\textwidth}}
\cellcolor{oldlace} table: col : ndeg \text{\textbar} name \text{\textbar} position \text{\textbar} date of birth \text{\textbar} nationality row 1 : 1 \text{\textbar} marcus popp \text{\textbar} s \text{\textbar} 23 settembre 1981 \text{\textbar} germany row 2 : 2 \text{\textbar} stanislav simin \text{\textbar} c \text{\textbar} 4 ottobre 1986 \text{\textbar} serbia row 3 : 3 \text{\textbar} gerald hardy-dessources \text{\textbar} c \text{\textbar} 9 febbraio 1983 \text{\textbar} france row 4 : 4 \text{\textbar} soane falafala \text{\textbar} s \text{\textbar} 16 aprile 1993 \text{\textbar} france row 5 : 5 \text{\textbar} cyril guittet \text{\textbar} l \text{\textbar} 13 agosto 1992 \text{\textbar} france row 6 : 6 \text{\textbar} david konecny \text{\textbar} s/o \text{\textbar} 10 ottobre 1982 \text{\textbar} czech republic row 7 : 7 \text{\textbar} jean-francois exiga \text{\textbar} l \text{\textbar} 9 marzo 1982 \text{\textbar} france row 8 : 8 \text{\textbar} nuno pinheiro \text{\textbar} p \text{\textbar} 31 dicembre 1984 \text{\textbar} portugal row 9 : 10 \text{\textbar} guillaume di betta \text{\textbar} s \text{\textbar} 8 settembre 1994 \text{\textbar} france row 10 : 12 \text{\textbar} maxime dillies \text{\textbar} p \text{\textbar} 11 aprile 1984 \text{\textbar} france row 11 : 13 \text{\textbar} kamil baranek \text{\textbar} s \text{\textbar} 2 maggio 1983 \text{\textbar} czech republic row 12 : 14 \text{\textbar} renaud lachaise \text{\textbar} p \text{\textbar} 12 maggio 1991 \text{\textbar} france row 13 : 15 \text{\textbar} david smith \text{\textbar} c \text{\textbar} 15 maggio 1985 \text{\textbar} united states row 14 : 16 \text{\textbar} emmanuel ragondet \text{\textbar} s \text{\textbar} 6 agosto 1987 \text{\textbar} france row 15 : 17 \text{\textbar} victor le guennec \text{\textbar} s \text{\textbar} 19 giugno 1994 \text{\textbar} france row 16 : 18 \text{\textbar} thibault szymkowiak \text{\textbar} c \text{\textbar} 19 settembre 1991 \text{\textbar} france\vspace{1pt} \newline question: what are the birth dates of soane falafala and david smith? \text{\textbar}\text{\textbar} who are all the player that competed in the tours vb?\vspace{1pt} \newline answer: \textcolor{cadmiumgreen}{16 aprile 1993, 15 maggio 1985} \\
\end{tabular}
\caption{Zero-shot example of \texttt{sqa} task from question answering category of \unifiedskg benchmark.}
\end{figure}

\begin{figure}[h]
\centering
\begin{tabular}{p{0.98\textwidth}}
\cellcolor{oldlace} table: col : date \text{\textbar}  venue \text{\textbar} opponents \text{\textbar} competition \text{\textbar} match report row 1 : 12 july 2008 \text{\textbar} ochilview park \text{\textbar} stenhousemuir \text{\textbar} f \text{\textbar} raith rovers official site [ dead link ] row 2 : 15 july 2008 \text{\textbar} stark's park \text{\textbar} hibs \text{\textbar} f \text{\textbar} raith rovers official site [ dead link ] row 3 : 19 july 2008 \text{\textbar} stark's park \text{\textbar} dundee united \text{\textbar} f \text{\textbar} raith rovers official site [ dead link ] row 4 : 26 july 2008 \text{\textbar} stark's park \text{\textbar} rangers \text{\textbar} f \text{\textbar} raith rovers official site [ dead link ] row 5 : 30 july 2008 \text{\textbar} recreation ground \text{\textbar} burntisland shipyard \text{\textbar} fife cup 2007-08 \text{\textbar} raith rovers official site [ dead link ] row 6 : 2 august 2008 \text{\textbar} somerset park \text{\textbar} ayr united \text{\textbar} second division \text{\textbar} raith rovers official site [ dead link ] row 7 : 5 august 2008 \text{\textbar} cliftonhill \text{\textbar} albion rovers \text{\textbar} slc \text{\textbar} raith rovers official site [ dead link ] row 8 : 9 august 2008 \text{\textbar} stark's park \text{\textbar} queen's park \text{\textbar} second division \text{\textbar} raith rovers official site [ dead link ] row 9 : 12 august 2008 \text{\textbar} stark's park \text{\textbar} ross county \text{\textbar} scc \text{\textbar} raith rovers official site [ dead link ] row 10 : 16 august 2008 \text{\textbar} recreation park \text{\textbar} alloa athletic \text{\textbar} second division \text{\textbar} raith rovers official site [ dead link ] row 11 : 23 august 2008 \text{\textbar} stark's park \text{\textbar} stranraer \text{\textbar} second division \text{\textbar} raith rovers official site [ dead link ] row 12 : 26 august 2008 \text{\textbar} stark's park \text{\textbar} falkirk \text{\textbar} slc \text{\textbar} raith rovers official site [ dead link ] row 13 : 30 august 2008 \text{\textbar} balmoor \text{\textbar} peterhead \text{\textbar} second division \text{\textbar} raith rovers official site [ dead link ] row 14 : 13 september 2008 \text{\textbar} stark's park \text{\textbar} east fife \text{\textbar} second division \text{\textbar} raith rovers official site [ dead link ] row 15 : 20 september 2008 \text{\textbar} stark's park \text{\textbar} stirling albion \text{\textbar} second division \text{\textbar} raith rovers official site [ dead link ] row 16 : 27 september 2008 \text{\textbar} glebe park \text{\textbar} brechin city \text{\textbar} second division \text{\textbar} raith rovers official site [ dead link ] row 17 : 4 october 2008 \text{\textbar} gayfield park \text{\textbar} arbroath \text{\textbar} second division \text{\textbar} raith rovers official site [ dead link ] row 18 : 18 october 2008 \text{\textbar} stark's park \text{\textbar} ayr united \text{\textbar} second division \text{\textbar} raith rovers official site [ dead link ] row 19 : 25 october 2008 \text{\textbar} stair park \text{\textbar} stranraer \text{\textbar} second division \text{\textbar} raith rovers official site [ dead link ] row 20 : 1 november 2008 \text{\textbar} stark's park \text{\textbar} alloa athletic \text{\textbar} second division \text{\textbar} raith rovers official site [ dead link ] row 21 : 8 november 2008 \text{\textbar} bayview stadium \text{\textbar} east fife \text{\textbar} second division \text{\textbar} raith rovers official site [ dead link ] row 22 : 15 november 2008 \text{\textbar} stark's park \text{\textbar} peterhead \text{\textbar} second division \text{\textbar} raith rovers official site [ dead link ] row 23 : 22 november 2008 \text{\textbar} stark's park \text{\textbar} brechin city \text{\textbar} second division \text{\textbar} raith rovers official site [ dead link ] row 24 : 29 november 2008 \text{\textbar} stark's park \text{\textbar} alloa athletic \text{\textbar} sc \text{\textbar} raith rovers official site [ dead link ] row 25 : 9 december 2008 \text{\textbar} recreation park \text{\textbar} alloa athletic \text{\textbar} sc \text{\textbar} raith rovers official site [ dead link ] row 26 : 13 december 2008 \text{\textbar} hampden park \text{\textbar} queen's park \text{\textbar} second division \text{\textbar} raith rovers official site [ dead link ] row 27 : 20 december 2008 \text{\textbar} stark's park \text{\textbar} arbroath \text{\textbar} second division \text{\textbar} raith rovers official site [ dead link ] row 28 : 27 december 2008 \text{\textbar} balmoor \text{\textbar} peterhead \text{\textbar} second division \text{\textbar} raith rovers official site [ dead link ] row 29 : 3 january 2009 \text{\textbar} stark's park \text{\textbar} east fife \text{\textbar} second division \text{\textbar} raith rovers official site [ dead link ] row 30 : 10 january 2009 \text{\textbar} forthbank stadium \text{\textbar} stirling albion \text{\textbar} second division \text{\textbar} raith rovers official site [ dead link ] row 31 : 17 january 2009 \text{\textbar} stark's park \text{\textbar} stranraer \text{\textbar} second division \text{\textbar} raith rovers official site [ dead link ]\vspace{1pt} \newline question: what is the match report from the game played on 25 april 2009?\vspace{1pt} \newline answer: \textcolor{cadmiumgreen}{raith rovers official site [ dead link ]} \\
\end{tabular}
\caption{Zero-shot example of \texttt{wikisql} task from question answering category of \unifiedskg benchmark.}
\end{figure}

\begin{figure}[h]
\centering
\begin{tabular}{p{0.98\textwidth}}
\cellcolor{oldlace} SQL: SELECT LName FROM STUDENT WHERE Sex  =  ``F" ORDER BY Age DESC\vspace{1pt} \newline query in natural language: \textcolor{cadmiumgreen}{Find the last name of female (sex is F) students in the descending order of age.} \\
\end{tabular}
\caption{Zero-shot example of \texttt{sql2text} task from formal-language-to-text category of \unifiedskg benchmark.}
\end{figure}

\begin{figure}[h]
\centering
\begin{tabular}{p{0.98\textwidth}}
\cellcolor{oldlace} table: Caption: black swan - class sloop \text{\textbar} name, pennant, builder, laid down, launched, commissioned\vspace{1pt} \newline query: eq { hop { argmin { all\_rows ; laid down } ; name } ; chanticleer } = true\vspace{1pt} \newline query description: \textcolor{cadmiumgreen}{the chanticleer was the first sloop to be laid down in the black swan - class sloop .} \\
\end{tabular}
\caption{Zero-shot example of \texttt{logic2text} task from formal-language-to-text category of \unifiedskg benchmark.}
\end{figure}

\begin{figure}[h]
\centering
\begin{tabular}{p{0.98\textwidth}}
\cellcolor{oldlace} table fact: $\langle$page\_title$\rangle$  Bruce Weber (basketball) $\langle$/page\_title$\rangle$  $\langle$section\_title$\rangle$  Head coaching record $\langle$/section\_title$\rangle$  $\langle$table$\rangle$  $\langle$cell$\rangle$  Illinois Fighting Illini (Big Ten Conference) (2003–2012) $\langle$col\_header$\rangle$  Season $\langle$/col\_header$\rangle$  $\langle$col\_header$\rangle$  Team $\langle$/col\_header$\rangle$  $\langle$col\_header$\rangle$  Overall $\langle$/col\_header$\rangle$  $\langle$col\_header$\rangle$  Conference $\langle$/col\_header$\rangle$  $\langle$col\_header$\rangle$  Standing $\langle$/col\_header$\rangle$  $\langle$col\_header$\rangle$  Postseason $\langle$/col\_header$\rangle$  $\langle$/cell$\rangle$  $\langle$cell$\rangle$  2007–08 $\langle$col\_header$\rangle$  Season $\langle$/col\_header$\rangle$  $\langle$/cell$\rangle$  $\langle$cell$\rangle$  16–19 $\langle$col\_header$\rangle$  Overall $\langle$/col\_header$\rangle$  $\langle$/cell$\rangle$  $\langle$cell$\rangle$  5–13 $\langle$col\_header$\rangle$  Conference $\langle$/col\_header$\rangle$  $\langle$/cell$\rangle$  $\langle$/table$\rangle$ \vspace{1pt} \newline textual description: \textcolor{cadmiumgreen}{Weber finished the 2007–08 Illini season with an overall record of 16–19 and 5–13 in the Big Ten.} \\
\end{tabular}
\caption{Zero-shot example of \texttt{totto} task from data to text category of \unifiedskg benchmark.}
\end{figure}


\begin{figure}
\centering
\begin{tabular}{p{0.98\textwidth}}
\cellcolor{oldlace} Select the sentence without English grammar error from the given pair. sentence 1: The took lights weren't orange. sentence 2: The taken lights weren't orange. A: \textcolor{cadmiumgreen}{sentence 2} \\
\end{tabular}
\caption{Zero-shot example of \texttt{blimp\_\_irregular\_past\_participle\_adjectives} task from linguistic probing category of CrossFit benchmark.}
\end{figure}

\begin{figure}
\centering
\begin{tabular}{p{0.98\textwidth}}
\cellcolor{oldlace} Select the sentence without English grammar error from the given pair. sentence 1: Kimberley has really ever found those actresses to compete. sentence 2: Kimberley has not ever found those actresses to compete. A: \textcolor{cadmiumgreen}{sentence 2} \\
\end{tabular}
\caption{Zero-shot example of \texttt{blimp\_\_sentential\_negation\_npi\_licensor\_present} task from linguistic probing category of CrossFit benchmark.}
\end{figure}

\begin{figure}
\centering
\begin{tabular}{p{0.98\textwidth}}
\cellcolor{oldlace} Why do some pimples hurt and swell more than others?\vspace{1pt} \newline Answer: \textcolor{cadmiumgreen}{Scalp pimples are the best: you're out and about with your day and feel an itch on your head and go to scratch it only to apply too much pressure to the overzealous zit and it bursts in an orgiastic explosion of bliss.} \\
\end{tabular}
\caption{Zero-shot example of \texttt{eli5\_\_eli5} task from question answering category of CrossFit benchmark.}
\end{figure}

\begin{figure}
\centering
\begin{tabular}{p{0.98\textwidth}}
\cellcolor{oldlace} Why do some pimples hurt and swell more than others?\vspace{1pt} \newline Answer: \textcolor{cadmiumgreen}{Scalp pimples are the best: you're out and about with your day and feel an itch on your head and go to scratch it only to apply too much pressure to the overzealous zit and it bursts in an orgiastic explosion of bliss.} \\
\end{tabular}
\caption{Zero-shot example of \texttt{eli5\_\_eli5} task from question answering category of CrossFit benchmark.}
\end{figure}

\begin{figure}
\centering
\begin{tabular}{p{0.98\textwidth}}
\cellcolor{oldlace} Predict the content that fills in the [MASK] placeholder to complete each fact statement:  Tiv people is located in [MASK] . Output: \textcolor{cadmiumgreen}{Nigeria} \\
\end{tabular}
\caption{Zero-shot example of \texttt{lama\_\_trex} task from information extraction category of CrossFit benchmark.}
\end{figure}

\begin{figure}
\centering
\begin{tabular}{p{0.98\textwidth}}
\cellcolor{oldlace} Predict the content that fills in the [MASK] placeholder to complete each fact statement:  Something you might do while helping someone is [MASK].\vspace{1pt} \newline output: \textcolor{cadmiumgreen}{dying} \\
\end{tabular}
\caption{Zero-shot example of \texttt{lama\_\_conceptnet} task from information extraction category of CrossFit benchmark.}
\end{figure}

\begin{figure}
\centering
\begin{tabular}{p{0.98\textwidth}}
\cellcolor{oldlace} Generate a normalized form of the entity denoted in the given passage with [START\_ENT] and [END\_ENT]. Passage: ATHLETICS - SALAH HISSOU BREAKS 10,000 METRES WORLD RECORD . [START\_ENT] BRUSSELS [END\_ENT] 1996-08-23  Morocco 's Salah Hissou broke the men 's 10,000 metres world record on Friday when he clocked 26 minutes 38.08 seconds at the Brussels grand prix on Friday .  The previous mark of 26:43.53 was set by Ethiopia 's Haile Gebreselassie in the Dutch town of Hengelo in June last year .\vspace{1pt} \newline Output: \textcolor{cadmiumgreen}{Brussels} \\
\end{tabular}
\caption{Zero-shot example of \texttt{kilt\_ay2} task from entity linking category of CrossFit benchmark.}
\end{figure}

\begin{figure}
\centering
\begin{tabular}{p{0.98\textwidth}}
\cellcolor{oldlace} Generate a normalized form of the entity denoted in the given passage with [START\_ENT] and [END\_ENT]. Passage: ATHLETICS - SALAH HISSOU BREAKS 10,000 METRES WORLD RECORD . [START\_ENT] BRUSSELS [END\_ENT] 1996-08-23  Morocco 's Salah Hissou broke the men 's 10,000 metres world record on Friday when he clocked 26 minutes 38.08 seconds at the Brussels grand prix on Friday .  The previous mark of 26:43.53 was set by Ethiopia 's Haile Gebreselassie in the Dutch town of Hengelo in June last year .\vspace{1pt} \newline Output: \textcolor{cadmiumgreen}{Brussels} \\
\end{tabular}
\caption{Zero-shot example of \texttt{kilt\_ay2} task from entity linking category of CrossFit benchmark.}
\end{figure}

\begin{figure}
\centering
\begin{tabular}{p{0.98\textwidth}}
\cellcolor{oldlace} In this sentence, replace verbs with their synonyms. \vspace{1pt} \newline \vspace{1pt} \newline  renewed efforts to fight drugs and crime will be costly \vspace{1pt} \newline \vspace{1pt} \newline  Answer: \textcolor{cadmiumgreen}{renewed efforts to fight\_back drugs and crime will be costly} \\
\end{tabular}
\caption{Zero-shot example of \texttt{styleptb} task from style transfer category of \exmix benchmark.}
\end{figure}

\begin{figure}
\centering
\begin{tabular}{p{0.98\textwidth}}
\cellcolor{oldlace} How would Shakespeare write the following sentence? \vspace{1pt} \newline \vspace{1pt} \newline  True , `` she said , ``a fine little one." ``No , `` I said , ``a huge wit." ``Right , `` she said , ``a hugely awful one." ``No , `` I said , ``he has a good wit." ``Exactly , `` she said , ``it's good and mild ; it doesn't hurt anyone." ``No , `` I said , ``Benedick is wise." ``He is certainly , `` she said , She cried a lot at that , and said she didn't care . \vspace{1pt} \newline \vspace{1pt} \newline  Answer: \textcolor{cadmiumgreen}{True , `` said she , ``a fine little one." ``No , `` said I , ``a great wit." ``Right , `` says she , ``a great gross one." ``Nay , `` said I , ``a good wit." ``Just , `` said she , ``it hurts nobody." ``Nay , `` said I , ``the gentleman is wise." ``Certain , `` said she , ``a wise gentleman." ``Nay , `` said I , ``he hath the tongues." ``That I believe , `` said she , ``for he swore a thing to me on Monday night , which he forswore on Tuesday morning ; there's a double tongue , there's two tongues." Thus did she an hour together transshape thy particular virtues .} \\
\end{tabular}
\caption{Zero-shot example of \texttt{shakespearizingmodernenglish} task from style transfer category of \exmix benchmark.}
\end{figure}

\begin{figure}
\centering
\begin{tabular}{p{0.98\textwidth}}
\cellcolor{oldlace} Transform the following sentence into funql. \vspace{1pt} \newline \vspace{1pt} \newline  what is the cost of a flight from boston to san francisco \vspace{1pt} \newline \vspace{1pt} \newline  Answer: \textcolor{cadmiumgreen}{answer(\_fare(\_flight(intersection(\_from\_2(city\_name(boston)),\_to\_2(city\_name(san\_francisco))))))} \\
\end{tabular}
\caption{Zero-shot example of \texttt{unimer} task from semantic parsing category of \exmix benchmark.}
\end{figure}

\begin{figure}
\centering
\begin{tabular}{p{0.98\textwidth}}
\cellcolor{oldlace} Construct a semantic representation of the following sentence. \vspace{1pt} \newline \vspace{1pt} \newline  Victoria awarded a cake to Olivia . \vspace{1pt} \newline \vspace{1pt} \newline  Answer: \textcolor{cadmiumgreen}{award . agent ( x \_ 1 , Victoria ) AND award . theme ( x \_ 1 , x \_ 3 ) AND award . recipient ( x \_ 1 , Olivia ) AND cake ( x \_ 3 )} \\
\end{tabular}
\caption{Zero-shot example of \texttt{cogs} task from semantic parsing category of \exmix benchmark.}
\end{figure}

\begin{figure}
\centering
\begin{tabular}{p{0.98\textwidth}}
\cellcolor{oldlace} Which Wikipedia entity does the marked entity correspond to? \vspace{1pt} \newline \vspace{1pt} \newline  Gmina Jaraczewo is a rural gmina ( administrative district ) in Jarocin County , Greater Poland Voivodeship , in west-central Poland .  Its seat is the village of Jaraczewo , which lies approximately west of Jarocin and south-east of the regional capital Poznań .  The gmina covers an area of , and as of 2006 its total population is 8,281 .  Gmina Jaraczewo contains the villages and settlements of Bielejewo , Brzostów , [START\_ENT] Cerekwica [END\_ENT] , Gola , Góra , Jaraczewo , Łobez , Łobzowiec , Łowęcice , Łukaszewo , Niedźwiady , Nosków , Nowa Cerekwica , Panienka , Parzęczew , Poręba , Rusko , Strzyżewko , Suchorzewko , Wojciechowo and Zalesie .  Gmina Jaraczewo is bordered by the gminas of Borek Wielkopolski , Dolsk , Jarocin , Koźmin Wielkopolski , Książ Wielkopolski and Nowe Miasto nad Wartą \vspace{1pt} \newline \vspace{1pt} \newline  Answer: \textcolor{cadmiumgreen}{Cerekwica, Jarocin County} \\
\end{tabular}
\caption{Zero-shot example of \texttt{wned} task from knowledge-intensive language tasks category of \exmix benchmark.}
\end{figure}

\begin{figure}
\centering
\begin{tabular}{p{0.98\textwidth}}
\cellcolor{oldlace} Which Wikipedia entity does the marked entity correspond to? \vspace{1pt} \newline \vspace{1pt} \newline  Afghan leader tells U.S. Congressman of peace plan .  Sayed Salahuddin  KABUL 1996-08-27  Afghan government military chief Ahmad Shah Masood briefed visiting U.S. Congressman Dana Rohrabacher on Tuesday on a peace plan for his wartorn country .  A spokesman for Masood said he had told the California Republican at a meeting in northern Kabul that President Burhanuddin Rabbani 's government favoured talks with all Afghan factions to set up an interim government .  The factions should agree to appoint a transitional leader , draft a new constitution , collect heavy weapons , create a national army and hold free elections in which the transitional leader would be barred from standing , he added .  Rohrabacher flew into Bagram military airbase north of Kabul in a Red Cross plane on Tuesday after meeting northern opposition militia leader General Abdul Rashid Dostum .  Masood 's spokesman Amrollah ( one name ) said Rohrabacher had recently visited Italy , Saudi Arabia and Pakistan as part of a mission to promote peace in [START\_ENT] Afghanistan [END\_ENT] .  `` We are certainly serious more than before to find a solution to the Afghan problem and support every U.N. plan , `` Amrollah quoted Rohrabacher as saying .  However , a spokesman for Prime Minister Gulbuddin Hekmatyar , a long-time rival of Masood , expressed concern at signs of renewed U.S. interest in Afghanistan .  `` America wants to block the establishment of a strong Islamic government in Afghanistan and the U.S. intends to neutralise the Afghan peace process initiated by the Afghans themselves , `` said the spokesman , Hamid Ibrahimi .  `` A great game has been started in Afghanistan as America feels that Tehran and Moscow have got stronger in the Afghan picture -- something Washington wants to change , `` he said .  Rohrabacher was expected to visit neutral faction leaders in the eastern city of Jalalabad and meet leaders of the rebel Islamic Taleban militia in the southern city of Kandahar .  Afghan guerrilla factions have been locked in a bloody power struggle since the fall of the communist government in April 1992 .  Hekmatyar , once Rabbani 's main rival , made a peace pact with him and rejoined the government as prime minister in June . \vspace{1pt} \newline \vspace{1pt} \newline  Answer: \textcolor{cadmiumgreen}{Afghanistan} \\
\end{tabular}
\caption{Zero-shot example of \texttt{aidayago2} task from knowledge-intensive language tasks category of \exmix benchmark.}
\end{figure}

\begin{figure}
\centering
\begin{tabular}{p{0.98\textwidth}}
\cellcolor{oldlace} What Lakers legend visited LeBron James less than 24 hours before the free agent became a Laker? A: \textcolor{cadmiumgreen}{Magic Johnson} \\
\end{tabular}
\caption{Zero-shot example of \texttt{newsquizqa} task from question answering category of ExMix benchmark.}
\end{figure}

\begin{figure}
\centering
\begin{tabular}{p{0.98\textwidth}}
\cellcolor{oldlace} What Lakers legend visited LeBron James less than 24 hours before the free agent became a Laker? A: \textcolor{cadmiumgreen}{Magic Johnson} \\
\end{tabular}
\caption{Zero-shot example of \texttt{newsquizqa} task from question answering category of \exmix benchmark.}
\end{figure}

\begin{figure}
\centering
\begin{tabular}{p{0.98\textwidth}}
\cellcolor{oldlace} Is the following comment toxic? \vspace{1pt} \newline \vspace{1pt} \newline  ``:Thanks for the comment and bringing this to my attention. This is my first time I have voted using that format. I didn't notice the comments at top and bottom  instead, I went directly to my comment. If you look look at the change I made, I added two words (my original vote with a strikethrough) that I had mistakenly deleted, instead of using the common method of altering one's vote by doing a strikethrough. So, I re-added the deleted words of my previous vote with a strikethrough.  \vspace{1pt} \newline \vspace{1pt} \newline On reflection, I think there are meaningful exceptions to rules. The change does not alter my comment or revised ``"vote.""  It preserves information about the change in status of my vote. In fact, it clarifies the vote change. So, in this case, as the intention of the request of do not edit is to keep a clear record of the vote, I think my very minor edit actually serves that. So, I'll leave it as is. Thanks,  \vspace{1pt} \newline \vspace{1pt} \newline ``
\vspace{1pt} \newline answer: \textcolor{cadmiumgreen}{No} \\
\end{tabular}
\caption{Zero-shot example of \texttt{wikipedia\_toxicity\_subtypes} task from toxic language detection category of \exmix benchmark.}
\end{figure}

\begin{figure}
\centering
\begin{tabular}{p{0.98\textwidth}}
\cellcolor{oldlace} Is the following comment toxic? \vspace{1pt} \newline \vspace{1pt} \newline  ``:Thanks for the comment and bringing this to my attention. This is my first time I have voted using that format. I didn't notice the comments at top and bottom  instead, I went directly to my comment. If you look look at the change I made, I added two words (my original vote with a strikethrough) that I had mistakenly deleted, instead of using the common method of altering one's vote by doing a strikethrough. So, I re-added the deleted words of my previous vote with a strikethrough.  \vspace{1pt} \newline \vspace{1pt} \newline On reflection, I think there are meaningful exceptions to rules. The change does not alter my comment or revised ``"vote.""  It preserves information about the change in status of my vote. In fact, it clarifies the vote change. So, in this case, as the intention of the request of do not edit is to keep a clear record of the vote, I think my very minor edit actually serves that. So, I'll leave it as is. Thanks,  \vspace{1pt} \newline \vspace{1pt} \newline ``\vspace{1pt} \newline answer: \textcolor{cadmiumgreen}{No} \\
\end{tabular}
\caption{Zero-shot example of \texttt{wikipedia\_toxicity\_subtypes} task from toxic language detection category of \exmix benchmark.}
\end{figure}

\begin{figure}
\centering
\begin{tabular}{p{0.98\textwidth}}
\cellcolor{oldlace} summarize: Editor's note: Ruben Navarrette is a nationally syndicated columnist and a member of the editorial board of the San Diego Union-Tribune. Read his column here. Ruben Navarrette asks whether Obama's cool, calm demeanor will be a plus in negotiating with world leaders. SAN DIEGO, California (CNN) -- Make no mistake, Barack Obama is one cool customer. Now, after the last debate, it seems all but certain that the Iceman cometh to the White House. Radio talk show hosts and rank-and-file Republicans spent the last few weeks pleading with John McCain to take the gloves off and take the fight to Obama. How's that working out, folks? In this week's match-up, Obama snatched the gloves out of McCain's hands and slapped him silly with them. I suppose the hope was that Obama would get rattled and make a mistake. But Obama doesn't get rattled or make many mistakes. I still have no idea what type of president Obama would make. But he's an extraordinary politician. In fact, he may even be better than Bill Clinton who, while he had the IQ and EQ, also had the burden of a legendary red-hot temper. Obviously, it takes a lot to get under Obama's skin. McCain sure tried. Maybe this is the guy we want negotiating with world leaders. Maybe after eight years of George W. Bush stubbornness, on the heels of eight years of Clinton emotiveness, we need to send out for ice. In a CNN/Opinion Research Corp. poll, 58 percent of those who watched the debate said Obama did the better job and 31 percent said that about McCain. That makes three skins for Obama. In earlier polls, 54 percent of those who watched thought Obama won the second debate, and 51 percent thought he won the first one. This week, McCain turned in his best performance of the debates, and the first 30 minutes -- with McCain bringing up Obama's problematic encounter with the now famous Joe the Plumber; and the quip about how he isn't Bush and how Obama should have run four years ago -- were near flawless for the Republican.  iReport.com: Are you Joe the Plumber? Get out your plunger and share your thoughts. McCain put Obama on the defensive, and it showed. If McCain had been that aggressive throughout the first two debates -- firm but not necessary unlikable -- we might be looking at a different race right now. But, over the next hour, Obama regained his stride and eventually dominated the exchange.  And, in the end, with his sarcastic crack about school vouchers -- ``Because there's not enough vouchers; therefore, we shouldn't do it, even though it's working. I got it." -- McCain was profoundly unlikable. So said the polls. In the CNN/Opinion Research Corp. poll, 70 percent of debate watchers found Obama more likable. Only 22 percent said that about McCain. McCain's supporters wanted him to bring up some of the allegedly shady characters from Obama's circle of acquaintances that give some Americans pause and lead them to question the Democrat's values. There are good reasons to have that conversation, and bad ones. A friend and fellow journalist told me Obama's Chicago posse was important because it formed ``the political womb where the fetal Obama grew into a politician. \textbf{$[$...continued$]$}
But, unless the political wind changes -- and quickly -- a promotion to the other end of Pennsylvania Avenue doesn't appear to be in the cards. The opinions expressed in this commentary are solely those of Ruben Navarrette.\vspace{1pt} \newline answer: \textcolor{cadmiumgreen}{Ruben Navarrette: McCain tried to get under Obama's skin with attacks.\vspace{1pt} \newline Obama remained cool and wound up ending strongly in the debate, he says.\vspace{1pt} \newline Navarrette says Ayers attacks backfired even though issue is legitimate.\vspace{1pt} \newline America may need the cool-headedness of Barack Obama, he says.} \\
\end{tabular}
\caption{Zero-shot example of \texttt{cnn\_dailymail\_v002} task from summarization category of T5 benchmark. We remove some text from the input and replace it with `\textbf{$[$...continued$]$}' for presentation purpose.}
\end{figure}

\begin{figure}
\centering
\begin{tabular}{p{0.98\textwidth}}
\cellcolor{oldlace} Answer the following question by reasoning step by step.\vspace{1pt} \newline \vspace{1pt} \newline `` Ai n't No Sunshine '' is a song by Bill Withers from his 1971 album Just As I Am , produced by Booker T. Jones . The record featured musicians Donald `` Duck '' Dunn on bass guitar , Al Jackson , Jr. on drums and Stephen Stills on guitar . String arrangements were done by Booker T. Jones , and recorded in Memphis by engineer Terry Manning . The song is in the key of A minor .\vspace{1pt} \newline who did ain't no sunshine when she's gone Output: \textcolor{cadmiumgreen}{`` Ai n't No Sunshine '' is a song by Bill Withers from his 1971 album Just As I Am , produced by Booker T. Jones .  The answer is Bill Withers} \\
\end{tabular}
\caption{Zero-shot example of \texttt{qed} task from Reasoning benchmark.}
\end{figure}

\begin{figure}
\centering
\begin{tabular}{p{0.98\textwidth}}
\cellcolor{oldlace} Answer the following question by reasoning step by step.\vspace{1pt} \newline \vspace{1pt} \newline How do most people feel about a person they love?\vspace{1pt} \newline popularity, know all, own house, care about, flu Output: \textcolor{cadmiumgreen}{we care about people we love. The answer is care about} \\
\end{tabular}
\caption{Zero-shot example of \texttt{cose} task from Reasoning benchmark.}
\end{figure}

\end{document}